THE UNIVERSITY OF EDINBURGH

MASTER DISSERTATION REPORT

# Generative Adversarial Network (GAN) based Image-Deblurring

*Author: Yuhong Lu*         *Supervisor: Dr Nick Polydorides*

*A thesis submitted in fulfilment of the requirements*
*for the degree of MSc. in Signal Processing and Communications*

*in the*

School of Engineering

October 2021

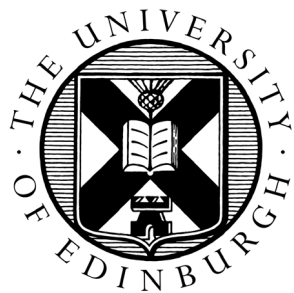

# Declaration of Authorship

I, Yuhong Lu, declare that this dissertation report titled, 'Generative Adversarial Network (GAN) based Image Deblurring' and the work presented in is my own work. Any uses made within it of the works of other authors in any form (e.g., ideas, equations, figures, text, tables, programs) are properly acknowledged at any point of their use. A list of the references employed is included.

Signed: *Yuhong Lu*

Date: 15 October 2021

# *Abstract*

This thesis analyzes the challenging problem of Image Deblurring based on classical theorems and state-of-art methods proposed in recent years. By spectral analysis we mathematically show the effective of spectral regularization methods, and point out the linking between the spectral filtering result and the solution of the regularization optimization objective. For ill-posed problems like image deblurring, the optimization objective contains a regularization term (also called the regularization functional) that encodes our prior knowledge into the solution. We demonstrate how to craft a regularization term by hand using the idea of maximum a posterior estimation. Then, we point out the limitations of such regularization-based methods, and step into the neural-network based methods.

Based on the idea of Wasserstein generative adversarial models, we can train a CNN to learn the regularization functional. Such data-driven approaches are able to capture the complexity, which may not be analytically modellable. Besides, in recent years with the improvement of architectures, the network has been able to output an image closely approximating the ground truth given the blurry observation. The Generative Adversarial Network (GAN) works on this Image-to-Image translation idea. We analyze the DeblurGAN-v2 method proposed by Orest Kupyn *et al*. [14] in 2019 based on numerical tests. And, based on the experimental results and our knowledge, we put forward some suggestions for improvement on this method.

# *Acknowledgements*

We are grateful for the MATLAB functions [43] provided by authors of the book, *Deblurring Images: Matrices, Spectra and Filtering*. We are also grateful for the open-source project and datasets provided by Orest Kupyn *et al*. [14]. And my supervisor, Dr Nick Polydorides, provides many constructive suggestions for the project. We have happy discussions in regular meetings.

# Symbols

| | |
|---|---|
| **B** | The blurry observation in its matrix form |
| **b** | The blurry observation in its vector form |
| **X**$_{true}$ | The ground truth image in its matrix form |
| **x**$_{true}$ | The ground truth image in its vector form |
| **E** | The additive noise in its matrix form |
| **e** | The additive noise in its vector form |
| **X**$_{recon}$ | The reconstructed image in its matrix form |
| **x**$_{recon}$ | The reconstructed image in its vector form |
| vec () | Vectorize a matrix by stacking its columns |
| vec$^{-1}$ () | The inverse function of vec () to turn a vector back to its matrix |
| **A** | The forward blurring operator |
| **A**$_r$ | The row blurring matrix |
| **A**$_c$ | The column blurring matrix |
| **P** | The Point Spread Function (PSF) |
| * | Mathematical convolution |
| $\otimes$ | Kronecker product |
| tr () | Trace of a matrix |
| Prob () | The probability of something |

# Contents

# Chapter 1

# Introduction

    Image Deblurring techniques find their applications in places where a restoration from blurry images favors follow-up works, such as object detection [1] and computer-aided diagnosis [2]. Blurs come for various reasons, in which out-of-focus and motion-induced are two typical examples. Deblurring approaches branch into several classes, namely the non-blind, semi-blind, and blind methods subject to the degree of our accessibility to the blurring kernel (or, equivalently, the Point-Spread Function we convolve the image with). If, for example, we are dealing with an out-of-focus blur in the scenario of atmospheric turbulence, then a parameterized 2-dimensional Gaussian-like PSF model [3] would be plausible. On the other hand, if we encounter a motion blur where the camera's movements during exposure are much more complicated, we have less prior knowledge of the actual PSF. Advanced Blind Image Deblurring (BID) approaches require dedicated design, but its success can build upon the experience and theoretical fundamentals established for non-Blind methods.

    The crux of deblurring lies in the ill-conditioned nature of the blurring operator. It means that the presence of even a tiny amount of noise can dramatically impact, in the form of the inverted noise [4], the quality of the (pseudo-) inversed solution. Noises are common and ubiquitous, and the additive noises in CCD camera alone can come in three forms [5]:

- the Poisson noise from fluctuated background photons,
- the readout noise from imperfectness of electronics, as well as
- the quantization error from the analog-to-digital conversion.

So, a successful deblurring method should be able to cope appropriately with the inverted noise and therefore produce a faithful reconstruction.

    In terms of the regularization-based methods, many treat deblurring as an optimization problem, with the objective being a weighted summation of the residual norm and a regularization term [6]. Almeida *et al*. [7] designed their optimization objective following the idea of maximum a posteriori estimation. A highlight is a prior form they used, assuming that edges around the image are sparse, to find the shape of the regularization term. Aware of humans' prohibitive workload to craft a regularization term, Lunz *et al*. [8] proposed an algorithm to train a critic network instead that can tell apart the authentic images and pseudo-inversed ones. After training, the network works as the regularization term in the optimization objective. The reconstructed images are then obtained from pseudo-inversed ones by gradient descent on that optimization objective learned. This method can receive good results on the LIDC/IDRI database [9] with PSNR over 30 dB and SSIM over 0.92. There are several points worth attention here for regularization-based approaches. One of them is the regularization (hyper) parameter, *i.e.*, the suitable amount of regularization. It is not a learnable parameter in the above cases but is strongly subjected to our observations, *e.g.*, the residual norm

and the noise in the data. So, if we want to generalize such regularization-based methods, then a better approach for choosing this parameter, not the trial-and-error ones based on experience, needs to be developed.

During the recent ten years, the significant upgrade of computing power has aroused great research passions on deep learning. In 2014, Ian et al. [10] suggested a novel idea of Generative Adversarial Network (GAN), which learns the probability distribution of the training data and generates samples closely approximating the real ones. After that, GAN began its attempts in various image-to-image translation tasks, including style transfer [11], where the network learns to add artistic effects on input photographs. Isola et al. [12] summarized previous works of colleagues and put forward a conditional GAN architecture known as pix2pix. Treating deblurring similarly as an image-to-image translation work and inspired by the pix2pix architecture, Orest Kupyn *et al*. [13][14] suggested DeblurGAN and DeblurGAN-v2, the updated version, in 2017 and 2019 respectively to deal with BID. In DeblurGAN-v2, they introduced Feature Pyramid Network (FPN) architecture for the generator and experimented, on the GoPro [15], DVD [16], and NFS [17] datasets, with multiple types of backbones, including Inception-ResNet-v2 [18], MobileNet [19], and MobileNet-DSC [20]. By the paper, their scheme receives good reconstructions from motion blurs (as shown by Figure 1). Their open-source project is available [21] on GitHub.

This thesis aims to analyze DeblurGAN-v2, the advanced approach proposed in 2019 for the BID. However, before we step into such a sophisticated method, a formal mathematical description of the image restoration problem for BID is given, along with a detailed literature review on forerunners' methods, *e.g.*, regularization-based approaches for blind or non-blind tasks. Elaborations on these methods help to draw a comparison with DeblurGAN-v2. Moreover, during that process, the merits and necessity of the latter are demonstrated as well. After analyzing the architectures and the numerical test results of DeblurGAN-v2, we seek to identify limitations, suggest improvements to this scheme and test the performance of any modifications on datasets available.

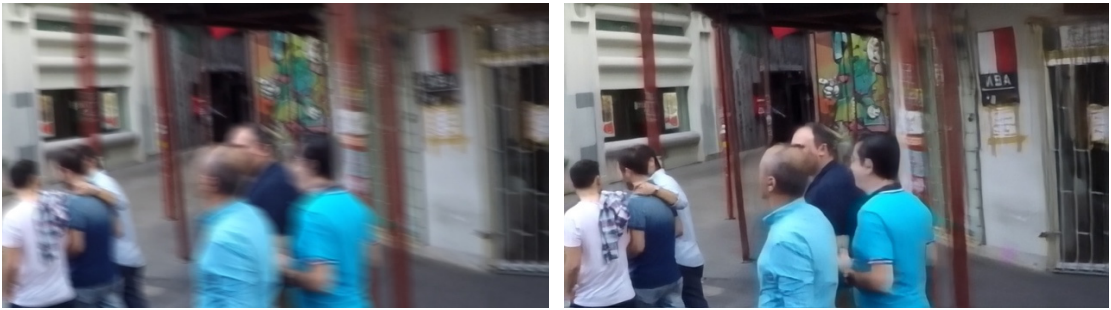

***Figure 1.*** *A perceptual quality test of DeblurGAN-v2 on the GoPro test dataset* [15], *where the blurry (left) image goes through a pre-trained model (available at* [21]*) with the InceptionResNet-v2 generator to produce a sharp (right) restoration.*

# Chapter 2

# Preliminary Works

In this Chapter we define the image restoration work for the Blind Image Deblurring (BID) problem. Based on a linear blurring model the generic solution form is given. Then for a better understanding of the blurring mechanism, the physical meaning of the Point Spread Function (PSF) is explained mathematically using the 2D convolution definition. Finally, we introduce the types of blurring we are going to deal with in this thesis.

## 2.1 A generic solution form for the Blind Image Deblurring (BID) problem

Suppose a color image $\mathbf{X}_{\text{true}} \in \mathbb{R}^{m \times n \times 3}$ undergoes a blurring process and turns into $\mathbf{B} \in \mathbb{R}^{m \times n \times 3}$ by a linear model

$$\mathbf{B} = \begin{bmatrix} \mathbf{B}(:,:,1) \\ \mathbf{B}(:,:,2) \\ \mathbf{B}(:,:,3) \end{bmatrix} = \begin{bmatrix} \mathbf{X}_{\text{true}}(:,:,1) * \mathbf{P} \\ \mathbf{X}_{\text{true}}(:,:,2) * \mathbf{P} \\ \mathbf{X}_{\text{true}}(:,:,3) * \mathbf{P} \end{bmatrix} + \begin{bmatrix} \mathbf{E}(:,:,1) \\ \mathbf{E}(:,:,2) \\ \mathbf{E}(:,:,3) \end{bmatrix} \tag{2.1-1}$$

where

$$\begin{cases} \mathbf{B}(:,:,i) \in \mathbb{R}^{m \times n} \text{ is the blurry image of the } i^{\text{th}} \text{ channel,} \\ \mathbf{X}_{\text{true}}(:,:,i) \in \mathbb{R}^{m \times n} \text{ is the underlying true image of the } i^{\text{th}} \text{ channel,} \\ \mathbf{P} \in \mathbb{R}^{k \times k} \text{ is the common 2D } \textit{unknown blurring kernel} \text{ (or } \textit{Point Spread Function} \text{ (}PSF\text{)) with odd } k \ll \min\{m,n\} \text{ applied to all channels,} \\ \mathbf{E}(:,:,i) \in \mathbb{R}^{m \times n} \text{ is the noise term of the } i^{\text{th}} \text{ channel, and} \\ * \text{ denotes the mathematical convolution operation.} \end{cases}$$

(2.1-2)

Now, given the blurry observation $\mathbf{B}$, we aim to reconstruct a sharp $\mathbf{X}_{\text{recon}} \in \mathbb{R}^{m \times n \times 3}$ as well as the blurring kernel $\mathbf{P}_{\text{recon}} \in \mathbb{R}^{k \times k}$ which satisfy

$$\langle \mathbf{X}_{\text{recon}}, \mathbf{P}_{\text{recon}} \rangle = \arg\max_{\langle \mathbf{X}, \mathbf{P} \rangle} \text{Prob}(\mathbf{X}, \mathbf{P} | \mathbf{B}) \tag{2.2}$$

where $\text{Prob}(\mathbf{X}, \mathbf{P} | \mathbf{B})$ is the Joint Posterior Probability of the $\langle \mathbf{X}, \mathbf{P} \rangle$ pair given $\mathbf{B}$.

The ideas behind (2.2) are straightforward. Given only a blurry image on hand, we traverse and evaluate all possibilities of <sharp image, blurring pattern> pairs, and then pick one pair that best fits the observation. However, the BID problem is highly ill-posed because there could be infinite numbers of such <sharp image, blurring pattern> pairs that suit the blurry observation, and only a few of the "sharp images" reconstructed look like a plausible solution. The effectiveness of any generalized-well algorithm lies in whether it can correctly capture the distribution of pixels in a real-look sharp image.

Then how to decide whether a reconstructed "sharp image" captures the pixel distribution of a real-look sharp image? In regularization-based BID methods such as [7], Almeida et al. assume that edges in sharp images are sparse (*i.e.*, more zero values on the Image Gradient plot with only a few non-zero places denoting edges). In contrast, edges in blurry images are less sparse because they are smoothed out. So, the authors designed a regularization term to penalize reconstructions with smoothed edges and thereby obtained a more real-look sharp reconstruction. We will discuss this approach in detail in Section 3.2. In GAN-based BID methods, well-designed network architectures learn the abstract mapping between the real blurry images and their real sharp correspondences. So, if the training dataset is well-generalized and the network learns well on it, it is reasonable to believe that the learned model is widely applicable.

Another comment, as indicated by Levin et al. [22], is that in (2.2) the number of unknowns (*i.e.*, entries in **X** and **P**) greatly exceeds the number of knowns (*i.e.*, entries in **B**). It is therefore sometimes a hard job to solve < **X**, **P** > from **B** directly. An alternative approach proposed is to find first the blurring kernel

$$\mathbf{P}_{recon} = \arg\max\nolimits_{<\mathbf{P}>} \text{Prob}(\mathbf{P}|\mathbf{B}) \tag{2.3}$$

with $\text{Prob}(\mathbf{P}|\mathbf{B}) = \int_{\mathbf{X}} \text{Prob}(\mathbf{X}, \mathbf{P}|\mathbf{B}) d\mathbf{X}$ calculated by Variational Bayesian methods [22]. It is much easier obtain a sparse and small $\mathbf{P}_{recon}$ from **B** now. After that, with an estimated blurring kernel available, a reconstruction of $\mathbf{X}_{recon}$ is solvable by some non-blind deblurring methods.

To summarize this section, for the blurring model (2.1) with an unknown blurring kernel, we define, mathematically, the solution form using the concept of the maximum posterior probability. The severely ill-posed problem can be done by either regularization- or GAN- based methods, but any success lies in how well they capture the essence of what a sharp and real-look image should be. Another note is that some methods [22] tend to solve the blurring kernel first and then treat the remaining as a non-blind deblurring problem. It, to a certain extent, reminds us of the importance of the non-blind image deblurring techniques while we are on the way to dealing with BID.

Therefore, in the next chapter, we carry out a literature review on regularization- and neural network-based approaches for blind and non-blind problems. But before that, in the next section, we explore more on the physics meaning of **P**, the blurring kernel in (2.1). With its physics meaning explicitly demonstrated in mathematics, we understand better the blurring mechanism. And in Section 2.3, we show some PSF examples of the blurring types we are going to deal with in this thesis.

## 2.2 Physics meaning of the Point Spread Function (PSF)

In (2.1) we presented a convolution model with a convolution kernel **P**, but we did not explain what this kernel means and why it has another name – the Point Spread Function (PSF). In this section we try to describe these in mathematics. Suppose there is a 2D image $\mathbf{F} \in \mathbb{R}^{m \times n}$ with only one single bright pixel, we can model the image by

$$\mathbf{F} = (\mathbf{F}(i,j))_{m \times n}, \text{ where } \mathbf{F}(i,j) = \begin{cases} 1, & i = p, j = q \\ 0, & \text{otherwise.} \end{cases} \quad (2.4)$$

And there is 2D kernel $\mathbf{P} \in \mathbb{R}^{k \times k}$ of a much smaller dimension

$$\mathbf{P} = (\mathbf{P}(i,j))_{k \times k}, \text{ where } k \text{ is an odd number satisfying } k \ll \min\{m,n\}. \quad (2.5)$$

During the convolution, the small square window $\mathbf{P}$ shuffles across $\mathbf{F}$. Let us say, at a time the center of that window locates at $(i_0, j_0)$ to produce a value there by

$$\mathbf{G}(i_0, j_0) = \mathbf{F} * \mathbf{P}\bigg|_{i=i_0, j=j_0} = \sum_{i=i_0 - \frac{1}{2}(k-1)}^{i_0 + \frac{1}{2}(k-1)} \sum_{j=j_0 - \frac{1}{2}(k-1)}^{j_0 + \frac{1}{2}(k-1)} \mathbf{F}(i,j) \mathbf{P}(2i_0 - i, 2j_0 - j). \quad (2.6)$$

If the only bright pixel, which locates at $(p, q)$, is not covered by the square window of $\mathbf{P}$ centered at $(i_0, j_0)$, i.e.,

$$\min\{|p - i_0|, |q - j_0|\} > \frac{1}{2}(k-1), \quad (2.7\text{-}1)$$

then the convolution yields zero there: by (2.4) and (2.6)

$$\mathbf{G}(i_0, j_0) = 0, \text{ because } \mathbf{F}(i,j) = 0 \text{ for } \begin{cases} i_0 - \frac{1}{2}(k-1) \le i \le i_0 + \frac{1}{2}(k-1) \\ j_0 - \frac{1}{2}(k-1) \le j \le j_0 + \frac{1}{2}(k-1) \end{cases}. \quad (2.7\text{-}2)$$

In other places where the bright pixel is covered by the window, *i.e.*,

$$\max\{|p - i_0|, |q - j_0|\} \le \frac{1}{2}(k-1), \quad (2.8\text{-}1)$$

then non-zero result is obtained as

$$\begin{aligned} \mathbf{G}(i_0, j_0) &= \mathbf{F} * \mathbf{P}\bigg|_{i=i_0, j=j_0} = \sum_{i=i_0 - \frac{1}{2}(k-1)}^{i_0 + \frac{1}{2}(k-1)} \sum_{j=j_0 - \frac{1}{2}(k-1)}^{j_0 + \frac{1}{2}(k-1)} \mathbf{F}(i,j) \mathbf{P}(2i_0 - i, 2j_0 - j) \\ &= \mathbf{F}(p,q) \mathbf{P}(2i_0 - p, 2j_0 - q) \\ &= 1 \cdot \mathbf{P}(2i_0 - p, 2j_0 - q) \\ &= \mathbf{P}(2i_0 - p, 2j_0 - q). \end{aligned} \quad (2.8\text{-}2)$$

We can write (2.7) and (2.8) together in a more compact way, like

$$\mathbf{G}(i_0, j_0) = \begin{cases} 0, & \min\{|p - i_0|, |q - j_0|\} > \frac{1}{2}(k-1); \text{ and} \\ \mathbf{P}(2i_0 - p, 2j_0 - q), & \max\{|p - i_0|, |q - j_0|\} \le \frac{1}{2}(k-1) \end{cases}, \text{ with } \begin{cases} 0 \le i_0 \le m-1 \\ 0 \le j_0 \le n-1 \\ k \ll \min\{m,n\} \end{cases}. \quad (2.9)$$

Now let us view the appearance of the blurry image $\mathbf{G}$, which is transformed from the point source $\mathbf{F}$ defined by (2.4). In the neighborhood of the point source, covered by the square window of $\mathbf{P}$, $\mathbf{G}$ is a mirrored version of $\mathbf{P}$. In other regions far from that point source, $\mathbf{G}$ is dark as well. So, we can conclude that the convolution kernel $\mathbf{P}$ depicts the blurring pattern of a point source. Because of this, people sometimes refer to **P** by another name – Point Spread Function (PSF) – which illustrates the blurring pattern of one type of blur. We can think of **P** as the "spreading trace" of one pixel when blurring occurs.

Sometimes the PSF can be analytically modeled by a set of parameters, like in the atmospheric turbulence blur [3] which can be described by a 2D Gaussian, as we shall see in Section 2.3. Sometimes the PSF is rather complicated, and we cannot summarize it systematically, like in the case of the motion blur.

For BID, there is no way to access the exact blurring mechanism in advance, although sometimes we know what type of blur we are handling. In algorithms like the DeblurGAN-v2 approach, the network learns the mapping between the blurry data and the sharp target. If the network performs well on the test dataset, then the blurring mechanism hidden behind is well captured by it.

Another note on $\mathbf{P} \in \mathbb{R}^{k \times k}$ is about its dimension. Compared to images, **P** is a much smaller square window with odd width and length. It is then easy for storage on computers with its center easily spotted.

**2.3 Two blurring patterns to be handled**

Blurs in the world are either regular or irregular. Regular blurs are ones whose PSF can be systematically modeled by several parameters. PSFs of irregular blurs, to the opposite, follow random patterns and there is not a simplified model to summarize them. In this thesis, for regular blurs we consider PSFs which can be expressed as 2D Gaussians. And for irregular blurs we consider motion blurs.

Motion blurs arise from, say, the shakes of the camera and the movements of objects during exposure. The PSF pattern can therefore be arbitrary with great complexity if we think of them as the spreading trace of one pixel, as shown in Figure 2.

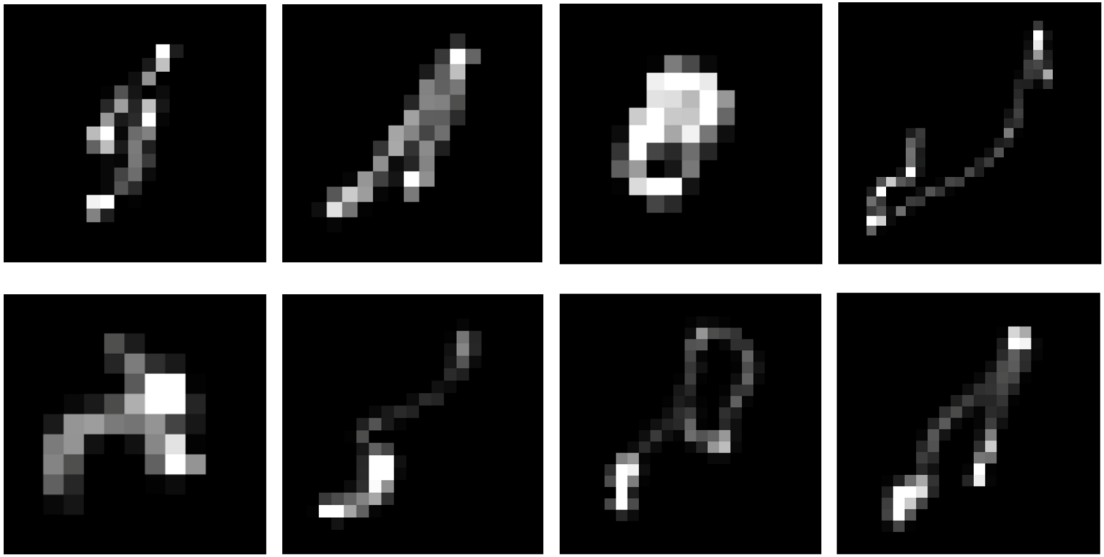

*Figure 2. Patterns of motion blur kernels obtained by Levin et al.* [22].

For Gaussian-like PSFs let us consider one type induced by atmospheric turbulence [3]. The PSF, centered at ($i_0, j_0$), can be systematically summarized as a 2D Gaussian

$$\mathbf{P} = \left(\mathbf{P}(i,j)\right)_{k\times k}, \text{ where } \mathbf{P}(i,j) \propto \exp\left(-\frac{1}{2}\begin{pmatrix}i-i_0\\j-j_0\end{pmatrix}^T \begin{pmatrix}\sigma_1^2 & \rho^2\\ \rho^2 & \sigma_2^2\end{pmatrix}^{-1} \begin{pmatrix}i-i_0\\j-j_0\end{pmatrix}\right)$$

(2.10-1)

in which

$$\begin{cases}\langle\sigma_1,\sigma_2\rangle \text{ denotes the } \langle\text{width, length}\rangle \\ \rho \text{ indicates the orientation}\end{cases} \text{ of the blurring kernel.}$$

(2.10-2)

Figure 3 shows several examples of such Gaussian-like PSFs with $k = 231$ and $(i_0, j_0) = (0,0)$. From (A) to (C) (or from (D) to (F)) where they share the same $\langle\sigma_1, \sigma_2\rangle$, we observe that the greater the $\rho$, the more oblique the pattern will be. When $\rho$ is the same, we observe (from the comparison of (A) and (D), or (B) and (E), or (C) and (F)) that $\langle\sigma_1, \sigma_2\rangle$ control the $\langle$width, length$\rangle$: the larger the $\sigma$, the wider the pattern will be in that direction.

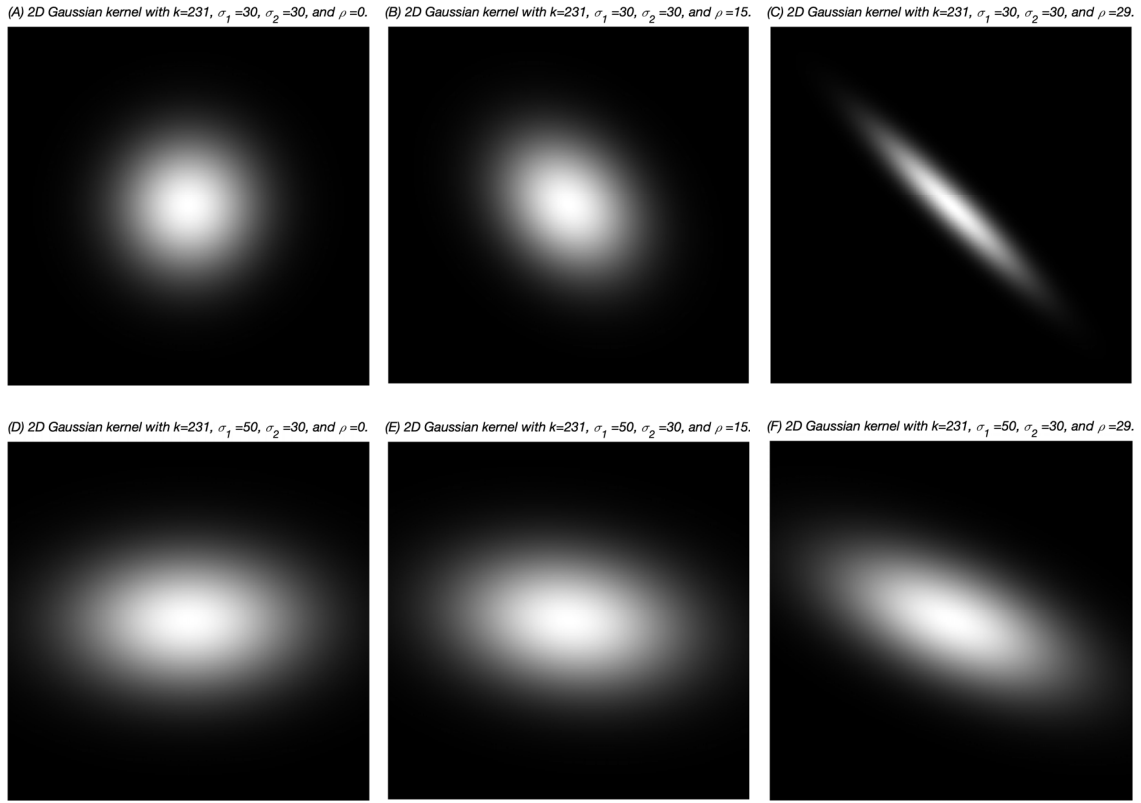

*Figure 3.* Gaussian-like PSFs with different sets of parameters.

# Chapter 3

# A literature review on Image Deblurring approaches

Before diving into the sophisticated DeblurGAN-v2 approach designed for the Blind Image Deblurring (BID) problem, it is beneficial to investigate the essence of Image Deblurring techniques theoretically. In this chapter, we focus on a literature review of the working mechanism of some model-based regularization approaches. We summarize their advantages and limitations, and then we compare them with those of neural network– based approaches. It helps to justify the necessity of the latter, for example the DeblurGAN-v2 approach which we will analyze in the next chapter.

## 3.1 Spectral Regularization Methods

Spectral Regularization methods are approaches using the Singular Value Decomposition (SVD) – or some other decompositions with spectral properties – on the Forward (blurring) Operator to introduce filtering or regularization in the image reconstructed. Spectral Analysis can help us to view the problem from the Frequency Domain, where phenomena are more convenient to explain mathematically. To compute the singular values of the operator, in this section we assume a known PSF.

### 3.1.1 An ill-conditioned problem

Image deblurring with a known PSF is an ill-conditioned problem vulnerable to the noise present in the data. Suppose the blurry observation $\mathbf{B} \in \mathbb{R}^{m \times n}$ is linearly associated with the underlying $\mathbf{X}_{\text{true}} \in \mathbb{R}^{m \times n}$ by a forward (blurring) operator $\mathbf{A}$ and a noise term $\mathbf{e}$ as

$$\text{vec}(\mathbf{B}) = \mathbf{A}\,\text{vec}(\mathbf{X}_{\text{true}}) + \mathbf{e}, \text{ where } N = m \cdot n, \text{ and } \begin{cases} \text{vec}(\mathbf{B}) \in \mathbb{R}^{N \times 1}, \\ \text{vec}(\mathbf{X}_{\text{true}}) \in \mathbb{R}^{N \times 1}, \\ \mathbf{e} \in \mathbb{R}^{N \times 1}, \text{ and} \\ \mathbf{A} \in \mathbb{R}^{N \times N}. \end{cases} \quad (3.1)$$

This model is, in principle, an alternative version of the convolution model introduced in (2.1). For theoretical analysis here we use (3.1), while a computer uses (2.1) to cause blurring since the kernel $\mathbf{P}$ is much more compact than $\mathbf{A}$ and thus easier to save in storage.

Now, assume a separable blur where the vertical and horizontal blur components are independent and thus can be separated, *i.e.*,

$$\mathbf{A} = \mathbf{A}_r \otimes \mathbf{A}_c, \text{ where } \begin{cases} \text{the row blurring matrix } \mathbf{A}_r \in \mathbb{R}^{n \times n}, \\ \text{the column blurring matrix } \mathbf{A}_c \in \mathbb{R}^{m \times m}, \text{ and} \\ \otimes \text{ denotes the Kronecker product.} \end{cases} \quad (3.2)$$

Then (3.1) becomes

$$\text{vec}(\mathbf{B}) = (\mathbf{A}_r \otimes \mathbf{A}_c) \text{vec}(\mathbf{X}_{\text{true}}) + \mathbf{e} = \text{vec}(\mathbf{A}_c \mathbf{X}_{\text{true}} \mathbf{A}_r^T) + \mathbf{e}. \quad (3.3)$$

Therefore, we can alternatively write (3.1) as

$$\mathbf{B} = \mathbf{A}_c \mathbf{X}_{\text{true}} \mathbf{A}_r^T + \mathbf{E}, \text{ where } \begin{cases} \mathbf{B} \in \mathbb{R}^{m \times n} \text{ is the blurry image,} \\ \mathbf{X}_{\text{true}} \in \mathbb{R}^{m \times n} \text{ is the underlying true image,} \\ \mathbf{A}_c \in \mathbb{R}^{m \times m} \text{ is the column blurring matrix operating on columns of } \mathbf{X}_{\text{true}}, \\ \mathbf{A}_r \in \mathbb{R}^{n \times n} \text{ is the row blurring matrix operating on rows of } \mathbf{X}_{\text{true}}, \text{ and} \\ \mathbf{E} \in \mathbb{R}^{m \times n} \text{ is the noise matrix rearranged from } \mathbf{e}, \text{ i.e., } \mathbf{E} = \text{vec}^{-1}(\mathbf{e}). \end{cases}$$

(3.4)

Given the observation $\mathbf{B}$ and the blurring matrices $\mathbf{A}_c$ and $\mathbf{A}_r$, if we want a reconstruction $\mathbf{X}_{\text{recon}}$ then intuitively we may do something like

$$\mathbf{X}_{\text{recon}}^{(\text{naive})} = \mathbf{A}_c^{-1} \mathbf{B} \mathbf{A}_r^{-T} = \mathbf{X}_{\text{true}} + \underbrace{\mathbf{A}_c^{-1} \mathbf{E} \mathbf{A}_r^{-T}}_{\text{Inverted Noise}}, \text{ where } \mathbf{A}_r^{-T} \triangleq (\mathbf{A}_r^{-1})^T = (\mathbf{A}_r^T)^{-1}. \quad (3.5)$$

The Inverted Noise demonstrates the contribution form from the noise to the naïve reconstruction. Then as indicated by [4], the relative error in the reconstruction (3.5) is upper bounded by

$$\frac{\|\mathbf{X}_{\text{recon}}^{(\text{naive})} - \mathbf{X}_{\text{true}}\|_F}{\|\mathbf{X}_{\text{true}}\|_F} \leq \text{cond}(\mathbf{A}_c) \text{cond}(\mathbf{A}_r) \frac{\|\mathbf{E}\|_F}{\|\mathbf{B}\|_F}, \text{ where } \begin{cases} \|\cdot\|_F \text{ denotes the Frobenius norm of a matrix, and} \\ \text{cond}(\cdot) \text{ denotes the condition number.} \end{cases}$$

(3.6)

So, if we want some quality guarantees on that reconstruction of (3.5), e.g., the relative error

$$\frac{\|\mathbf{X}_{\text{recon}}^{(\text{naive})} - \mathbf{X}_{\text{true}}\|_F}{\|\mathbf{X}_{\text{true}}\|_F} \leq \varepsilon, \text{ where } \varepsilon \text{ can be a small percentage } (e.g., 2\%) \quad (3.7)$$

then a sufficient condition for the establishment of (3.7) is

$$\text{cond}(\mathbf{A}_c) \text{cond}(\mathbf{A}_r) \frac{\|\mathbf{E}\|_F}{\|\mathbf{B}\|_F} \leq \varepsilon \quad \Rightarrow \quad \|\mathbf{E}\|_F \leq \varepsilon \frac{\|\mathbf{B}\|_F}{\text{cond}(\mathbf{A}_c) \text{cond}(\mathbf{A}_r)}. \quad (3.8)$$

The ill-conditioned nature of the non-blind deblurring problem lies in the fact that, the condition numbers of blurring matrices (i.e., cond($\mathbf{A}_c$) and cond($\mathbf{A}_r$) here) are very large. So, a great denominator in (3.8) significantly "squeezes" the upper bound on $\|\mathbf{E}\|_F$, and such a small upper bound is unachievable by almost any noise in the world. Practical $\|\mathbf{E}\|_F$ exceeds that bound by a large amount, and consequently the naïve reconstruction (3.5) exhibits a noise-like appearance as shown in Figure 4 (B).

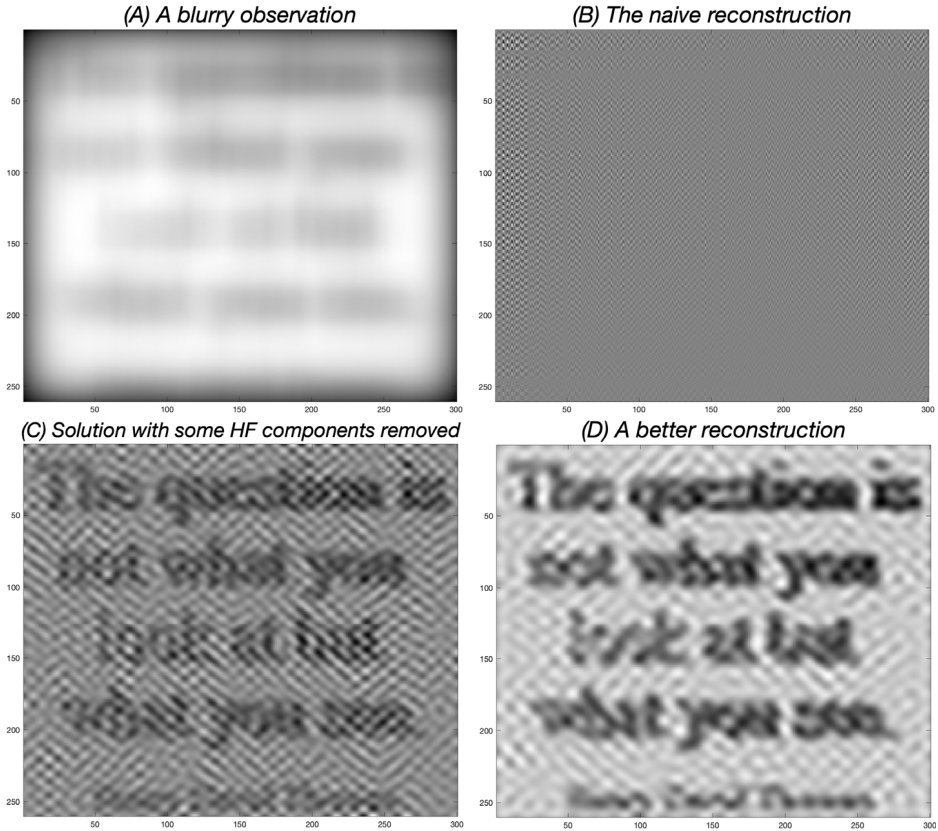

*Figure 4. Several reconstructions from a blurry observation (A) where the underlying text is "The question is not what you look at but what you see"* [43]. *The naïve reconstruction (B) is so noisy that we cannot read even a single word from it. Reconstructions get better after we remove some high-frequency components, as (C) and (D) tell.*

We have seen that the Inverted Noise term in (3.5) introduces great turbulence to the unfiltered solution, because of the large condition numbers of $\mathbf{A}_c$ and $\mathbf{A}_r$. Spectral Regularization methods works to mitigate this impact, as we shall see in the next few sections.

### 3.1.2 First attempt on deblurring

In the last section we demonstrated that image deblurring is an ill condition problem where the inverted noise exerts a great influence on the naïve reconstruction. Here let us investigate it further by spectral analysis. Back to (3.1), if we do a Singular Value Decomposition (SVD) on the forward blurring operator $\mathbf{A}$, *i.e.*,

$$\mathbf{A} = \mathbf{U}\mathbf{\Sigma}\mathbf{V}^\mathrm{T}, \text{ where } \begin{cases} \text{the forward operator } \mathbf{A} \in \mathbb{R}^{N \times N}, \\ \text{the orthogonal matrices } \mathbf{U}, \mathbf{V} \in \mathbb{R}^{N \times N} \text{ which satisfy} \begin{cases} \mathbf{U}^\mathrm{T}\mathbf{U} = \mathbf{U}\mathbf{U}^\mathrm{T} = \mathbf{I}_N \\ \mathbf{V}^\mathrm{T}\mathbf{V} = \mathbf{V}\mathbf{V}^\mathrm{T} = \mathbf{I}_N \end{cases}, \text{ and} \\ \text{the diagonal matrix } \mathbf{\Sigma} = \mathrm{diag}(\sigma_1, \sigma_2, \ldots, \sigma_N) \in \mathbb{R}^{N \times N} \text{ where singular values } \sigma_1 \geq \sigma_2 \geq \ldots \geq \sigma_N \geq 0 \end{cases}$$

(3.9)

then

$$\mathbf{A}^{-1} = \left(\mathbf{U}\mathbf{\Sigma}\mathbf{V}^\mathrm{T}\right)^{-1} = \mathbf{V}\mathbf{\Sigma}^{-1}\mathbf{U}^\mathrm{T} = \left(\mathbf{v}_1, \mathbf{v}_2, \ldots, \mathbf{v}_N\right) \begin{pmatrix} \sigma_1^{-1} & 0 & \cdots & 0 \\ 0 & \sigma_2^{-1} & \cdots & 0 \\ \vdots & \vdots & \ddots & \vdots \\ 0 & 0 & \cdots & \sigma_N^{-1} \end{pmatrix} \begin{pmatrix} \mathbf{u}_1^\mathrm{T} \\ \mathbf{u}_2^\mathrm{T} \\ \vdots \\ \mathbf{u}_N^\mathrm{T} \end{pmatrix}$$

$$= \left(\sigma_1^{-1}\mathbf{v}_1, \sigma_2^{-1}\mathbf{v}_2, \ldots, \sigma_N^{-1}\mathbf{v}_N\right) \begin{pmatrix} \mathbf{u}_1^\mathrm{T} \\ \mathbf{u}_2^\mathrm{T} \\ \vdots \\ \mathbf{u}_N^\mathrm{T} \end{pmatrix}$$

$$= \sigma_1^{-1}\mathbf{v}_1\mathbf{u}_1^\mathrm{T} + \sigma_2^{-1}\mathbf{v}_2\mathbf{u}_2^\mathrm{T} + \ldots + \sigma_N^{-1}\mathbf{v}_N\mathbf{u}_N^\mathrm{T}$$

$$= \sum_{i=1}^{N} \sigma_i^{-1}\mathbf{v}_i\mathbf{u}_i^\mathrm{T}, \text{ where } \mathbf{u}_i, \mathbf{v}_i \in \mathbb{R}^{N \times 1}.$$

(3.10)

So, the vector-form naïve reconstruction, is

$$\operatorname{vec}\left(\mathbf{X}_{\mathrm{recon}}^{(\mathrm{naive})}\right) = \mathbf{A}^{-1}\operatorname{vec}(\mathbf{B}) = \mathbf{A}^{-1}\left(\mathbf{A}\operatorname{vec}(\mathbf{X}_{\mathrm{true}}) + \mathbf{e}\right) = \operatorname{vec}(\mathbf{X}_{\mathrm{true}}) + \underbrace{\mathbf{A}^{-1}\mathbf{e}}_{\text{Inverted Noise}}$$

$$= \operatorname{vec}(\mathbf{X}_{\mathrm{true}}) + \left(\sum_{i=1}^{N} \sigma_i^{-1}\mathbf{v}_i\mathbf{u}_i^\mathrm{T}\right)\mathbf{e}$$

$$= \operatorname{vec}(\mathbf{X}_{\mathrm{true}}) + \sum_{i=1}^{N} \sigma_i^{-1}\mathbf{v}_i \langle \mathbf{u}_i, \mathbf{e}\rangle$$

$$= \operatorname{vec}(\mathbf{X}_{\mathrm{true}}) + \underbrace{\sum_{i=1}^{N} \frac{\langle \mathbf{u}_i, \mathbf{e}\rangle}{\sigma_i}\mathbf{v}_i}_{\text{Inverted Noise}}$$

$$= \sum_{i=1}^{N} \frac{\langle \mathbf{u}_i, \operatorname{vec}(\mathbf{B})\rangle}{\sigma_i}\mathbf{v}_i.$$

(3.11)

As (3.11) suggests, the naïve reconstruction is a weighted summation of ALL the singular vectors $\mathbf{v}_i$ of the forward operator A. The weight of $\mathbf{v}_i$ is the projection of the blurry observation vec(**B**) onto the singular vector $\mathbf{u}_i$, and then divided by $\sigma_i$ which is the $i^\text{th}$ singular value of **A**. Singular vectors have their spectral properties, *i.e.*, high-order singular vectors correspond to high-frequency components. We can observe that from a series of eigen pictures, as shown in Figure 5 (D)~(I), which are rearranged versions of the singular vectors.

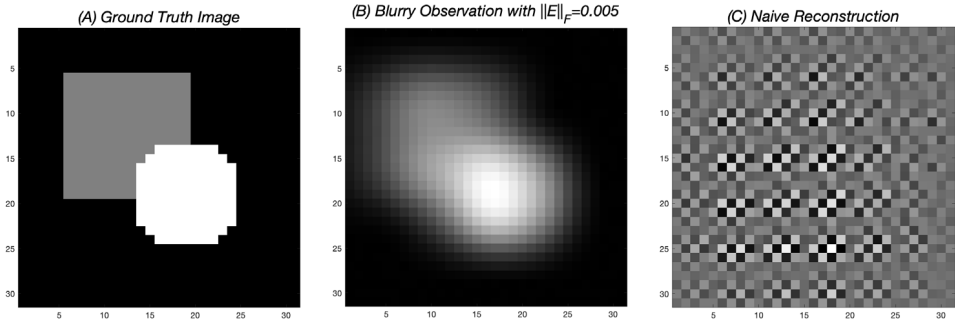

(A) Ground Truth Image  (B) Blurry Observation with $\|E\|_F = 0.005$  (C) Naive Reconstruction

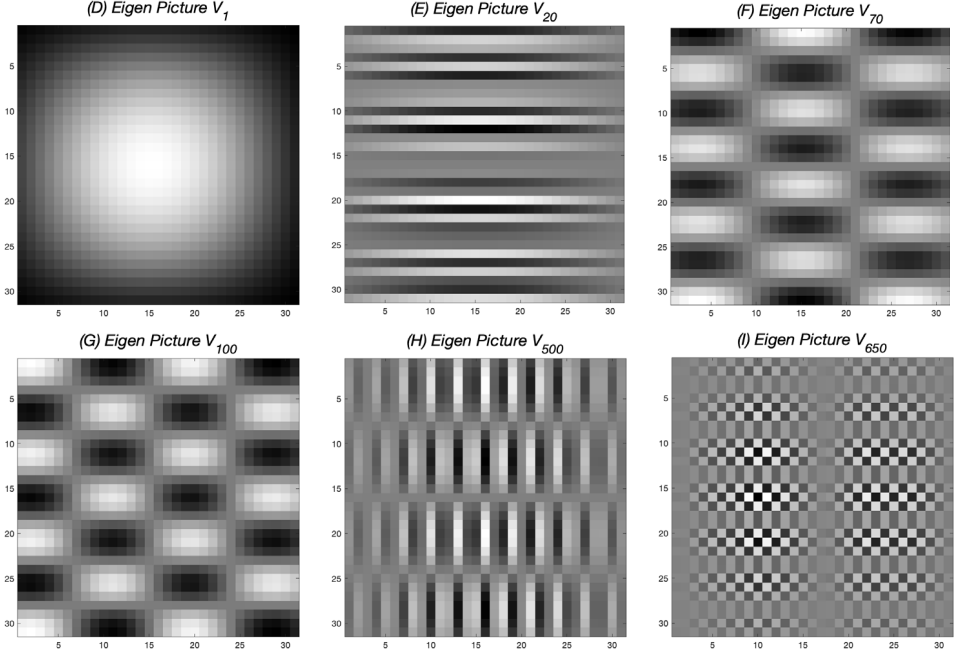

***Figure 5.*** *Eigen pictures of an image deblurring example* [43]. *The higher the order of the eigen picture, the more it corresponds to high-frequency information that changes drastically.*

The matrix-form of naïve reconstruction is

$$\mathbf{X}_{\text{recon}}^{(\text{naïve})} = \mathbf{X}_{\text{true}} + \underbrace{\sum_{i=1}^{N} \frac{\langle \mathbf{u}_i, \mathbf{e} \rangle}{\sigma_i} \mathbf{V}_i}_{\text{Inverted Noise}}, \text{ where } \mathbf{V}_i \text{ is the } i^{\text{th}} \text{ eigen picture.} \quad (3.12)$$

Figure 5 collectively displays the terms we have discussed till now. Given a blurry observation and the forward operator, we can have a naïve reconstruction like (C). From (3.12) we know that it is a weighted summation of eigen pictures like the ones displayed from (D) to (I). Due to the large condition number of A, high-order singular values are significantly lower than low-order singular values. From [23] we learn that in (3.12) |<$\mathbf{u}_i$, $\mathbf{e}$>| are roughly at the same magnitude for all *i*. So, the larger the *i* is, the larger is the weight before the $i^{\text{th}}$ eigen picture in (3.12).

Images of good perceptual quality, like the one in Figure 5 (A), are mainly low-frequency stuffs. However, the naïve reconstruction is now dominated by high-order eigen pictures and thus goes away from the underlying truth. High-order eigen pictures, as we see from (G) to (I), correspond to high-frequency information. So, meaningful context information gets overwhelmed and the naïve reconstruction in (C) puts on a quite noisy appearance just like high-order eigen pictures, *e.g.*, (I).

Therefore, an effective attempt is to "filter out" high-frequency components by an appropriate amount. Rather than a weighted summation of ALL the singular vectors as (3.12) tells, now we adopt a filtered reconstruction

$$\mathbf{X}_{\text{recon}}^{(\text{filtered})} = \sum_{i=1}^{k} \frac{\langle \mathbf{u}_i, \text{vec}(\mathbf{B}) \rangle}{\sigma_i} \mathbf{V}_i, \text{ where } k < N \quad (3.13)$$

and we test performances on different *k*. We demonstrate, in Figure 6, a series of experimental results on the blurry observation shown before in Figure 4 (A):

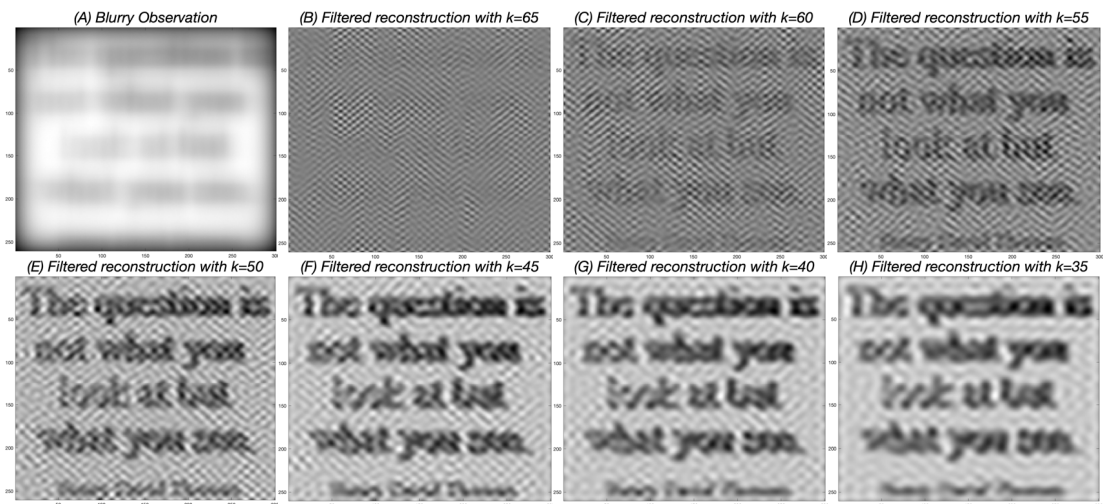

***Figure 6.*** *A series of deblurring results by different k in (3.13).*

We see from Figure 6 that with a high *k*, *e.g.*, *k*=65, the reconstruction is still dominated by high-frequency interruptions and the underlying text is hard to read. As the value of *k* decreases, more and more high-frequency parts are removed, and the text gradually becomes clear. However, a too-small *k* will also cause meaningful contexts of the image to be removed along with high-frequency components, as indicated by (G) and (H). So, an appropriate *k* in (3.13) should not only remove high-frequency components that cause interference, but also retain the low-frequency parts that make up the image information.

We tested the effectiveness of (3.13) through a series of experiments in Figure 6. However, we need to investigate it in a more theoretical way. The followings are a few aspects to consider:
- Are there any conditions for (3.13) to work?
- Apart from choosing *k* based on observations, are there any methods to find its optimum value?
- We know from (3.9) that **A** is a huge matrix, and (3.13) is based on the Singular Value Decomposition. So, can we do SVD on **A** directly?

We will explore these things in the next few sections.

### 3.1.3 Fast Algorithm for Spectral Decomposition

The forward blurring operator **A** is a square matrix with side length being the product of the width and length of the image to be handled. For a small image of 260 × 300, the forward operator **A** is of (260×300) × (260×300). It is unlikely to do SVD directly on an array of such a large dimension, at least for common PCs. A good news is that the forward operator **A** is highly structured if some (simplified) boundary conditions are met. It helps us to calculate the singular values of **A** through manipulations on a much smaller PSF instead of **A**.

As mentioned at the beginning of Section 3.1.1, computers produce a blurry image through convolving the image with a small kernel called PSF. We described the 2D convolution process in Section 2.2. And to allow pixels at the edge of the image to have the same chance of participating in convolution as central pixels, we pad the image before convolving with the PSF. Images are padded based on different boundary conditions, of which Zero Boundary Condition, Periodic Boundary Condition, and Reflexive Boundary Condition are three common examples, as shown in Figure 7 (B)~(D).

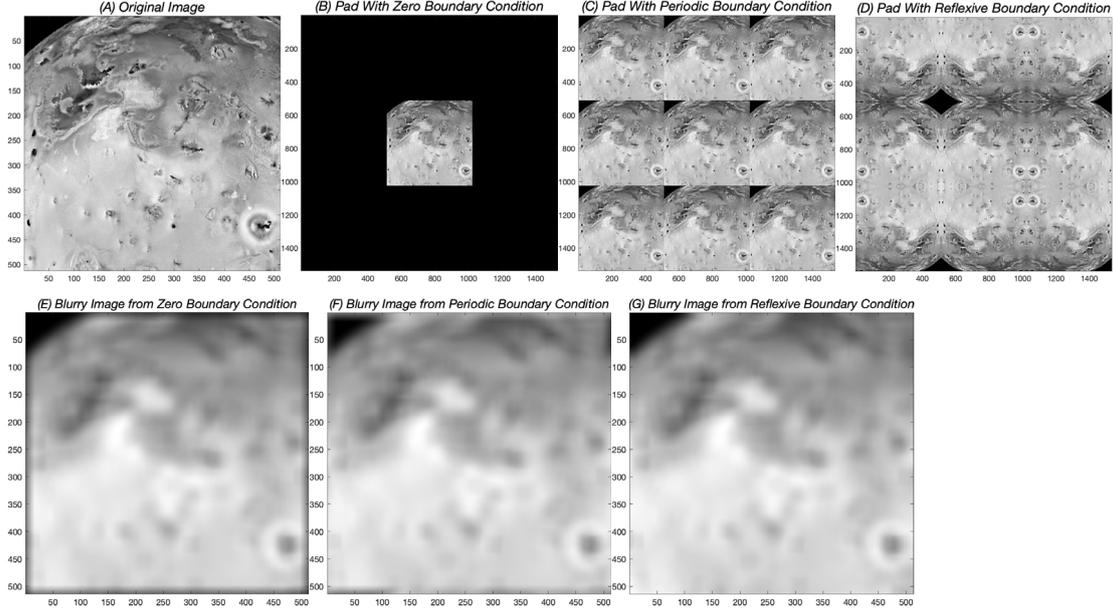

*Figure 7. (A)~(D) are the original image [43] and three padded versions based on different boundary conditions. I~(G) show the three blurry images obtained by convolving the same Gaussian kernel with (B)~(D), respectively.*

The Reflexive Boundary Condition is the most popular and most used among the three. The "mirror image relationship" ensures the smooth transition of pixel values to the boundary. Therefore, the convolution window moving on the boundary will not contain drastically changing or irrelevant information. The resultant blurry image appears much real as (G) suggests. Comparatively, blurry images obtained from Zero or Periodic Boundary Conditions have more or less "black artifacts" around the edges, as (I) and (F) indicate.

Based on these boundary condition assumptions, spectral decomposition of **A** can be efficiently computed through Fast Flourier Transformation (FFT) or Discrete Cosine Transformation (DCT) of the PSF. For detailed derivations and MATLAB functions please have a look at [24]. The following table [25] summarizes the cases when such efficient algorithms are applicable.

*Table 1. Cases when efficient algorithms are applicable*

| PSF | Boundary Condition | Forward Operator A | Fast Algorithm |
| --- | --- | --- | --- |
| Arbitrary | Periodic | *BCCB | 2D FFT |
| Doubly Symmetric | Reflexive | *BTTB+*BTHB+*BHTB+*BHHB | 2D DCT |
| Separable | Arbitrary | Kronecker product of $\mathbf{A}_r \otimes \mathbf{A}_c$ | Two small SVDs |

*BCCB[24]: *Block circulant with circulant blocks*; *BTTB[24]: *Block Toeplitz with Toeplitz blocks*;
*BTHB[24]: *Block Toeplitz with Hankel blocks*; *BHTB[24]: *Block Hankel with Toeplitz blocks*;
*BHHB[24]: *Block Hankel with Hankel blocks.*

As shown in the table, we can use fast algorithms to decompose **A** onto different bases and do image reconstructions, provided that we are in one of the three conditions listed above. We have illustrated the concept of Boundary Condition in Figure 7. For the "PSF" column in the table, a Double-Symmetric PSF is the one whose non-zero part is symmetrical in both horizontal and vertical directions [24], and a Separable PSF is for the case of separable blurs as we mentioned in Section 3.1.1.

### 3.1.4 View Spectral Regularization from the Spectral Filtering perspective

In this section let us view the Spectral Regularization Methods from the perspective of Spectral Filtering. Following the notation of (3.13), spectral filtering produces a reconstruction by the following generic form

$$\mathbf{X}_{\text{recon}}^{(\text{filtered})} = \sum_{i=1}^{N} \phi_i \frac{\langle \mathbf{u}_i, \text{vec}(\mathbf{B}) \rangle}{\sigma_i} \mathbf{V}_i \triangleq \sum_{i=1}^{N} \phi_i \frac{\langle \mathbf{u}_i, \mathbf{b} \rangle}{\sigma_i} \mathbf{V}_i \qquad (3.14\text{-}1)$$

where

$$\begin{cases} \phi_i \text{ is the } i^{\text{th}} \text{ filter factor}, \\ \langle \mathbf{u}_i, \mathbf{b} \rangle \text{ is the } i^{\text{th}} \text{ Spectral Coefficient of observation } \mathbf{B}, \\ \sigma_i \text{ is the } i^{\text{th}} \text{ Singular Value of the forward operator } \mathbf{A}, \text{ and} \\ \mathbf{V}_i \text{ is the } i^{\text{th}} \text{ eigen picture}. \end{cases} \qquad (3.14\text{-}2)$$

The filtered reconstruction is the weighted summation of all the eigen pictures but regularized by a series of filter factors $\phi_i$. We denote the inner product $|\langle \mathbf{u}_i, \text{vec}(\mathbf{B})\rangle|$ by the name – "Spectral Coefficient" of the observation **B** – in the sense that we are now in the coordinate system of

$$\mathbf{U}^{\text{T}}\mathbf{b} = \left(\mathbf{u}_1, \mathbf{u}_2, \ldots, \mathbf{u}_i, \ldots, \mathbf{u}_N\right)^{\text{T}} \mathbf{b} = \begin{pmatrix} \mathbf{u}_1^{\text{T}} \\ \mathbf{u}_2^{\text{T}} \\ \vdots \\ \mathbf{u}_i^{\text{T}} \\ \vdots \\ \mathbf{u}_N^{\text{T}} \end{pmatrix} \mathbf{b} = \begin{pmatrix} \mathbf{u}_1^{\text{T}}\mathbf{b} \\ \mathbf{u}_2^{\text{T}}\mathbf{b} \\ \vdots \\ \mathbf{u}_i^{\text{T}}\mathbf{b} \\ \vdots \\ \mathbf{u}_N^{\text{T}}\mathbf{b} \end{pmatrix}, \text{ where } \begin{cases} \mathbf{u}_i, \mathbf{b} \in \mathbb{R}^{N\times 1}, \text{ and} \\ \mathbf{u}_i^{\text{T}}\mathbf{b} \in \mathbb{R}^{1\times 1} \text{ is the } i^{\text{th}} \text{ component} \end{cases}$$

(3.14-3)

where the orthogonal matrix **U** linearly transposes **b**, the vector-form observation, into the frequency domain.

### 3.1.4.1 Analysis on the Spectral Coefficients and Singular Values

We have discussed the Eigen Picture $\mathbf{V}_i$ in Section 3.1.2. Now before we talk about the filter factor $\phi_i$, it is interesting to explore the behaviors of the Spectral Coefficient as well as the Singular Values, as shown in Figure 8 below.

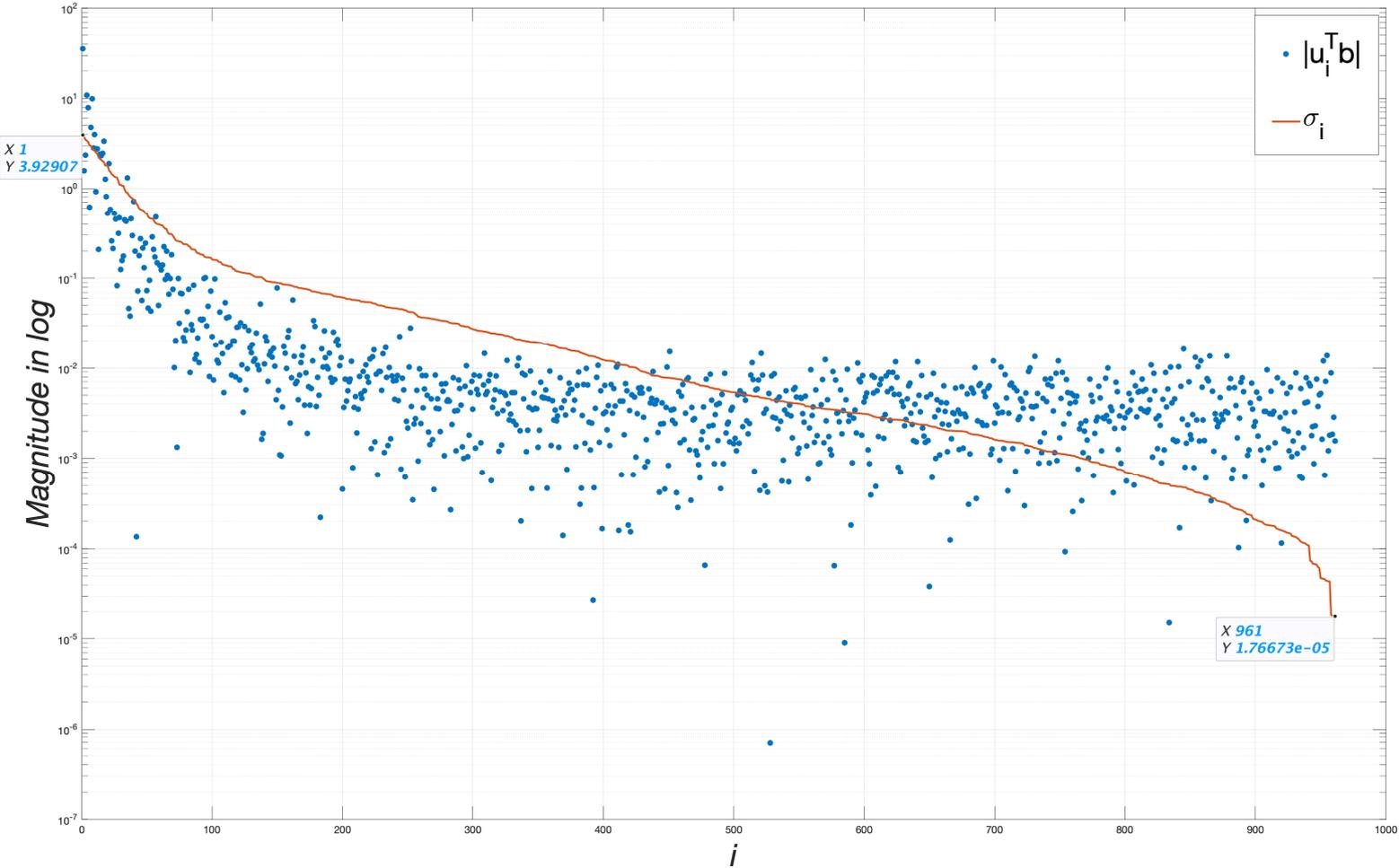

**Figure 8.** *Investigation on the Spectral Coefficients and the Singular Values. Operations are done towards the 31×31 ground-truth image and the blurry observation in Figure 5 (A) and (B).*

Here a higher $i$ corresponds to the Eigen Picture $\mathbf{V}_i$ of a higher frequency in (3.14-1). It is observed that the magnitude of Spectral Coefficient and the Singular Value both decrease against the index $i$ but with different behaviors. The Singular Value decreases from $\sigma_1$=3.92907 to $\sigma_{961}$=1.76673×10$^{-5}$, which validates a large condition number of $\mathbf{A}$ of $\sigma_1/\sigma_{961}$≈222392. As shown in the plot the red line decreases steadily in logarithm, which means a quite rapid fall on the linear scale. The decrease of Spectral Coefficient acts differently. For the first 100 values it goes down more rapidly than Singular Value does. Nevertheless, the decreasing speed gradually vanishes after that and consequently, the Spectral Coefficient levels off at a noise plateau after its 300$^{th}$ value.

The noisy behavior of the Spectral Coefficient is easily explainable. Let us assume the vector-form blurry observation **b** is the exact blurry **b**$_{exact}$ plus an additive noise term, *i.e.*,

$$\mathbf{b} = \mathbf{b}_{exact} + \mathbf{e}, \text{ where } \mathbf{b}_{exact} = \mathbf{A} \cdot \text{vec}(\mathbf{X}_{true}) = \mathbf{A}\mathbf{x}_{true}. \tag{3.15}$$

And the additive noise **e** has statistically independent components, with zero mean and identical variation $\eta^2$, *i.e.*,

$$\begin{cases} \varepsilon[\mathbf{e}] = \mathbf{0}, \text{ and} \\ \text{Cov}[\mathbf{e}] = \varepsilon\left[(\mathbf{e}-\varepsilon[\mathbf{e}])(\mathbf{e}-\varepsilon[\mathbf{e}])^T\right] = \varepsilon[\mathbf{ee}^T] = \varepsilon\left[\begin{pmatrix}e_1\\e_2\\\vdots\\e_N\end{pmatrix}(e_1, e_2, \ldots, e_N)\right] = \begin{pmatrix}\eta^2 & 0 & \cdots & 0\\0 & \eta^2 & \cdots & 0\\\vdots & \vdots & \ddots & \vdots\\0 & 0 & \cdots & \eta^2\end{pmatrix} = \eta^2 \mathbf{I}_N. \end{cases} \tag{3.16}$$

Then the expectation of the squared Spectral Coefficient $(\mathbf{u}_i^T\mathbf{b})^2$ is given by

$$\varepsilon\left[(\mathbf{u}_i^T\mathbf{b})^2\right] = \varepsilon\left[(\mathbf{u}_i^T\mathbf{b}_{exact} + \mathbf{u}_i^T\mathbf{e})^2\right] = \varepsilon\left[(\mathbf{u}_i^T\mathbf{b}_{exact})^2 + 2\mathbf{u}_i^T\mathbf{b}_{exact}\mathbf{u}_i^T\mathbf{e} + (\mathbf{u}_i^T\mathbf{e})^2\right]$$

$$= (\mathbf{u}_i^T\mathbf{b}_{exact})^2 + 2\mathbf{u}_i^T\mathbf{b}_{exact} \cdot \varepsilon[\mathbf{u}_i^T\mathbf{e}] + \varepsilon\left[(\mathbf{u}_i^T\mathbf{e})^2\right] \tag{3.17-1}$$

$$= (\mathbf{u}_i^T\mathbf{b}_{exact})^2 + \eta^2$$

where

$$\begin{cases} \varepsilon[\mathbf{u}_i^T\mathbf{e}] = \mathbf{u}_i^T \cdot \varepsilon[\mathbf{e}] = 0, \text{ and} \\ \varepsilon\left[(\mathbf{u}_i^T\mathbf{e})^2\right] = \varepsilon\left[(\mathbf{u}_i^T\mathbf{e})(\mathbf{u}_i^T\mathbf{e})^T\right] = \varepsilon[\mathbf{u}_i^T\mathbf{ee}^T\mathbf{u}_i] = \mathbf{u}_i^T\varepsilon[\mathbf{ee}^T]\mathbf{u}_i = \mathbf{u}_i^T \cdot \eta^2\mathbf{I}_N \cdot \mathbf{u}_i = \eta^2\mathbf{u}_i^T\mathbf{I}_N\mathbf{u}_i = \eta^2. \end{cases}$$

$$\tag{3.17-2}$$

The magnitude expectation of the Spectral Coefficient is therefore

$$\varepsilon\left[|\mathbf{u}_i^T\mathbf{b}|\right] = \sqrt{\varepsilon\left[(\mathbf{u}_i^T\mathbf{b})^2\right]} = \sqrt{(\mathbf{u}_i^T\mathbf{b}_{exact})^2 + \eta^2} \approx \begin{cases} |\mathbf{u}_i^T\mathbf{b}_{exact}|, & \text{when } |\mathbf{u}_i^T\mathbf{b}_{exact}| \gg \eta \\ \eta, & \text{when } |\mathbf{u}_i^T\mathbf{b}_{exact}| \ll \eta. \end{cases} \tag{3.18}$$

So, only a few large Spectral Coefficients survive, and the rest small ones are overwhelmed by the noise plateau. The survived ones carry clear and effective information of the obervation, but the rest are meaningless. And according to (3.18), the greater the intensity of the noise, the fewer Spectral Coefficients can survive, and the less information can be used for reconstruction, as illustrated in the left column of Figure 9.

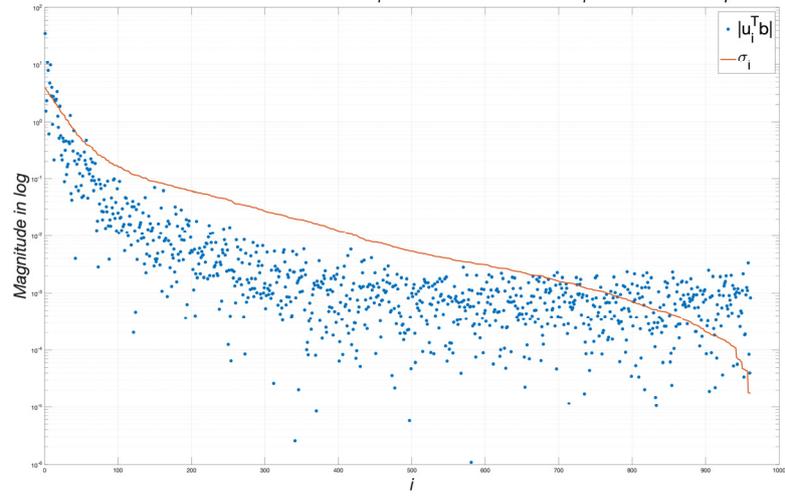
(A1) Variation of the $i^{th}$ Spectral Coefficient $|u_i^T b|$ and the singular value $\sigma_i$ against $i$, with $\|E\|_F=0.001$

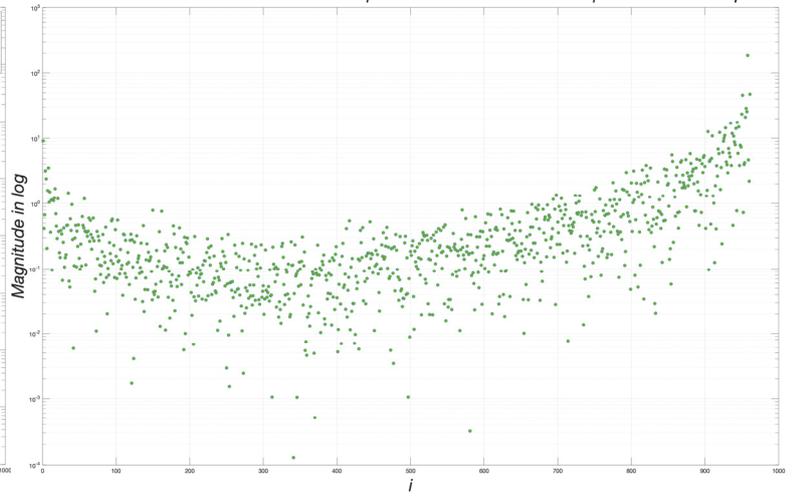
(A2) Variation of the $i^{th}$ Spectral Coefficient $|u_i^T b|$ divided by singular value $\sigma_i$ against $i$, with $\|E\|_F=0.001$

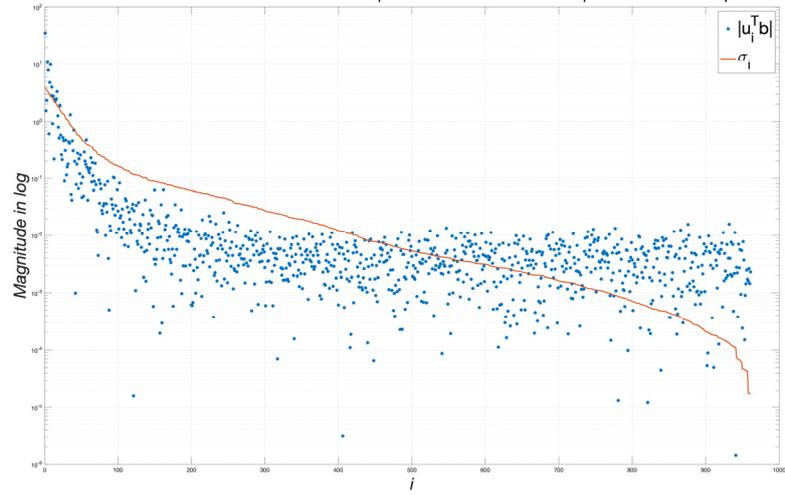
(B1) Variation of the $i^{th}$ Spectral Coefficient $|u_i^T b|$ and the singular value $\sigma_i$ against $i$, with $\|E\|_F=0.005$

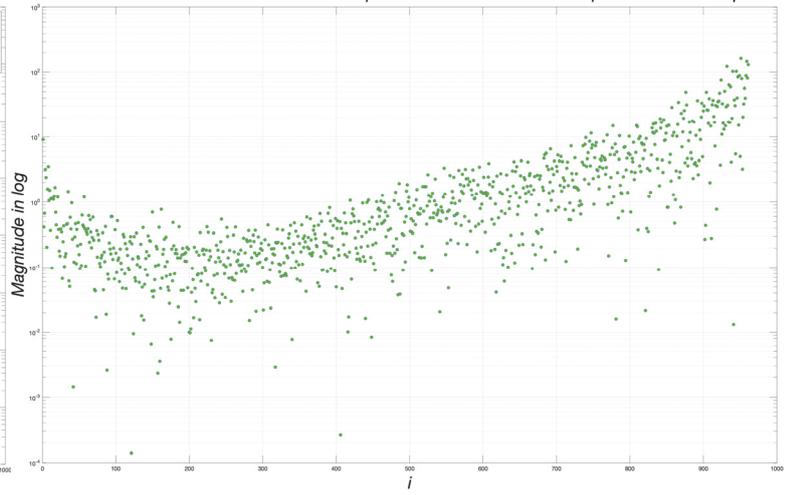
(B2) Variation of the $i^{th}$ Spectral Coefficient $|u_i^T b|$ divided by singular value $\sigma_i$ against $i$, with $\|E\|_F=0.005$

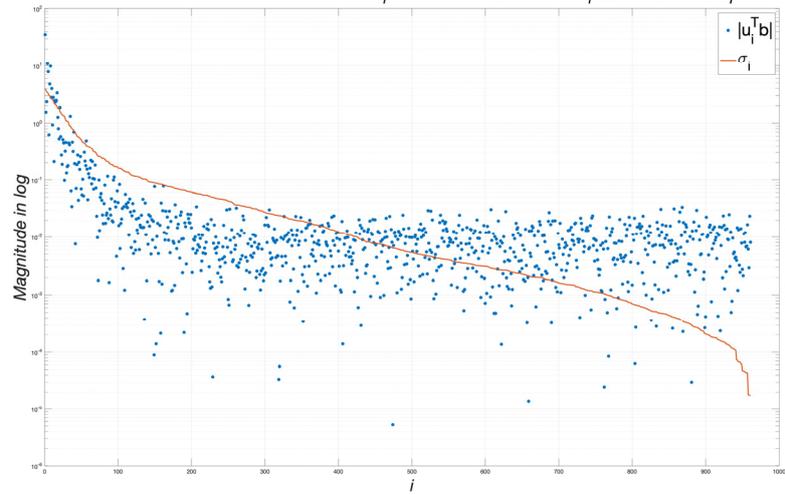
(C1) Variation of the $i^{th}$ Spectral Coefficient $|u_i^T b|$ and the singular value $\sigma_i$ against $i$, with $\|E\|_F=0.01$

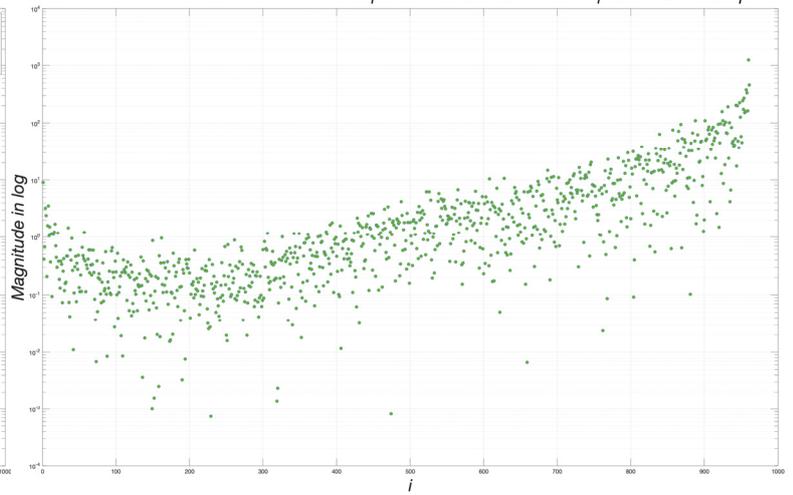
(C2) Variation of the $i^{th}$ Spectral Coefficient $|u_i^T b|$ divided by singular value $\sigma_i$ against $i$, with $\|E\|_F=0.01$

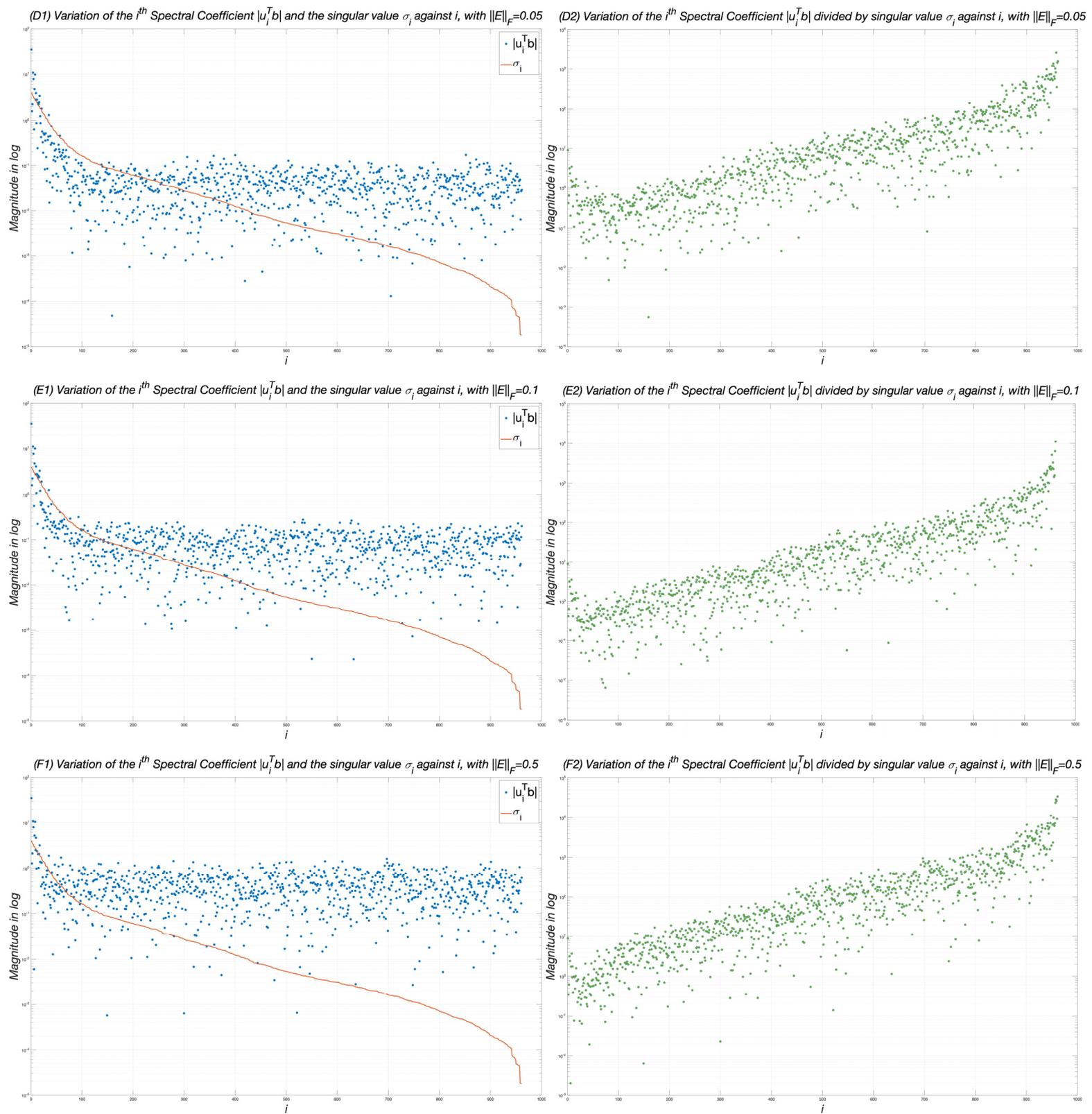

*Figure 9.* The greater the noise, the less information from data can be used for reconstruction.

### 3.1.4.2 The Discrete Picard Condition

Now, it is a good time to introduce the ***Discrete Picard Condition*** for such filtering methods to work. From the discussion of (3.18) we learned that the effective information we can get from <$\mathbf{u}_i$, $\mathbf{b}$> is only <$\mathbf{u}_i$, $\mathbf{b}_{exact}$>, *i.e.*, a few Spectral Coefficients at the early stage of the decay which satisfy

$$\varepsilon\left[\left|\mathbf{u}_i^T \mathbf{b}\right|\right] \approx \left|\mathbf{u}_i^T \mathbf{b}_{exact}\right|, \quad \text{when } \left|\mathbf{u}_i^T \mathbf{b}_{exact}\right| \gg \eta. \tag{3.19}$$

Now, if we go back to the linear blurring model

$$\mathbf{b}_{exact} = \mathbf{A}\mathbf{x}_{true} = \mathbf{U}\mathbf{\Sigma}\mathbf{V}^T\mathbf{x}_{true}, \text{ where } \begin{cases} \mathbf{U} = (\mathbf{u}_1, \mathbf{u}_2, \ldots, \mathbf{u}_N) \text{ with } \mathbf{u}_i \in \mathbb{R}^{N \times 1}, \\ \mathbf{V} = (\mathbf{v}_1, \mathbf{v}_2, \ldots, \mathbf{v}_N) \text{ with } \mathbf{v}_i \in \mathbb{R}^{N \times 1}, \text{ and} \\ \mathbf{\Sigma} = \text{diag}(\sigma_1, \sigma_2, \ldots, \sigma_N) \in \mathbb{R}^{N \times N} \end{cases} \tag{3.20}$$

then the Spectral Coefficients of $\mathbf{b}_{exact}$ are collectively stored in

$$\mathbf{U}^T\mathbf{b}_{exact} = \mathbf{U}^T \cdot \mathbf{U}\mathbf{\Sigma}\mathbf{V}^T\mathbf{x}_{true} = \mathbf{\Sigma}\mathbf{V}^T\mathbf{x}_{true} \tag{3.21-1}$$

where

$$\begin{cases} \mathbf{U}^T\mathbf{b}_{exact} = \begin{pmatrix} \mathbf{u}_1^T \\ \mathbf{u}_2^T \\ \vdots \\ \mathbf{u}_i^T \\ \vdots \\ \mathbf{u}_N^T \end{pmatrix} \mathbf{b}_{exact} = \begin{pmatrix} \mathbf{u}_1^T\mathbf{b}_{exact} \\ \mathbf{u}_2^T\mathbf{b}_{exact} \\ \vdots \\ \mathbf{u}_i^T\mathbf{b}_{exact} \\ \vdots \\ \mathbf{u}_N^T\mathbf{b}_{exact} \end{pmatrix}, \\ \mathbf{\Sigma}\mathbf{V}^T\mathbf{x}_{true} = \begin{pmatrix} \sigma_1 & 0 & \cdots & 0 & \cdots & 0 \\ 0 & \sigma_2 & \cdots & 0 & \cdots & 0 \\ \vdots & \vdots & \ddots & \vdots & \ddots & \vdots \\ 0 & 0 & \cdots & \sigma_i & \cdots & 0 \\ \vdots & \vdots & \ddots & \vdots & \ddots & \vdots \\ 0 & 0 & \cdots & 0 & \cdots & \sigma_N \end{pmatrix} \begin{pmatrix} \mathbf{v}_1^T \\ \mathbf{v}_2^T \\ \vdots \\ \mathbf{v}_i^T \\ \vdots \\ \mathbf{v}_N^T \end{pmatrix} \mathbf{x}_{true} = \begin{pmatrix} \sigma_1 & 0 & \cdots & 0 & \cdots & 0 \\ 0 & \sigma_2 & \cdots & 0 & \cdots & 0 \\ \vdots & \vdots & \ddots & \vdots & \ddots & \vdots \\ 0 & 0 & \cdots & \sigma_i & \cdots & 0 \\ \vdots & \vdots & \ddots & \vdots & \ddots & \vdots \\ 0 & 0 & \cdots & 0 & \cdots & \sigma_N \end{pmatrix} \begin{pmatrix} \mathbf{v}_1^T\mathbf{x}_{true} \\ \mathbf{v}_2^T\mathbf{x}_{true} \\ \vdots \\ \mathbf{v}_i^T\mathbf{x}_{true} \\ \vdots \\ \mathbf{v}_N^T\mathbf{x}_{true} \end{pmatrix} = \begin{pmatrix} \sigma_1\mathbf{v}_1^T\mathbf{x}_{true} \\ \sigma_2\mathbf{v}_2^T\mathbf{x}_{true} \\ \vdots \\ \sigma_i\mathbf{v}_i^T\mathbf{x}_{true} \\ \vdots \\ \sigma_N\mathbf{v}_N^T\mathbf{x}_{true} \end{pmatrix}. \end{cases}$$

(3.21-2)

So

$$\mathbf{U}^T\mathbf{b}_{exact} = \mathbf{\Sigma}\mathbf{V}^T\mathbf{x}_{true} \quad \Leftrightarrow \quad \underbrace{\mathbf{u}_i^T\mathbf{b}_{exact}}_{i^{\text{th}} \text{ Spectral Coefficient of } \mathbf{b}_{exact}} = \sigma_i \cdot \underbrace{\mathbf{v}_i^T\mathbf{x}_{true}}_{i^{\text{th}} \text{ Spectral Coefficient of } \mathbf{x}_{true}}. \tag{3.21-3}$$

We have observed the decay of $|\mathbf{u}_i^T\mathbf{b}_{exact}|$ – the first few values of $|\mathbf{u}_i^T\mathbf{b}|$ larger than the noise plateau – from the left column of Figure 9. And we should be aware that, the decay of $|\mathbf{u}_i^T\mathbf{b}_{exact}|$ is bound to be faster than the dacay of $|\mathbf{v}_i^T\mathbf{x}_{true}|$, because the blurry observation contains less high-frequency information than the original sharp image has. It means, if we want to approximate the ground-truth image from the blurry obervation, then, dividing the decaying $\mathbf{u}_i^T\mathbf{b}_{exact}$ by the decaying $\sigma_i$ should yield a decaying $\mathbf{v}_i^T\mathbf{x}_{true}$

as well. In other words,

***(The Discrete Picard Condition)*** *If we want a proper restoration from the blurry data, then the Spectral Coefficients of the observation **B** should decay faster than the singular values of the forward operator **A** do.*

The Discrete Picard Condition can be assessed if we observe the variation of $|\mathbf{u}_i^T\mathbf{b}|/\sigma_i$ against $i$ as shown in the right column of Figure 9. The condition is met if the value can go down before increasing. The amount of green dots going down indicates exactly how many useful data we can have, dependent on the intensity of the noise, as we discussed at the end of Section 3.1.4.1.

### 3.1.4.3 Two Spectral Filtering Methods

Back to the beginning of Section 3.1.4 we provided a generic form for the filtered reconstruction, and for convenience we rewrite it here as

$$\mathbf{X}_{\text{recon}}^{(\text{filtered})} = \sum_{i=1}^{N} \phi_i \frac{\langle \mathbf{u}_i, \mathbf{b} \rangle}{\sigma_i} \mathbf{V}_i, \text{ where } \begin{cases} \phi_i \text{ is the } i^{\text{th}} \text{ filter factor,} \\ \langle \mathbf{u}_i, \mathbf{b} \rangle \text{ is the } i^{\text{th}} \text{ spectral coefficient of the blurry data,} \\ \sigma_i \text{ is the } i^{\text{th}} \text{ singular value of the forward operator, and} \\ \mathbf{V}_i \text{ is the } i^{\text{th}} \text{ eigen picture.} \end{cases}$$

(3.22)

Spectral filtering methods can be divided into different categories based on different designs on $\phi_i$. Here we present two of them, namely the *Truncated Singular Value Decomposition* (TSVD) *Method* and the *Tikhonov Method.*

➢ *The TSVD Method*

For the TSVD Method the filter factors are designed by

$$\phi_i = \begin{cases} 1, & i = 1, 2, \ldots, k \\ 0, & i = k+1, \ldots, N \end{cases}, \text{ where } k \text{ is the truncation parameter.} \quad (3.23)$$

It is like an ideal low-pass filter, passing all elements before the $k^{\text{th}}$ without any modification while removing completely elements after that.

➢ *The Tikhonov Method*

For this method the filter factors are designed by

$$\phi_i = \frac{\sigma_i^2}{\sigma_i^2 + \alpha^2}, \text{ where } \begin{cases} \text{the regularization parameter } \alpha \in [\sigma_N, \sigma_1], \text{ and} \\ i = 1, 2, \ldots, N. \end{cases} \quad (3.24\text{-}1)$$

The choice of $\alpha$ – the regularization parameter – affects the shape of the filter. It is a pre-determined value chosen within the wide range of $\sigma_i$, the singular values of the forward operator. For the $i^{\text{th}}$ filter factor,

$$\phi_i = \begin{cases} \dfrac{\sigma_i^2}{\sigma_i^2+\alpha^2} = \dfrac{1}{1+\alpha^2/\sigma_i^2} = \left(1+\dfrac{\alpha^2}{\sigma_i^2}\right)^{-1} = 1-\dfrac{\alpha^2}{\sigma_i^2}+\dfrac{\alpha^4}{\sigma_i^4}+\ldots, & \text{if } \alpha \ll \sigma_i. \\[2ex] \dfrac{\sigma_i^2}{\sigma_i^2+\alpha^2} = \dfrac{\sigma_i^2}{\alpha^2}\cdot\dfrac{1}{1+\sigma_i^2/\alpha^2} = \dfrac{\sigma_i^2}{\alpha^2}\cdot\left(1+\dfrac{\sigma_i^2}{\alpha^2}\right)^{-1} = \dfrac{\sigma_i^2}{\alpha^2}\cdot\left(1-\dfrac{\sigma_i^2}{\alpha^2}+\dfrac{\sigma_i^4}{\alpha^4}+\ldots\right), & \text{if } \alpha \gg \sigma_i. \end{cases}$$

(3.24-2)

In other words,

$$\phi_i = \begin{cases} 1-\dfrac{\alpha^2}{\sigma_i^2}+o\!\left(\left(\dfrac{\alpha}{\sigma_i}\right)^{4}\right), & \text{for } \alpha \ll \sigma_i. \\[2ex] \dfrac{\sigma_i^2}{\alpha^2}+o\!\left(\left(\dfrac{\sigma_i}{\alpha}\right)^{4}\right), & \text{for } \alpha \gg \sigma_i. \end{cases}$$

(3.24-3)

For large $\sigma_i$ corresponding to small index $i$ and low-frequency information, $\phi_i$ is closer to 1 to pass those eigen pictures. $\Phi_i$ reduces as $\sigma_i$ decreases, and the degree of regularization is controlled by the regularization parameter $\alpha$. Figure 10 illustrates the role of the regularization parameter $\alpha$: the higher $\alpha$, the fewer $\phi_i$ closer to one, and it means a stricter regulation we adopted in the Tikhonov Method.

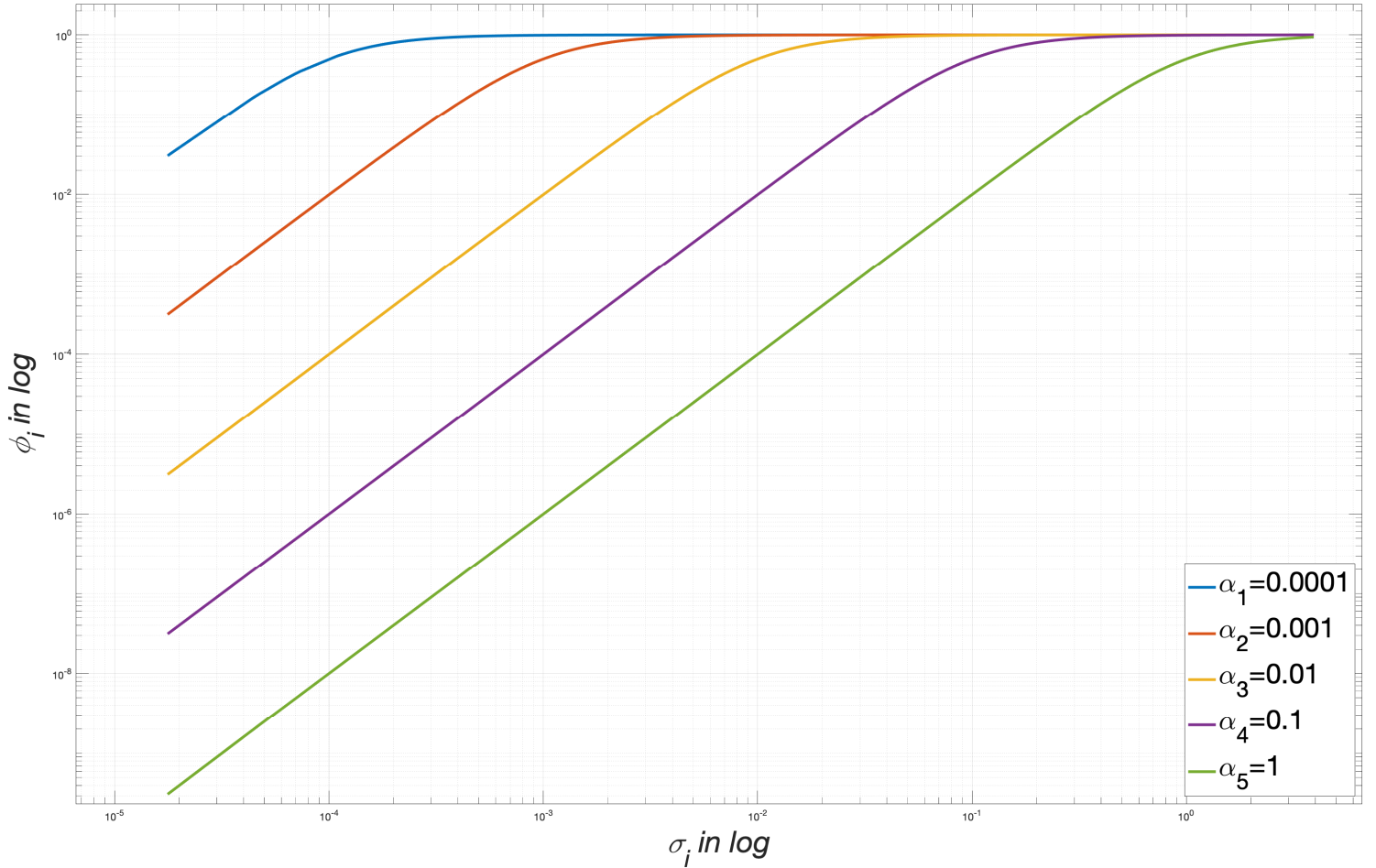

*Figure 10. Different sets of Tikhonov filter factors based on different regularization parameter $\alpha$. The higher $\alpha$ is, the fewer filter factors closer to one, and the stricter our regularization is.*

### 3.1.4.4 Validation of Spectral Filtering Methods

The effectiveness of these spectral filtering methods can be justified through an investigation into the ***Regularization Error*** and the ***Perturbation Error***. Recall (3.11), the vector-form naïve reconstruction is the inversed observation like

$$\mathbf{x}_{\text{recon}}^{(\text{naive})} = \mathbf{A}^{-1}\mathbf{b} = \left(\mathbf{U}\mathbf{\Sigma}\mathbf{V}^T\right)^{-1}\mathbf{b} = \mathbf{V}\mathbf{\Sigma}^{-1}\mathbf{U}^T\mathbf{b}, \quad \text{where} \begin{cases} \mathbf{U}\mathbf{U}^T = \mathbf{U}^T\mathbf{U} = \mathbf{I}_N \\ \mathbf{V}\mathbf{V}^T = \mathbf{V}^T\mathbf{V} = \mathbf{I}_N \end{cases}. \quad (3.25)$$

Now we filter in the frequency domain, the filtered reconstruction is

$$\mathbf{x}_{\text{recon}}^{(\text{filtered})} = \mathbf{V}\mathbf{\Phi}\mathbf{\Sigma}^{-1}\mathbf{U}^T\mathbf{b}, \quad \text{where} \begin{cases} \mathbf{\Phi} = \begin{pmatrix} \phi_1 & 0 & \cdots & 0 & \cdots & 0 \\ 0 & \phi_2 & \cdots & 0 & \cdots & 0 \\ \vdots & \vdots & \ddots & \vdots & \ddots & \vdots \\ 0 & 0 & \cdots & \phi_i & \cdots & 0 \\ \vdots & \vdots & \ddots & \vdots & \ddots & \vdots \\ 0 & 0 & \cdots & 0 & \cdots & \phi_N \end{pmatrix}, \text{ and} \\ \phi_i \text{ can be either the TSVD or the Tikhonov fator we discussed.} \end{cases}$$

(3.26-1)

Following the notation of (3.15) to split the observation into the exact blurred image and the additive noise, we have a filtered reconstruction of the form

$$\begin{aligned} \mathbf{x}_{\text{recon}}^{(\text{filtered})} &= \mathbf{V}\mathbf{\Phi}\mathbf{\Sigma}^{-1}\mathbf{U}^T\mathbf{b} = \mathbf{V}\mathbf{\Phi}\mathbf{\Sigma}^{-1}\mathbf{U}^T\left(\mathbf{b}_{\text{exact}} + \mathbf{e}\right) \\ &= \mathbf{V}\mathbf{\Phi}\mathbf{\Sigma}^{-1}\mathbf{U}^T\mathbf{b}_{\text{exact}} + \mathbf{V}\mathbf{\Phi}\mathbf{\Sigma}^{-1}\mathbf{U}^T\mathbf{e} \\ &= \mathbf{V}\mathbf{\Phi}\mathbf{\Sigma}^{-1}\mathbf{U}^T\mathbf{A}\mathbf{x}_{\text{true}} + \mathbf{V}\mathbf{\Phi}\mathbf{\Sigma}^{-1}\mathbf{U}^T\mathbf{e} \\ &= \mathbf{V}\mathbf{\Phi}\mathbf{\Sigma}^{-1}\mathbf{U}^T \cdot \left(\mathbf{U}\mathbf{\Sigma}\mathbf{V}^T\right)\mathbf{x}_{\text{true}} + \mathbf{V}\mathbf{\Phi}\mathbf{\Sigma}^{-1}\mathbf{U}^T\mathbf{e} \\ &= \mathbf{V}\mathbf{\Phi}\mathbf{V}^T\mathbf{x}_{\text{true}} + \mathbf{V}\mathbf{\Phi}\mathbf{\Sigma}^{-1}\mathbf{U}^T\mathbf{e}. \end{aligned} \quad (3.26\text{-}2)$$

It has a difference to the $\mathbf{x}_{\text{true}}$ by

$$\mathbf{x}_{\text{true}} - \mathbf{x}_{\text{recon}}^{(\text{filtered})} = \underbrace{\left(\mathbf{I}_N - \mathbf{V}\mathbf{\Phi}\mathbf{V}^T\right)\mathbf{x}_{\text{true}}}_{\text{Regularization Error}} - \underbrace{\mathbf{V}\mathbf{\Phi}\mathbf{\Sigma}^{-1}\mathbf{U}^T\mathbf{e}}_{\text{Perturbation Error}}. \quad (3.27)$$

The first error term is introduced by regularization, because if there is no filtering then $\mathbf{\Phi}$ is an identity matrix such that the Regularization Error is gone. The second error is the perturbation error to be reduced by regularization, because

$$\left\|\mathbf{V}\mathbf{\Phi}\mathbf{\Sigma}^{-1}\mathbf{U}^T\mathbf{e}\right\|_2^2 = \left\|\mathbf{V}^T\left(\mathbf{V}\mathbf{\Phi}\mathbf{\Sigma}^{-1}\mathbf{U}^T\mathbf{e}\right)\right\|_2^2 = \left\|\mathbf{\Phi}\mathbf{\Sigma}^{-1}\mathbf{U}^T\mathbf{e}\right\|_2^2 \quad (3.28)$$

$$= \left\|\begin{pmatrix} \phi_1 & 0 & \cdots & 0 \\ 0 & \phi_2 & \cdots & 0 \\ \vdots & \vdots & \ddots & \vdots \\ 0 & 0 & \cdots & \phi_N \end{pmatrix} \begin{pmatrix} \sigma_1^{-1} & 0 & \cdots & 0 \\ 0 & \sigma_2^{-1} & \cdots & 0 \\ \vdots & \vdots & \ddots & \vdots \\ 0 & 0 & \cdots & \sigma_N^{-1} \end{pmatrix} \begin{pmatrix} \mathbf{u}_1^T\mathbf{e} \\ \mathbf{u}_2^T\mathbf{e} \\ \vdots \\ \mathbf{u}_N^T\mathbf{e} \end{pmatrix}\right\|_2^2$$

$$= \left\|\begin{pmatrix} \phi_1\sigma_1^{-1}\mathbf{u}_1^T\mathbf{e} \\ \phi_2\sigma_2^{-1}\mathbf{u}_2^T\mathbf{e} \\ \vdots \\ \phi_N\sigma_N^{-1}\mathbf{u}_N^T\mathbf{e} \end{pmatrix}\right\|_2^2 = \sum_{i=1}^N \left(\phi_i\sigma_i^{-1}\mathbf{u}_i^T\mathbf{e}\right)^2$$

we can eliminate the magnification effect of large $\sigma_i^{-1}$ by setting $\phi_i$ to zero for that $\mathbf{u}_i^T \mathbf{e}$, which is small and roughly at the same order of magnitude for all $i$ according to [23].

The regularization-based methods work in the sense that they can strike a balance between Perturbation Error and the Regularization Error, by keeping these two simultaneously at a small level. We have seen how $\mathbf{\Phi}$ can reduce the Perturbation Error in (3.28), and now let us see how $\mathbf{\Phi}$ works to limit the Regularization Error caused by itself. The squared 2-norm of the Regularization Error is

$$\left\| \left( \mathbf{I}_N - \mathbf{V}\mathbf{\Phi}\mathbf{V}^T \right) \mathbf{x}_{true} \right\|_2^2 = \left\| \mathbf{V}^T \left( \mathbf{I}_N - \mathbf{V}\mathbf{\Phi}\mathbf{V}^T \right) \mathbf{x}_{true} \right\|_2^2 = \left\| \mathbf{V}^T \mathbf{x}_{true} - \mathbf{\Phi}\mathbf{V}^T \mathbf{x}_{true} \right\|_2^2 = \left\| \left( \mathbf{I}_N - \mathbf{\Phi} \right) \mathbf{V}^T \mathbf{x}_{true} \right\|_2^2$$

$$= \left\| \left( \mathbf{I}_N - \mathbf{\Phi} \right) \mathbf{V}^T \mathbf{A}^{-1} \mathbf{b}_{exact} \right\|_2^2 = \left\| \left( \mathbf{I}_N - \mathbf{\Phi} \right) \mathbf{V}^T \left( \mathbf{U}\mathbf{\Sigma}\mathbf{V}^T \right)^{-1} \mathbf{b}_{exact} \right\|_2^2$$

$$= \left\| \left( \mathbf{I}_N - \mathbf{\Phi} \right) \mathbf{V}^T \mathbf{V}\mathbf{\Sigma}^{-1}\mathbf{U}^T \mathbf{b}_{exact} \right\|_2^2 = \left\| \left( \mathbf{I}_N - \mathbf{\Phi} \right) \mathbf{\Sigma}^{-1}\mathbf{U}^T \mathbf{b}_{exact} \right\|_2^2$$

$$= \left\| \begin{pmatrix} 1-\phi_1 & 0 & \cdots & 0 \\ 0 & 1-\phi_2 & \cdots & 0 \\ \vdots & \vdots & \ddots & \vdots \\ 0 & 0 & \cdots & 1-\phi_N \end{pmatrix} \begin{pmatrix} \sigma_1^{-1} & 0 & \cdots & 0 \\ 0 & \sigma_2^{-1} & \cdots & 0 \\ \vdots & \vdots & \ddots & \vdots \\ 0 & 0 & \cdots & \sigma_N^{-1} \end{pmatrix} \begin{pmatrix} \mathbf{u}_1^T \\ \mathbf{u}_2^T \\ \vdots \\ \mathbf{u}_N^T \end{pmatrix} \mathbf{b}_{exact} \right\|_2^2$$

$$= \left\| \begin{pmatrix} (1-\phi_1)\sigma_1^{-1} & 0 & \cdots & 0 \\ 0 & (1-\phi_2)\sigma_2^{-1} & \cdots & 0 \\ \vdots & \vdots & \ddots & \vdots \\ 0 & 0 & \cdots & (1-\phi_N)\sigma_N^{-1} \end{pmatrix} \begin{pmatrix} \mathbf{u}_1^T \mathbf{b}_{exact} \\ \mathbf{u}_2^T \mathbf{b}_{exact} \\ \vdots \\ \mathbf{u}_N^T \mathbf{b}_{exact} \end{pmatrix} \right\|_2^2$$

$$= \left\| \begin{pmatrix} \left( (1-\phi_1)\sigma_1^{-1} \right) \mathbf{u}_1^T \mathbf{b}_{exact} \\ \left( (1-\phi_2)\sigma_2^{-1} \right) \mathbf{u}_2^T \mathbf{b}_{exact} \\ \vdots \\ \left( (1-\phi_N)\sigma_N^{-1} \right) \mathbf{u}_N^T \mathbf{b}_{exact} \end{pmatrix} \right\|_2^2 = \sum_{i=1}^{N} \left( (1-\phi_i) \frac{\mathbf{u}_i^T \mathbf{b}_{exact}}{\sigma_i} \right)^2. \tag{3.29}$$

By a proper design of $\phi_i$, we can limit (3.29) to a small value since

$$\begin{cases} \left| \dfrac{\mathbf{u}_i^T \mathbf{b}_{exact}}{\sigma_i} \right| \text{ decays on average against the index } i, \text{ according to the } \textit{Discrete Picard Condition} \text{ discussed in Section 3.1.4.2.} \\ \phi_i \text{ is designed to be closer to 1 for small index } i, \text{ while small for large index } i. \end{cases}$$

(3.30)

So, large values of $\mathbf{u}_i^T \mathbf{b}_{exact} / \sigma_i$ for small indexes $i$ are damped by small $1-\phi_i$, and large values of $1-\phi_i$ for large indexes $i$ are damped by small $\mathbf{u}_i^T \mathbf{b}_{exact} / \sigma_i$.

### 3.1.4.5 Summarize on the Spectral Filtering points of view

It is worth a summary here. Spectral Filtering can mitigate the impact of inverted noise by introducing a parameterized filter $\mathbf{\Phi}$. This $\mathbf{\Phi}$ is controlled by parameters such as the truncation parameter $k$ in the TSVD Method, or the regularization parameter $\alpha$ in the Tikhonov Method. The extra Regularization Error introduced can be controlled by an appropriate choice of such (regularization) parameters if the Discrete Picard

Condition is met.

In the next section, we analyze the Spectral Regularization Methods from the perspective of a minimization problem with a residual term and a regularization term. Although the presentation may be different, the essence is the same as we shall see in Section 3.1.5.2.

### 3.1.5 View Spectral Regularization from the Optimization perspective

#### 3.1.5.1 An Optimization Objective

For the linear blurring model

$$\mathbf{b} = \mathbf{A}\mathbf{x} + \mathbf{e}, \text{ where } \begin{cases} \mathbf{b} \in \mathbb{R}^{N \times 1} \text{ is the vectorized blurry observation,} \\ \mathbf{A} \in \mathbb{R}^{N \times N} \text{ is the forward operator,} \\ \mathbf{x} \in \mathbb{R}^{N \times 1} \text{ is a vectorized sharp image,} \\ \mathbf{e} \in \mathbb{R}^{N \times 1} \text{ is the vectorized noise,} \end{cases} \quad (3.31)$$

we seek a reconstruction $\mathbf{x}_{\text{recon}}$ from the following optimization objective

$$\mathbf{x}_{\text{recon}} = \arg\min_{\mathbf{x}} \left\{ \underbrace{\|\mathbf{b} - \mathbf{A}\mathbf{x}\|_2^2}_{\text{Square of the Residual Norm}} + \alpha^2 \cdot \underbrace{\|\mathbf{D}\mathbf{x}\|_2^2}_{\text{Regularization Term}} \right\}, \text{ where } \mathbf{D} \text{ is the regularization matrix.}$$

(3.32)

This is called the *General Tikhonov Regularization,* or the *Damped Least Squares Method* [26]. We will discuss some other regularization types in Section 3.1.5.3, Section 3.2 and Section 3.3. For now, let us start from (3.32).

The minimization objective in (3.32) is the square of the *Residual Norm* plus a *Regularization Term* weighted by $\alpha^2$ – the regularization parameter squared. Let us discuss them one by one:

➢ The Residual Norm $\|\mathbf{b} - \mathbf{A}\mathbf{x}\|_2$ measures how well the reconstruction fits the blurry observation. On one hand we want the restored image matches the observation data; on the other hand, the "exact match" yields the inversed observation (*i.e.*, the naïve solution) $\mathbf{A}^{-1}\mathbf{b}$ dominated by high-frequency stuffs as we saw in Figure 4 (B) and Figure 5 (C). Clearly, we need to control the amount of goodness-of-fit and draw the solution away from the inversed observation.

➢ The Regularization Term $\|\mathbf{D}\mathbf{x}\|_2^2$ reflects our expectation of a well-reconstructed image, such that for a good reconstruction, $\|\mathbf{D}\mathbf{x}_{\text{recon}}^{(\text{good})}\|_2^2$ should be small. For example, we know the naïve solution is dominated by high-frequency interruptions.

Then we can design a regularization matrix $\mathbf{D}$ such that $\|\mathbf{Dx}\|_2^2$ produces the intensity of edges in an image. The intensity of edges can be an indication suggesting the proportion of high-frequency information in an image. So, by reducing $\|\mathbf{Dx}\|_2^2$ the image is smoothed out, and some of the high-frequency interruptions are removed as well. The regularization term is typically designed subject to our prior knowledge of the image, and we shall see an example in Section 3.2.

➢ The regularization parameter $\alpha$ controls the amount of regularization. For a small $\alpha$ the Residual Norm dominates in the optimization and then the result tends to show more properties of the naïve reconstruction (under-smoothed). With a large $\alpha$ the Regularization Term governs, and the reconstruction is likely to be over-smoothed. In contrast with the case for designing the Regularization Term when prior knowledge is applicable, a proper $\alpha$ depends on information that is inaccessible before processing, such as the data misfit $\|\mathbf{b} - \mathbf{Ax}\|_2$, the noise $\mathbf{e}$, *etc*.

In the next section, we will introduce several approaches for choosing such $\alpha$. And many algorithms, including the ones we shall see in Section 3.3, treat $\alpha$ as a hyperparameter up to trial and error.

The Tikhonov Regularization of (3.32) can be treated as a Linear Least Squares problem for numerical least-squares algorithms to carry out efficiently. For two vectors $\mathbf{a}$ and $\mathbf{b}$, the sum of their squared 2-norm is going to be

$$\|\mathbf{a}\|_2^2 + \|\mathbf{b}\|_2^2 = \mathbf{a}^T\mathbf{a} + \mathbf{b}^T\mathbf{b} = \begin{pmatrix}\mathbf{a}\\\mathbf{b}\end{pmatrix}^T \begin{pmatrix}\mathbf{a}\\\mathbf{b}\end{pmatrix} = \left\|\begin{pmatrix}\mathbf{a}\\\mathbf{b}\end{pmatrix}\right\|_2^2, \text{ where } \begin{cases}\mathbf{a} \in \mathbb{R}^{N\times 1}\\ \mathbf{b} \in \mathbb{R}^{N\times 1}\end{cases}. \qquad (3.33)$$

So, the sum of the Residual Norm and the weighted Regularization Term is

$$\|\mathbf{b}-\mathbf{Ax}\|_2^2 + \alpha^2 \|\mathbf{Dx}\|_2^2 = \|\mathbf{b}-\mathbf{Ax}\|_2^2 + \|(-\alpha)\mathbf{Dx}\|_2^2 = \|\mathbf{b}-\mathbf{Ax}\|_2^2 + \|\mathbf{0}-\alpha\mathbf{Dx}\|_2^2$$
$$= \left\|\begin{pmatrix}\mathbf{b}-\mathbf{Ax}\\ \mathbf{0}-\alpha\mathbf{Dx}\end{pmatrix}\right\|_2^2, \text{ where } \begin{cases}\mathbf{b}-\mathbf{Ax} \in \mathbb{R}^{N\times 1}\\ \mathbf{0}-\alpha\mathbf{Dx} \in \mathbb{R}^{N\times 1}\end{cases}. \qquad (3.34)$$

Therefore, the optimization problem in (3.32) can be rewritten as

$$\mathbf{x}_{\text{recon}} = \arg\min_{\mathbf{x}}\left\{\|\mathbf{b}-\mathbf{Ax}\|_2^2 + \alpha^2 \cdot \|\mathbf{Dx}\|_2^2\right\} = \arg\min_{\mathbf{x}}\left\{\left\|\begin{pmatrix}\mathbf{b}-\mathbf{Ax}\\ \mathbf{0}-\alpha\mathbf{Dx}\end{pmatrix}\right\|_2^2\right\}$$
$$= \arg\min_{\mathbf{x}}\left\{\left\|\begin{pmatrix}\mathbf{b}\\ \mathbf{0}\end{pmatrix}-\begin{pmatrix}\mathbf{A}\\ \alpha\mathbf{D}\end{pmatrix}\mathbf{x}\right\|_2^2\right\}, \text{ where } \begin{cases}\alpha \text{ is the regularization parameter, and}\\ \mathbf{D} \text{ is the regularization matrix.}\end{cases}$$

(3.35)

Numerical computations can be carried out directly on (3.35) because it is now simply a Linear Least Squares problem on $\mathbf{x}$.

### 3.1.5.2 Methods for choosing the Regularization Parameter

In this section we discuss how to choose a regularization parameter for the Spectral Regularization Methods. Recall that in Section 3.1.4.3, we presented two spectral filtering methods – the TSVD and the Tikhonov Method – with (regularization) parameter $k$ and $\alpha$, respectively. And there is a link between the optimization objective (3.32) and the Tikhonov filter factors in the form of (3.24). Let us reveal it first.

Suppose the regularization matrix $\mathbf{D}$ in (3.32) is now an identity matrix, *i.e.*,

$$\mathbf{D} = \mathbf{I}_N \quad \Rightarrow \quad \arg\min_{\mathbf{x}} \left\{ \|\mathbf{b} - \mathbf{A}\mathbf{x}\|_2^2 + \alpha^2 \cdot \|\mathbf{D}\mathbf{x}\|_2^2 \right\} = \arg\min_{\mathbf{x}} \left\{ \|\mathbf{b} - \mathbf{A}\mathbf{x}\|_2^2 + \alpha^2 \cdot \|\mathbf{x}\|_2^2 \right\}.$$

(3.36)

And we have already known, from (3.26-1), that for a $\mathbf{x}_{\text{recon}}^{(\text{filtered})}$,

$$
\begin{aligned}
\left\| \mathbf{b} - \mathbf{A}\mathbf{x}_{\text{recon}}^{(\text{filtered})} \right\|_2^2 &= \left\| \mathbf{b} - \left( \mathbf{U}\boldsymbol{\Sigma}\mathbf{V}^\text{T} \right) \cdot \left( \mathbf{V}\boldsymbol{\Phi}\boldsymbol{\Sigma}^{-1}\mathbf{U}^\text{T}\mathbf{b} \right) \right\|_2^2 \\
&= \left\| \mathbf{b} - \mathbf{U}\boldsymbol{\Sigma}\boldsymbol{\Phi}\boldsymbol{\Sigma}^{-1}\mathbf{U}^\text{T}\mathbf{b} \right\|_2^2 \\
&= \left\| \mathbf{U}^\text{T} \left( \mathbf{b} - \mathbf{U}\boldsymbol{\Sigma}\boldsymbol{\Phi}\boldsymbol{\Sigma}^{-1}\mathbf{U}^\text{T}\mathbf{b} \right) \right\|_2^2 \\
&= \left\| \mathbf{U}^\text{T}\mathbf{b} - \boldsymbol{\Sigma}\boldsymbol{\Phi}\boldsymbol{\Sigma}^{-1}\mathbf{U}^\text{T}\mathbf{b} \right\|_2^2, \text{ where } \begin{cases} \boldsymbol{\Sigma} = \text{diag}(\sigma_1, \sigma_2, \ldots, \sigma_N) \\ \boldsymbol{\Phi} = \text{diag}(\phi_1, \phi_2, \ldots, \phi_N) \end{cases} \\
&= \left\| \mathbf{U}^\text{T}\mathbf{b} - \boldsymbol{\Phi}\boldsymbol{\Sigma}\boldsymbol{\Sigma}^{-1}\mathbf{U}^\text{T}\mathbf{b} \right\|_2^2 \\
&= \left\| \mathbf{U}^\text{T}\mathbf{b} - \boldsymbol{\Phi}\mathbf{U}^\text{T}\mathbf{b} \right\|_2^2 = \left\| (\mathbf{I}_N - \boldsymbol{\Phi})\mathbf{U}^\text{T}\mathbf{b} \right\|_2^2 \\
&= \left\| \begin{pmatrix} 1-\phi_1 & 0 & \cdots & 0 \\ 0 & 1-\phi_2 & \cdots & 0 \\ \vdots & \vdots & \ddots & \vdots \\ 0 & 0 & \cdots & 1-\phi_N \end{pmatrix} \begin{pmatrix} \mathbf{u}_1^\text{T}\mathbf{b} \\ \mathbf{u}_2^\text{T}\mathbf{b} \\ \vdots \\ \mathbf{u}_N^\text{T}\mathbf{b} \end{pmatrix} \right\|_2^2 \\
&= \left\| \begin{pmatrix} (1-\phi_1)\mathbf{u}_1^\text{T}\mathbf{b} \\ (1-\phi_2)\mathbf{u}_2^\text{T}\mathbf{b} \\ \vdots \\ (1-\phi_N)\mathbf{u}_N^\text{T}\mathbf{b} \end{pmatrix} \right\|_2^2 = \sum_{i=1}^{N} \left( (1-\phi_i)\mathbf{u}_i^\text{T}\mathbf{b} \right)^2.
\end{aligned}
$$

(3.37)

Also, according to (3.22) the vectorized reconstruction is

$$\mathbf{x}_{\text{recon}}^{(\text{filtered})} = \sum_{i=1}^{N} \phi_i \frac{\langle \mathbf{u}_i, \mathbf{b} \rangle}{\sigma_i} \mathbf{v}_i = \sum_{i=1}^{N} \phi_i \frac{\mathbf{u}_i^\text{T}\mathbf{b}}{\sigma_i} \mathbf{v}_i \quad (3.38)$$

which means

$$\left\| \mathbf{x}_{\text{recon}}^{(\text{filtered})} \right\|_2^2 = \sum_{i=1}^{N} \left( \phi_i \frac{\mathbf{u}_i^\text{T}\mathbf{b}}{\sigma_i} \right)^2. \quad (3.39)$$

Therefore,

$$\left\|\mathbf{b}-\mathbf{A}\mathbf{x}_{\text{recon}}^{(\text{filtered})}\right\|_2^2 + \alpha^2 \left\|\mathbf{x}_{\text{recon}}^{(\text{filtered})}\right\|_2^2 = \sum_{i=1}^{N}\left((1-\phi_i)\mathbf{u}_i^{\text{T}}\mathbf{b}\right)^2 + \alpha^2 \cdot \sum_{i=1}^{N}\left(\phi_i \frac{\mathbf{u}_i^{\text{T}}\mathbf{b}}{\sigma_i}\right)^2$$

$$= \sum_{i=1}^{N}\left\{\left[(1-\phi_i)^2 + \alpha^2\left(\frac{\phi_i}{\sigma_i}\right)^2\right](\mathbf{u}_i^{\text{T}}\mathbf{b})^2\right\} \quad (3.40)$$

$$= \sum_{i=1}^{N}\left\{\left[\frac{1}{\sigma_i^2}\left(\sigma_i^2(1-\phi_i)^2 + \alpha^2\phi_i^2\right)\right](\mathbf{u}_i^{\text{T}}\mathbf{b})^2\right\}.$$

We want the filtered reconstruction (3.38), which is controlled by $\phi_i$, to minimize (3.40). As shown above, it is a quadratic function of $\phi_i$, with the derivative with respect to $\phi_i$ as

$$\frac{d}{d\phi_i}\left[\left\|\mathbf{b}-\mathbf{A}\mathbf{x}_{\text{recon}}^{(\text{filtered})}\right\|_2^2 + \alpha^2 \left\|\mathbf{x}_{\text{recon}}^{(\text{filtered})}\right\|_2^2\right] = \sum_{i=1}^{N}\left\{\left[\frac{1}{\sigma_i^2}\cdot\frac{d}{d\phi_i}\left(\sigma_i^2(1-\phi_i)^2 + \alpha^2\phi_i^2\right)\right](\mathbf{u}_i^{\text{T}}\mathbf{b})^2\right\}$$

(3.41)

Thus, a proper $\phi_i$ should satisfy

$$\frac{d}{d\phi_i}\left(\sigma_i^2(1-\phi_i)^2 + \alpha^2\phi_i^2\right) = 0 \;\Rightarrow\; -2\sigma_i^2(1-\phi_i) + 2\alpha^2\phi_i = 0 \;\Rightarrow\; \phi_i = \frac{\sigma_i^2}{\sigma_i^2 + \alpha^2}.$$

(3.42)

Now, we have successfully demonstrated that the solution to the optimization problem

$$\mathbf{x}_{\text{recon}} = \arg\min_{\mathbf{x}}\left\{\left\|\mathbf{b}-\mathbf{A}\mathbf{x}\right\|_2^2 + \alpha^2\left\|\mathbf{x}\right\|_2^2\right\} \quad (3.43)$$

has the form

$$\mathbf{x}_{\text{recon}} = \sum_{i=1}^{N}\phi_i\frac{\mathbf{u}_i^{\text{T}}\mathbf{b}}{\sigma_i}\mathbf{v}_i, \text{ where } \phi_i = \frac{\sigma_i^2}{\sigma_i^2 + \alpha^2} \quad (3.44)$$

which is exactly result of the Tikhonov filter specified in (3.24).

There are three approaches – the Discrepancy Principle [26], the Generalized Cross Validation (GCV) [27], and the L-Curve Criterion [28, 29] – to decide the regularization parameter $\alpha$ in (3.43). The Discrepancy Principle depends on an accurate estimate of $\|\mathbf{e}\|_2$ – the 2-norm of the noise – in the blurry observation. So, plots like Figure 8 could be useful since the noise plateau there indicating the intensity of the noise. The L-Curve Criterion is quite straightforward. For each candidate in a set of regularization parameters, it records the norm of the corresponding residual, $\|\mathbf{b}-\mathbf{A}\mathbf{x}_{\text{recon}}\|_2$, and the norm of the corresponding solution, $\|\mathbf{x}_{\text{recon}}\|_2$. Then the norms of residual are plotted against the norms of solution, and the best regularization parameter would lie at the corner of the "L-shape" curve. Figure 11 shows an example of such L-curves.

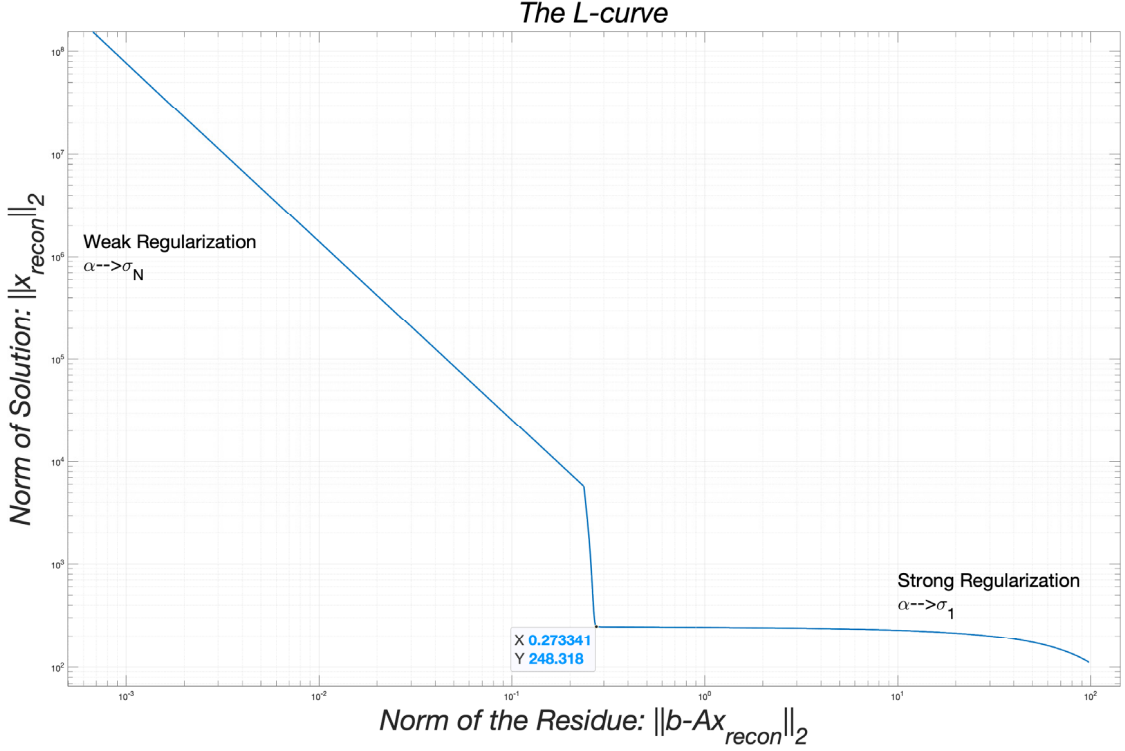

***Figure 11.*** *An L-curve example with its corner spotted.*

As we discussed in Section 3.1.5.1, a weak regularization leads to a under-smoothed reconstruction which is much like the naïve solution; with a strong regularization, however, the solution is over-smoothed. We want a balance between the norms of the solution and the residual, *i.e.*, they are simultaneously small. So, the reconstruction would be neither dominated by high frequency nor over-smoothed. It is the idea behind the L-Curve Criterion.

The Generalized Cross Validation (GCV) produces an optimal value for the regularization parameter $\alpha$ by

$$\alpha_{opt} = \arg\min_{\alpha} G(\alpha) = \arg\min_{\alpha} \left\{ \frac{\left\| \left( \mathbf{I}_N - \mathbf{AV\Phi\Sigma}^{-1}\mathbf{U}^T \right)\mathbf{b} \right\|_2^2}{\left( \mathrm{tr}\left( \mathbf{I}_N - \mathbf{AV\Phi\Sigma}^{-1}\mathbf{U}^T \right) \right)^2} \right\}. \tag{3.45}$$

From (3.26-1) and (3.37) we know the numerator of $G(\alpha)$ is

$$\left\| \left( \mathbf{I}_N - \mathbf{AV\Phi\Sigma}^{-1}\mathbf{U}^T \right)\mathbf{b} \right\|_2^2 = \left\| \mathbf{b} - \mathbf{Ax}_{\text{recon}}^{(\text{filtered})} \right\|_2^2 = \sum_{i=1}^{N} \left( (1-\phi_i)\mathbf{u}_i^T\mathbf{b} \right)^2. \tag{3.46}$$

And the trace in the denominator is

$$\begin{aligned}
\mathrm{tr}\left( \mathbf{I}_N - \mathbf{AV\Phi\Sigma}^{-1}\mathbf{U}^T \right) &= \mathrm{tr}\left( \mathbf{I}_N - \left(\mathbf{U\Sigma V}^T\right)\cdot\left(\mathbf{V\Phi\Sigma}^{-1}\mathbf{U}^T\right) \right) \\
&= \mathrm{tr}\left( \mathbf{I}_N - \mathbf{U\Sigma\Phi\Sigma}^{-1}\mathbf{U}^T \right) = \mathrm{tr}\left( \mathbf{I}_N - \mathbf{U\Phi\Sigma\Sigma}^{-1}\mathbf{U}^T \right) \\
&= \mathrm{tr}\left( \mathbf{I}_N - \mathbf{U\Phi U}^T \right) = \mathrm{tr}\left( \mathbf{U}\left(\mathbf{I}_N - \mathbf{\Phi}\right)\mathbf{U}^T \right) = \mathrm{tr}\left( \left(\mathbf{I}_N - \mathbf{\Phi}\right)\mathbf{U}^T\mathbf{U} \right) \\
&= \mathrm{tr}\left( \mathbf{I}_N - \mathbf{\Phi} \right) = \sum_{i=1}^{N}(1-\phi_i) = N - \sum_{i=1}^{N}\phi_i.
\end{aligned} \tag{3.47}$$

Then (3.45) becomes

$$\alpha_{opt} = \arg\min_{\alpha} G(\alpha) = \arg\min_{\alpha}\left\{\frac{\sum_{i=1}^{N}\left((1-\phi_i)\mathbf{u}_i^T\mathbf{b}\right)^2}{\left(N-\sum_{i=1}^{N}\phi_i\right)^2}\right\}. \tag{3.48}$$

For the $\alpha$-Tikhonov filter introduced by (3.24),

$$G(\alpha) = \frac{\sum_{i=1}^{N}\left((1-\phi_i)\mathbf{u}_i^T\mathbf{b}\right)^2}{\left(N-\sum_{i=1}^{N}\phi_i\right)^2} = \frac{\sum_{i=1}^{N}\left(\frac{\alpha^2}{\sigma_i^2+\alpha^2}\mathbf{u}_i^T\mathbf{b}\right)^2}{\left(\sum_{i=1}^{N}\frac{\alpha^2}{\sigma_i^2+\alpha^2}\right)^2} = \frac{\alpha^4 \cdot \sum_{i=1}^{N}\left(\frac{1}{\sigma_i^2+\alpha^2}\mathbf{u}_i^T\mathbf{b}\right)^2}{\left(\alpha^2 \cdot \sum_{i=1}^{N}\frac{1}{\sigma_i^2+\alpha^2}\right)^2} = \frac{\sum_{i=1}^{N}\left(\frac{1}{\sigma_i^2+\alpha^2}\mathbf{u}_i^T\mathbf{b}\right)^2}{\left(\sum_{i=1}^{N}\frac{1}{\sigma_i^2+\alpha^2}\right)^2}$$

(3.49)

so, the optimal regularization parameter $\alpha$ is found by

$$\alpha_{opt} = \arg\min_{\alpha} G(\alpha) = \arg\min_{\alpha}\left\{\frac{\sum_{i=1}^{N}\left(\frac{1}{\sigma_i^2+\alpha^2}\mathbf{u}_i^T\mathbf{b}\right)^2}{\left(\sum_{i=1}^{N}\frac{1}{\sigma_i^2+\alpha^2}\right)^2}\right\}. \tag{3.50}$$

Similarly, for the $k$-TSVD filter introduced by (3.23),

$$G(k) = \frac{\sum_{i=1}^{N}\left((1-\phi_i)\mathbf{u}_i^T\mathbf{b}\right)^2}{\left(N-\sum_{i=1}^{N}\phi_i\right)^2} = \frac{\sum_{i=k+1}^{N}\left(\mathbf{u}_i^T\mathbf{b}\right)^2}{\left(N-\sum_{i=1}^{k}\phi_i\right)^2} = \frac{\sum_{i=k+1}^{N}\left(\mathbf{u}_i^T\mathbf{b}\right)^2}{(N-k)^2} \tag{3.51}$$

and the optimal truncation parameter $k$ is found by

$$k_{opt} = \arg\min_{k} G(k) = \arg\min_{k}\left\{\frac{\sum_{i=k+1}^{N}\left(\mathbf{u}_i^T\mathbf{b}\right)^2}{(N-k)^2}\right\}. \tag{3.52}$$

Figure 12 displays three deblurring results produced by the Tikhonov filter (3.44), with the Discrepancy Principle, the Generalized Cross Validation, and the L-Curve Criterion to determine the regularization parameter $\alpha$, respectively.

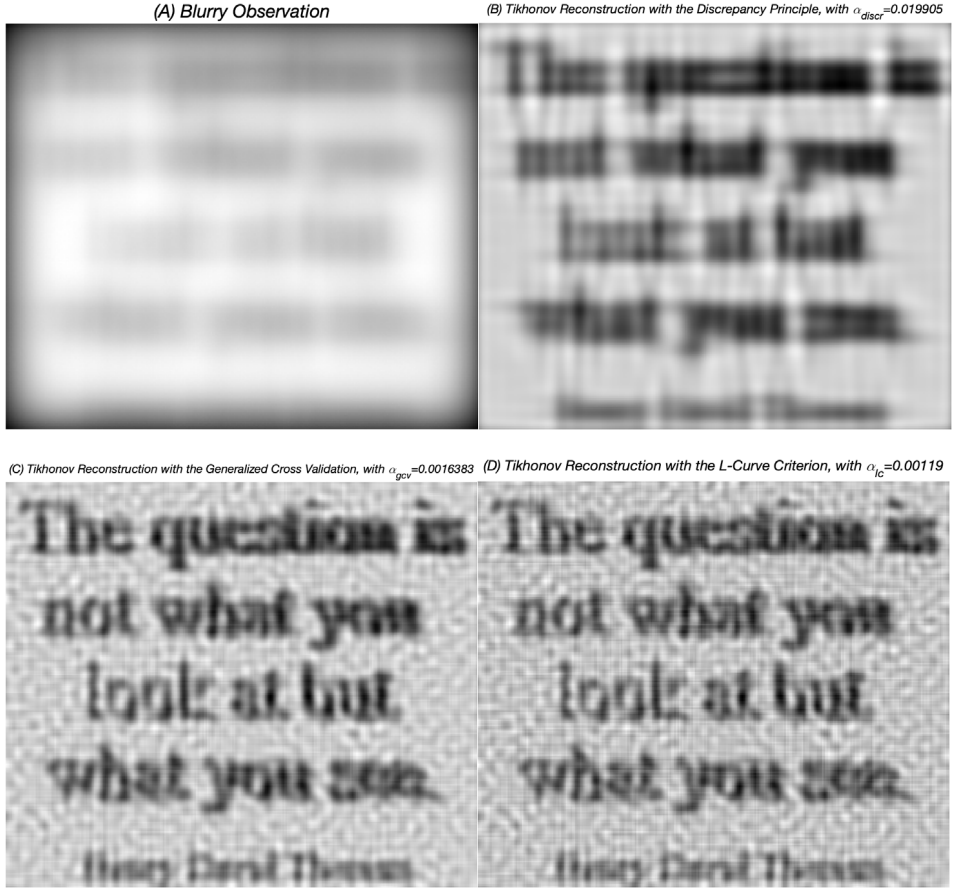

*Figure 12. Tikhonov reconstructions, with the Discrepancy Principle, the Generalized Cross Validation, and the L-Curve Criterion finding $\alpha$ to be 0.019905, 0.0016383, and 0.00119, respectively.*

According to Figure 12, we notice that the Discrepancy Principle over-smooths the reconstruction by producing a large regularization parameter, while others can reconstruct an image with good perceptual quality. The Discrepancy Principle relies on a good estimate of the noise level in the observation but sometimes this information is inaccurate. And as indicated by [30], it still tends to over-smooth the image even on an accurate estimation. The other two methods also have some flaws. They both cannot converge to the true solution if the error norm decays to zero [31]. And for the Generalized Cross Validation method, the function G can have very gentle gradients near the minimizer [32]. This challenges numerical methods when determining a good value of $\alpha$.

A proper choice of the regularization parameter is an essential point, and we shall see, evaluate, and justify them in methods we shall discuss later.

### 3.1.5.3 Regularization with different smoothing norms

The General Tikhonov Regularization objective introduced in (3.32) have some variations according to the specific case we are dealing with. A practical concern is that the Regularization Term, in the form of the smoothing norm $\|\mathbf{Dx}\|_2^2$, tends to smooth the reconstruction by too much. Recall the definition of the $p$-norm

$$\|\mathbf{y}\|_p = \left(|y_1|^p + |y_2|^p + \ldots + |y_N|^p\right)^{1/p}, \text{ where } \begin{cases} \mathbf{y} = (y_1, y_2, \ldots, y_N)^T, \text{ and} \\ 1 \le p \le 2, \end{cases} \quad (3.53)$$

it is easy to see $\|\mathbf{y}\|_p^p$ increases with $p$. We can do a numerical test on $\mathbf{y}^{(1)} = (1, 2, 3, 4)^T$, $\mathbf{y}^{(2)} = (1, 0.2, 3, 4)^T$, and $\mathbf{y}^{(3)} = (1, 0.2, 0.3, 4)^T$, as shown in the Table 2 below. Now, suppose $\mathbf{Dx}$ computes the derivative of an image, which is indicative of the edge intensity. For a large $p$ the measurement $\|\mathbf{Dx}\|_p^p$ is high, and as one component in the minimization objective (3.32) to be penalized, it results in a reconstruction $\mathbf{x}_{recon}$ with a very low edge intensity. Therefore, this $\mathbf{x}_{recon}$ tends to be very smoothed. For a small $p$ such a smoothing impact can be weakened. So, instead of (3.32) where $p=2$, as an alternative we can find the optimum by

$$\mathbf{x}_{recon} = \arg\min_{\mathbf{x}} \left\{ \|\mathbf{b} - \mathbf{Ax}\|_2^2 + \alpha^2 \cdot \|\mathbf{Dx}\|_p^p \right\}, \text{ where } 1 \le p < 2. \quad (3.54)$$

*Table 2. Numerical Test for* $\|\cdot\|_p^p$ *with three vectors*

| $p$ | 1 | 1.2 | 1.4 | 1.7 | 2 |
|---|---|---|---|---|---|
| $\|\mathbf{y}^{(1)}\|_p^p$ | 10.0 | 12.31 | 15.26 | 21.28 | 30 |
| $\|\mathbf{y}^{(2)}\|_p^p$ | 8.2 | 10.16 | 12.73 | 18.09 | 26.04 |
| $\|\mathbf{y}^{(3)}\|_p^p$ | 5.5 | 6.66 | 8.25 | 11.75 | 17.13 |

Another variation to (3.32) is one called *Sparsity Regularization* [33]. This time the optimum $\mathbf{x}_{recon}$ is found by

$$\mathbf{x}_{recon} = \arg\min_{\mathbf{x}} \left\{ \|\mathbf{b} - \mathbf{Ax}\|_2^2 + \alpha^2 \cdot \|\mathbf{Dx}\|_0 \right\}, \text{ where } \|\cdot\|_0 \text{ finds the number of non-zero entries in a vector.}$$

(3.55)

Taking part in the minimization and getting rewards for small values, $\|\mathbf{Dx}\|_0$ leads to a solution $\mathbf{x}_{recon}$ with sparse edges if $\mathbf{D}$ is the derivative operator. Now, $\mathbf{Dx}_{recon}$ has more zero entries, which means $\mathbf{x}_{recon}$ has more blocks with the same pixels. If we want to reconstruct an image with high contrast, then this sparsity regularization can prevent our solution from being smoothed out.

## 3.2 A MAP-based regularization method [7]

In Section 3.1 we gave a detailed account of the spectral regularization methods where analysis and operations are in the SVD coordinate system. In Section 3.1.5 we presented the minimization objective, which is the summation of a residual norm and a regularization term weighted by a regularization parameter. We mentioned that the regularization term can be designed in accordance with the prior information we assume for the image. In 2008, Almeida *et al.* [7] proposed a method to cope with the blind image deblurring (BID) problem, where the regularization term was designed based on the idea of maximum a posteriori (MAP). It is a good example to demonstrate how the prior knowledge can be used for the optimization objective. In this section, we focus on the handcraft process.

### 3.2.1 Maximum A Posterior Estimation

The Maximum A Posterior (MAP) Estimation can be seen as a regularized version of the Maximum Likelihood Estimation (MLE) when extra information of a related event can be used as prior knowledge. MLE works in a straightforward sense. Suppose there is a series of observation $x$ parameterized by the unknown $\boldsymbol{\theta}$, and we want to estimate $\boldsymbol{\theta}$. This problem can be modeled by finding

$$\hat{\boldsymbol{\theta}}_{\text{MLE}} = \arg\max_{\boldsymbol{\theta}} \text{Prob}(x|\boldsymbol{\theta}) \qquad (3.56)$$

which means that we want the best parameter $\boldsymbol{\theta}$ to make this observation $x$ the most likely. Now assume we have some prior knowledge about $\boldsymbol{\theta}$, *e.g.*, we know in advance the distribution of it. And given the observation $x$, we want to pick the most likely $\boldsymbol{\theta}$. Referring to the Bayesian Theorem this problem can be modeled by finding

$$\hat{\boldsymbol{\theta}}_{\text{MAP}} = \arg\max_{\boldsymbol{\theta}} \text{Prob}(\boldsymbol{\theta}|x) = \arg\max_{\boldsymbol{\theta}} \frac{\text{Prob}(x|\boldsymbol{\theta}) \cdot \text{Prob}(\boldsymbol{\theta})}{\text{Prob}(x)}$$
$$= \arg\max_{\boldsymbol{\theta}} \text{Prob}(x|\boldsymbol{\theta}) \cdot \text{Prob}(\boldsymbol{\theta}) \qquad (3.57\text{-}1)$$

since the denominator

$$\text{Prob}(x) = \int_{\boldsymbol{\theta}} \text{Prob}(x|\boldsymbol{\theta}) \cdot \text{Prob}(\boldsymbol{\theta}) \, d\boldsymbol{\theta} \qquad (3.57\text{-}2)$$

is irrelevant to $\boldsymbol{\theta}$ after the integration over all possible $\boldsymbol{\theta}$. This is the Maximum A Posterior (MAP) Estimation, where we want the mode of posterior distribution of the random variable $\boldsymbol{\theta}$. As (3.57) suggests, compared with MLE this estimation is regularized by the prior knowledge we have.

### 3.2.2 Problem definition

Similar to (2.1), they assume a linear blurring model

$$y = h * x + n \tag{3.58}$$

where the small case *y*, *h*, *x*, and *n* refers to the blurry observation, the PSF, the underlying truth, and the additive noise, respectively. They are all 2D signals, but the authors did not adopt the capital notation, perhaps because there is almost nothing need to be demonstrated by matrix operations as we did in Section 3.1. They designed an edge detector *f(x)* to produce an "derivative image" saving the intensities of edges in an image. And for the blind image deblurring (BID) problem the job is to find

$$\langle x_{\text{recon}}, h_{\text{recon}} \rangle = \arg\min_{\langle x, h \rangle} \left\{ \frac{1}{2} \| y - h * x \|_2^2 + \lambda \, \text{R}[f(x)] \right\}. \tag{3.59}$$

We saw the optimization objective in the right-hand side is quite like the form we defined in (3.32) with a residual norm and a regularization term. The problem now is how to craft by hand the regularization term R[*f*(x)], and therefore form an optimization objective (3.59).

### 3.2.3 Craft by hand the Optimization Objective

There are three key hypotheses here. The first hypothesis is made on the probability distribution of edge intensities in a natural image, *i.e.*, the edge intensity of the $i^{\text{th}}$ pixel follows an exponential distribution model

$$\text{Prob}(f_i(x)|x) \propto e^{-k(f_i(x)+\varepsilon)^q}. \tag{3.60}$$

where *k*, *ε*, and *q* are adjustable parameters. Figure 13 illustrates an example of such a probability density distribution.

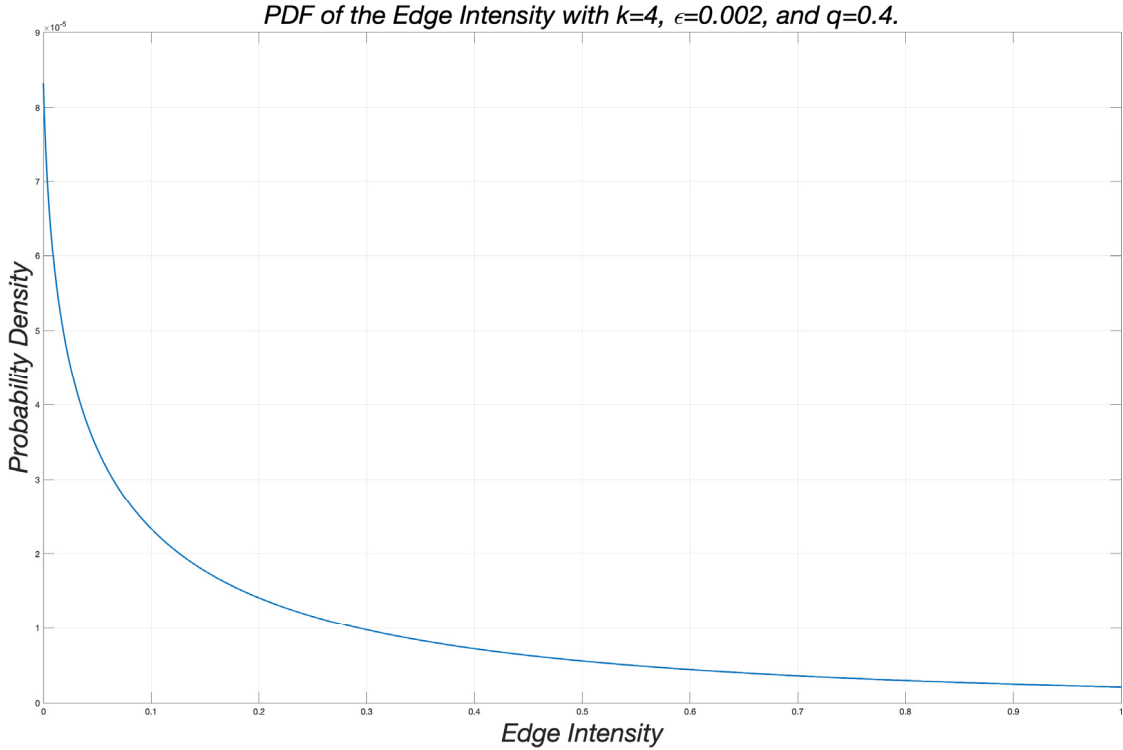

*Figure 13. An example of the sparse prior defined in (3.60).*

This exponential distribution reflects the Sparse Edge Assumption – zero edge intensities dominate in a nature image with only a few leading edges exhibit strong intensities. The second hypothesis is that the noise term, $n$ in (3.58), follows a Gaussian distribution with zero mean and variance $\sigma^2$. So, the posterior distribution of $<x, f(x), h>$ given the observation $y$ is

$$\text{Prob}(x, f(x), h | y) = \text{Prob}(y | x, f(x), h) \cdot \text{Prob}(x, f(x), h)$$
$$\propto e^{-\frac{1}{2\sigma^2}\|y - h*x\|_2^2} \cdot \text{Prob}(x, f(x), h). \tag{3.61}$$

The PSF $h$ is independent from the image $x$ and its derivative $f(x)$, so

$$\text{Prob}(x, f(x), h) = \text{Prob}(x, f(x)) \cdot \text{Prob}(h)$$
$$= \text{Prob}(f(x) | x) \cdot \text{Prob}(x) \cdot \text{Prob}(h). \tag{3.62}$$

If there is no prior knowledge on the PSF $h$ and the image $x$ itself, then assume that they follow the uniform distribution on their respective ranges. In this sense,

$$\text{Prob}(x, f(x), h) \propto \text{Prob}(f(x) | x) \tag{3.63}$$

and (3.61) becomes

$$\text{Prob}(x, f(x), h | y) \propto e^{-\frac{1}{2\sigma^2}\|y - h*x\|_2^2} \cdot \text{Prob}(f(x) | x). \tag{3.64}$$

The last hypothesis is that edge intensities for different pixels are independent from each other, i.e.,

$$\text{Prob}(f(x) | x) = \prod_i \text{Prob}(f_i(x) | x) = \prod_i e^{-k(f_i(x) + \varepsilon)^q}. \tag{3.65}$$

Therefore, the posterior distribution is

$$\text{Prob}(x, f(x), h | y) \propto e^{-\frac{1}{2\sigma^2}\|y - h*x\|_2^2} \prod_i e^{-k(f_i(x) + \varepsilon)^q} \tag{3.66}$$

and by maximizing it the solution $<x_{\text{recon}}, h_{\text{recon}}>$ can be found.

The log-likelihood of (3.66) is then

$$L(x, f(x), h | y) = \log(\text{Prob}(x, f(x), h | y)) = -\frac{1}{2\sigma^2}\|y - h*x\|_2^2 - k\sum_i (f_i(x) + \varepsilon)^q + \text{constant}. \tag{3.67}$$

Maximizing (3.67) is equivalent to minimizing

$$G(x, f(x), h | y) = \frac{1}{2\sigma^2}\|y - h*x\|_2^2 + k\sum_i (f_i(x) + \varepsilon)^q$$
$$= \frac{1}{\sigma^2}\left\{\frac{1}{2}\|y - h*x\|_2^2 + k\sigma^2\sum_i (f_i(x) + \varepsilon)^q\right\}. \tag{3.68}$$

Therefore, the solution $<x_{\text{recon}}, h_{\text{recon}}>$ is obtained by finding

$$\langle x_{\text{recon}}, h_{\text{recon}} \rangle = \arg\min_{\langle x, h \rangle} \left\{\frac{1}{2}\|y - h*x\|_2^2 + k\sigma^2\sum_i (f_i(x) + \varepsilon)^q\right\}. \tag{3.69}$$

Look back at (3.59), now the crafted regularization term is

$$R[f(x)] = \sum_i (f_i(x)+\varepsilon)^q, \text{ where } \begin{cases} f_i(x) \text{ detects the edge intensity for the } i^{th} \text{ pixel,} \\ \varepsilon \text{ and } q \text{ model the Sparse Edge Assumption specified in (3.60).} \end{cases}$$

(3.70)

And the regularization parameter $\lambda$ corresponds to $k\sigma^2$ in (3.69), where $k$ controls (3.60) and $\sigma^2$ is the variance of the Gaussian additive noise assumed.

One thing to note here is that for a well-reconstructed $x_{recon}$, the regularization term $R[f(x_{recon})]$ should be small because a small $R[f(x_{recon})]$ can increase the posterior probability specified in (3.66). In this sense this reconstructed image is up to our expectation specified in the model (3.60), *i.e.*, the Sparse Edge Assumption for a good image.

### 3.2.4 Some Comments

After crafting the minimization objective (3.69), they optimize $<x, h>$ in an alternating fashion. For more technical details please check [7]. One thing to note is that they let the regularization parameter $\lambda$ decline on a geometric progression, like

$$\lambda_{n+1} = \lambda_n / r, \text{ where } \begin{cases} \lambda_1 = 2, \\ r = \begin{cases} 3 \text{ for real blurry monochrome pictures} \\ 1.5 \text{ for synthetic blurry monochrome pictures} \end{cases}, \\ \text{the number of iterations } n \text{ is flexible depending on current restoration quality.} \end{cases}$$

(3.71)

So initially, the reconstruction only contains a few leading sparse edges because of the strong regularization and the Sparse Edge Assumption specified in (3.60). With more iterations, the strength of regularization decreases gradually. Therefore, based on previous results the reconstruction can restore more details.

We present this MAP-based regularization method just for demonstrating the laborious work for crafting the regularization term by hand. This approach is highly model based, as we saw in Section 3.2.3. We have to design carefully the parameters in the model (3.60), and this exponential model is only an approximation to complex cases in real life. Besides, the noise can only be additive here and independent from the intensity of the image. Moreover, the additive noise should follow a Gaussian process with *i.i.d* components. These assumptions may limit the application of this method in real life. In the next section, we will introduce a network-based approach and during the training process, the network will learn a proper regularization term from data. So, we do not need to craft the regularization term by hand there and thus do not need to make such simplified assumptions.

### 3.3. A Deep Learning-based regularization method [8]

The hand-crafted regularization term is based on a simplified model controlled by a few parameters. For example, in Section 3.2.3 the sparse edge assumption is modeled by a simplified exponential PDF. However, the description of sparsity, smoothness, or other prior information we want to involve in the regularization function, can sometimes be very complicated. There may be no room for simplification. In these cases, it is helpful to take advantage of the neural network, which is designed to learn a complex mapping between the input and the output.

In this section, we focus on the method proposed by Lunz et al. [8], where they train a network to learn the regularization functional from data.

### 3.3.1 Problem definition

The generic inverse problem is defined as to recover the image $x \in X$ from the measurement $y \in Y$, under the linear model

$$y = Ax + e. \tag{3.72}$$

The linear forward operator $A$ maps the two spaces, i.e., $A: X \longrightarrow Y$, and $e \in Y$ indicates the random noise term. The distribution of ground truth images is denoted as $P_r$, and the distribution of measurements is denoted as $P_Y$. Suppose the distribution $P_Y$ on the measurement space $Y$ can be mapped back to the image space $X$ with a new distribution $P_n$, by a pseudo-inverse operator $A_\delta^+$, i.e.,

$$P_n = \left(A_\delta^+\right)_\# P_Y, \text{ where } A_\delta^+ Y \sim \left(A_\delta^+\right)_\# P_Y \text{ with } Y \sim P_Y. \tag{3.73}$$

Literature such as [34] has shown that such pseudo-inverse operators are calculable for a variety of forward operators. The pseudo-inverse images on the distribution $P_n$ behaves quite like the noise-corrupted "naïve reconstructions", as we discussed in Section 3.1.1 and 3.1.2. So, the job is to pull the reconstructed image back from the distribution $P_n$ to the distribution $P_r$ where true images lie on.

Previous methods we discussed solve this by finding

$$x_{\text{recon}} = \arg\min_x \left\{\|Ax - y\|_2^2 + \lambda f(x)\right\} \tag{3.74}$$

where $f(x)$ is a hand-crafted regularization functional with our prior knowledge inserted, e.g., the sparsity, smoothness of a good image, etc. So, by giving a small $f(x_{\text{recon}})$ the reconstruction $x_{\text{recon}}$ is up to these expectations, as we discussed in Section 3.2.3.

Now, the hand-crafted $f(x)$ is replaced by a neural network $\Psi_\Theta$ with learnable parameters $\Theta$. Like $f(x)$, we hope it can produce low values for good reconstructions, while high values for bad reconstructions. If we manage to train such a neural network, then the optimization problem is to find

$$x_{\text{recon}} = \arg\min_x \left\{\|Ax - y\|_2^2 + \lambda \cdot \Psi_\Theta(x)\right\}. \tag{3.75}$$

### 3.3.2 Two algorithms

Suppose there are two spaces, $X$ for the Image Space and $Y$ for the Measurement Space. Now we have a dataset containing ground truth images $x_{\text{true}} \in X$. And we transform these images by the model specified in (3.72) to have measurements $y \in Y$. The ground truth images follow a distribution $P_r$, *i.e.*, $x_{\text{true}} \sim P_r$, on the Image Space $X$; and the measurements follow a distribution $P_Y$, *i.e.*, $y \sim P_Y$, on the Measurement Space $Y$. The pseudo-inverse operator $A_\delta^+$ is able to map measurements on $Y$ back to images on $X$, but the pseudo-inversed images $x_{\text{pseudo}}$ follows a different distribution $P_n$ away from $P_r$, *i.e.*,

$$P_n = \left(A_\delta^+\right)_\# P_Y \neq P_r, \text{ where } \begin{cases} x_{\text{pseudo}} \sim P_n \text{ and } x_{\text{pseudo}} \in X, \\ x_{\text{true}} \sim P_r \text{ and } x_{\text{true}} \in X. \end{cases} \quad (3.76)$$

Now, we want a neural network $\Psi_\Theta(x)$ which can distinguish images from these two distributions. The network is designed to achieve this by producing a small scalar for the images in $P_r$, while a large scalar for images in $P_n$. The Loss function is therefore

$$L(\Theta) = \underbrace{E_{X \sim P_r}\left[\Psi_\Theta(X)\right]}_{\text{expected to be small}} - \underbrace{E_{X \sim P_n}\left[\Psi_\Theta(X)\right]}_{\text{expected to be large}} + \mu \cdot E_X\left[\left(\|\nabla_x \Psi_\Theta(X)\| - 1\right)^2\right]. \quad (3.77\text{-}1)$$

The last term in the loss function

$$E_X\left[\left(\|\nabla_x \Psi_\Theta(X)\| - 1\right)^2\right] \quad (3.77\text{-}2)$$

ensures the network $\Psi_\Theta(x)$ to be *Lipschitz-continuous with constant one*, which is necessary. As we shall later in the Algorithm 2, we need to do gradient descent on the optimization objective, so it is beneficial to have a network $\Psi_\Theta(x)$ differentiable everywhere with gradients with respect to $x$ fluactuating closely around a constant (*i.e.*, 1 here). The parameter $\mu$ is the gradient penalty coefficient associated with this Lipschitz-continuous condition.

Figure 14 illustrates the architecture of $\Psi_\Theta$ they used. The CNN takes as input 128×128 grayscale images and outputs two kinds of scalars, corresponding to the pseudo-inversed images and the ground truth images, respectively.

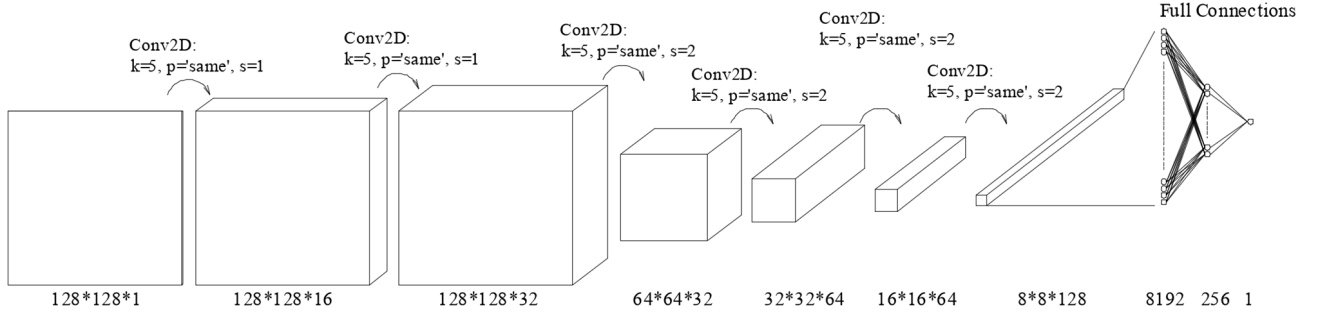

*Figure 14. The CNN structure to learn a regularization functional. Six convolution layers are used before the full connections in the end. For each layer, k means the Kernel Size, p specifies the Padding Pattern, and s stands for the Stride. As we go deeper, the more abstract the features we extract, and the thicker the volume is while it shrinks in length and width. The network outputs a scalar in the end for the 128×128 input image.*

The network is trained on the following Algorithm 1 [8]. As indicated in the loss function (3.77-1), we expect the network to produce a low value to images following the distribution $P_r$ while a high value to ones following $P_n$. Therefore, as a component in the minimization objective, $\Psi_\Theta$ rewards images whose distribution is closer to $P_r$ and punishes images whose distribution is closer to $P_n$. So, by gradient descent on that minimization objective as Algorithm 2 [8] shows, the distribution of reconstruction becomes more and more like $P_r$ and goes away from $P_n$.

---

**Algorithm 1** Learning a regularization functional [8]

---

**Require:** Gradient penalty coefficient $\mu$, batch size $m$, RMS-Prop learning rate $lr$, and pseudo-inverse operator $A_\delta^+$.

  **while** the weight $\Theta$ has not converged **do**
    **for** $i \in 1, 2, \ldots, m$ **do**
      Sample the ground truth image $x_r \sim P_r$, the measurement $y \sim P_Y$, and the random number $\varepsilon \sim U[0, 1]$.
      $x_n \leftarrow A_\delta^+ y$           % *The pseudo-inversed image from the sampled measurement*
      $x_i = \varepsilon x_r + (1-\varepsilon) x_n$  % *The $i^{th}$ random image in the batch, with its pixel distribution located somewhere between $P_r$ and $P_n$*
      $L_i \leftarrow \Psi_\Theta(x_r) - \Psi_\Theta(x_n) + \mu \left( \left\| \nabla_{x_i} \Psi_\Theta(x_i) \right\| - 1 \right)^2$   % *The $i^{th}$ loss function in the batch*
    **end for**
    $\Theta \leftarrow \text{RMSProp}\left( \nabla_\Theta \sum_{i=1}^{m} L_i, lr \right)$     % *Update the weight $\Theta$ based on averaged $L_i$ in a batch*
  **end while**

**Algorithm 2** Applying the learned regularization functional with gradient descent to find the reconstruction [8]

**Require:** Learned regularization functional $\Psi_\Theta$, measurements $y$, regularization parameter $\lambda$, step size $\tau$, operator $A$, pseudo-inverse operator $A_\delta^+$, and stopping criterion $S$.

$x \leftarrow A_\delta^+ y$            *% Start from the pseudo-inversed image*

**while** $S$ not satisfied **do**

$x \leftarrow x - \tau \cdot \nabla_x \left[ \|Ax - y\|_2^2 + \lambda \Psi_\Theta(x) \right]$    *% Gradient Descent to pull the distribution of x from $P_n$ to $P_r$ gradually*

**end while**

**return** $x$

### 3.3.3 Analysis on the results

In Algorithm 1, we set the gradient penalty coefficient $\mu = 20$, the batch size m = 16, and the RMS-Prop learning rate $lr = 0.0001$. We use 217 images for training and 24 images for evaluation, and the data are downloaded from the LIDC/IDRI database [35]. We transform these clear data to produce degraded measurements. The forward operator A corresponds to the Ray Transform in Computed Tomography. The additive noise obeys a Gaussian distribution with zero mean and $0.01^2$ variance, *i.e.*, $e \sim N(0, 0.01^2)$. We provide convergence opportunity to the network by setting 7 epochs with 1000 steps each. And for each step, the algorithm collects samples for a batch by randomly selecting a path pointing at an image in the corresponding folder at each time. The performance of the network in training is tested and recorded for every 100 global steps in global, using the 24 evaluation images. Let us now investigate them in detail.

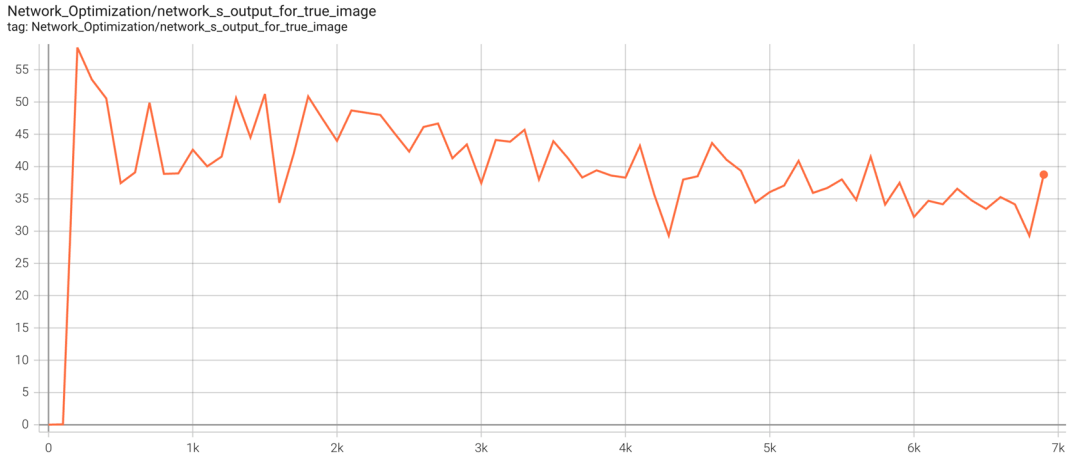

(A) $\Psi_\Theta(x_r)$

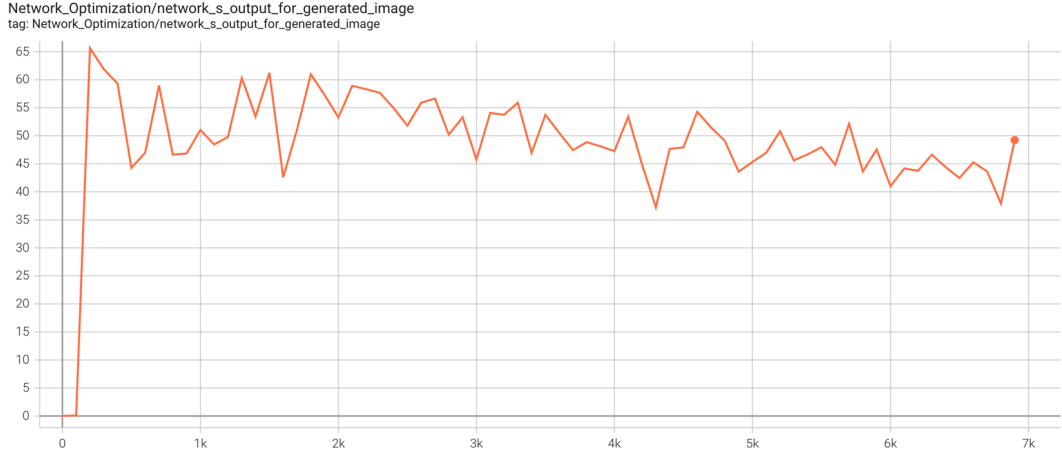

(*B*) $\Psi_\Theta(x_n)$

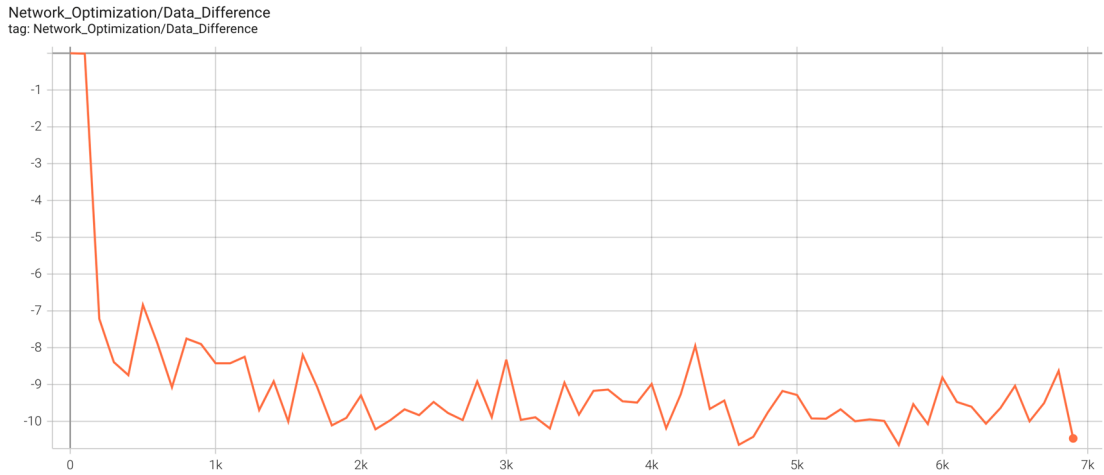

(*C*) $\Psi_\Theta(x_r) - \Psi_\Theta(x_n)$

*Figure 15.* *Performance of the network during training.* (*A*) *Responses of the network to the ground truth images.* (*B*) *Responses of the network to the pseudo-inversed images.* (*C*) *Differences between these two responses.*

As shown above, the network gradually learns to tell apart the ground truth images and the pseudo-inversed ones by producing a low value for the former while a high value for the latter. And the greater the difference between the two, the stronger the network's distinguishing ability. As shown in (C), the network has acquired this ability quickly for the first 1000 steps. After 2000 global steps, its performance has become stable without much room for improvements.

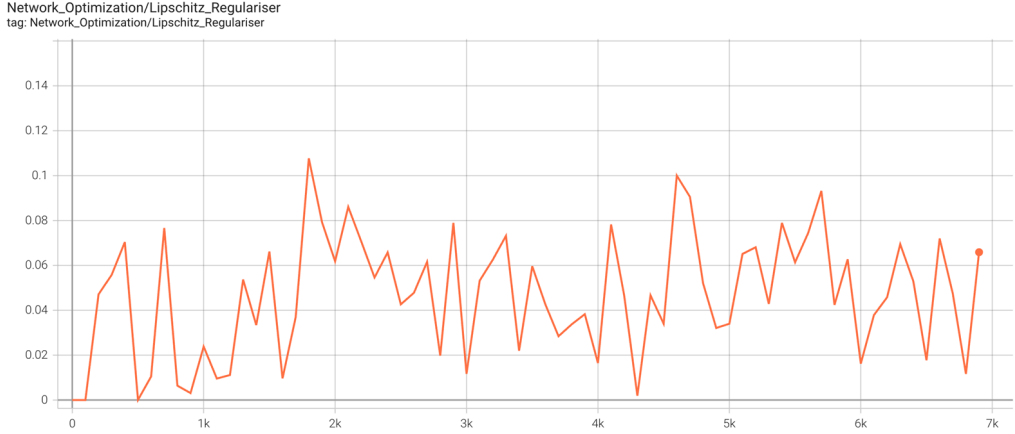

***Figure 16.*** *Evaluation of* $\left(\left\|\nabla_x \Psi_\Theta(x)\right\| - 1\right)^2$ *showing how well the 1-Lip condition is satisfied.*

As shown above, during training the network's gradients with respect to an input image fluctuate around 1 with only a minor variance of no more than 0.12. It suggests that the 1-Lipschitz condition is well satisfied by the network.

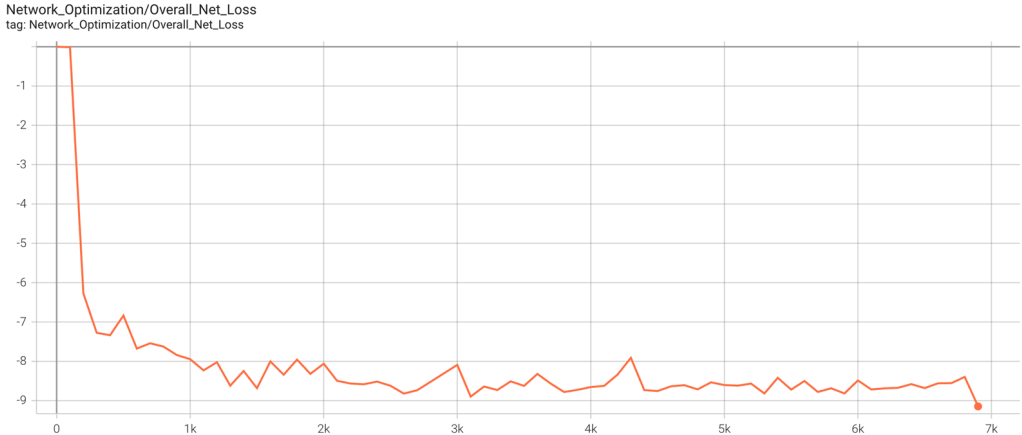

***Figure 17.*** *Variation of the loss function specified in* (*3.77-1*) *with batches of 16 images each.*

The loss function, as indicated by (3.77-1), is the weighted summation of the "Data_Difference" and the "Lipschitz_Regulariser" which we plotted in Figure 15 (C) and Figure 16, respectively. During the initial stage (before 1000 steps) the loss function decreases quickly from 0 to -8, and after 2000 steps it stabilizes at around -8.5. Loss function of this shape suggests a successful training process.

Algorithm 2 works to rectify the pseudo-inversed images by pulling its distribution from $P_n$ to something closer to $P_r$. It is also used to test the network's performance every 100 steps, *i.e.*, for every 100 global steps we use the network to do reconstruction and record its performance. Here we set the regularization parameter $\lambda = 0.3$, the step size $\tau = 0.7$. And we do 1500 steps of gradient decent just to provide enough opportunity for it to converge to the optimal reconstruction $x_{\text{recon}}$. Here we record and analyze three measurements every 100 global steps in training, *i.e.*,

(1) Data_Loss: $\|Ax_{\text{recon}} - y\|_2^2$, which is the residual norm in the optimization objective;

(2) Wasserstein_Loss: $\Psi_\Theta(x_{\text{recon}})$, which is the output of the regularization functional;

(3) L2_to_ground_truth: $\|x_{\text{true}} - x_{\text{recon}}\|_2$, where $x_{\text{true}}$ is the ground truth image $\sim P_r$.

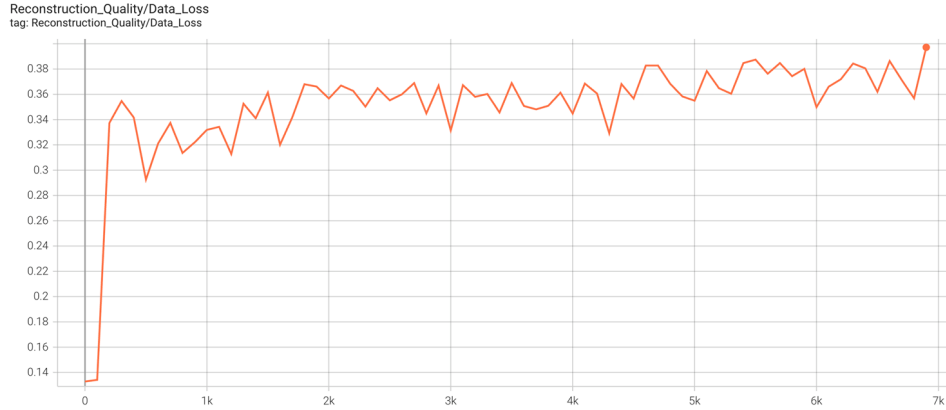

***Figure 18.*** *Variation of the residual norm* (*i.e.*, Data_Loss) *during training.*

It is interesting to observe this variation of the residual norm, as shown above. In the early stage of training, the neural network's ability to discriminate good and bad images is very limited. At this time, the result of reconstruction with such a regularization function will be very close to the pseudo-inversed image to produce an almost zero residual norm. With the improvement of network recognition capabilities, the images recovered with it will deviate more and more from the distribution of pseudo-inversed images. Therefore, the residual norm is increasing. Good news is that residual norm stabilizes at a small value of 0.37 in the end of training. It means that our reconstruction does not deviate far from the observation, and the result is convergent.

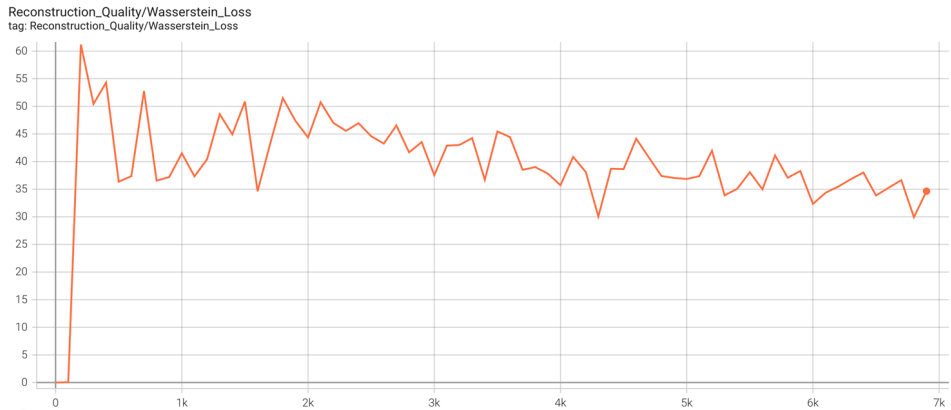

***Figure 19.*** *Variation of the outputs of the regularization functional* (*i.e.,* Wasserstein_Loss) *during training.*

As shown above, the outputs of the network to the reconstructed images grow smaller as training goes. In the middle and late stages of training, the outputs of the network to the reconstructed images fluctuate roughly in the range of 35 ~ 45; looking back at Figure 15 (A), this is exactly the range of the network's outputs for the ground truth images. Comparatively, the network gives outputs of 45 ~ 55 for pseudo-inverse images, as Figure 15 (B) shows. This means that the learned regularization function can rectify the distribution of the pseudo-inverse images, making it closer to the distribution of the real images, and thus produce more real-like reconstructed images.

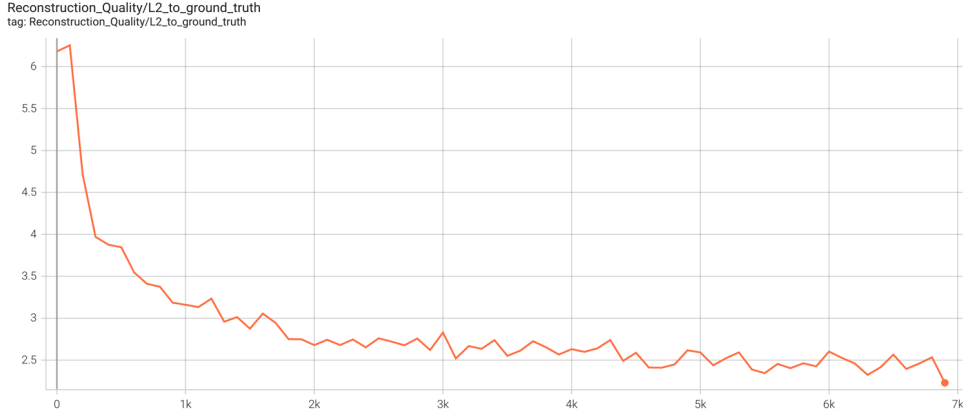

***Figure 20.*** *Variation of the $l_2$-norm (i.e., L2_to_ground_truth) to the ground truth image during training.*

This indicator measures the degree to which the reconstructed image is close to the ground truth image. As training goes, the reconstructed image gets closer and closer to its ground truth.

The regularization parameter $\lambda$ in the optimization objective is a hyperparameter not learnable from data. The paper provides a theoretical model to estimate it, depending on the noise level. Following the notation of (3.75) we want to minimize

$$g(x) = \|Ax - y\|_2^2 + \lambda \Psi_\Theta(x). \tag{3.78}$$

Assume that the solution $x$ lies on the critical point of this function, then the derivative of it with respect to $x$ should be zero, *i.e.*,

$$\begin{aligned}\frac{\partial g(x)}{\partial x} &= \frac{\partial \|Ax-y\|_2^2}{\partial x} + \frac{\partial(\lambda\Psi_\Theta(x))}{\partial x} \\ &= \frac{\partial(Ax-y)}{\partial x} \cdot \frac{\partial \|Ax-y\|_2^2}{\partial(Ax-y)} + \lambda \nabla_x \Psi_\Theta(x) \\ &= A^\mathrm{T} \cdot 2(Ax-y) + \lambda \nabla_x \Psi_\Theta(x) \\ &= -2A^\mathrm{T}e + \lambda \nabla_x \Psi_\Theta(x) \triangleq 0.\end{aligned} \tag{3.79}$$

Then

$$\lambda \nabla_x \Psi_\Theta(x) = 2A^\mathrm{T}e \;\;\Rightarrow\;\; \|\lambda \nabla_x \Psi_\Theta(x)\|_2 = 2\|A^\mathrm{T}e\|_2 \;\;\Rightarrow\;\; |\lambda| = \frac{2\|A^\mathrm{T}e\|_2}{\|\nabla_x \Psi_\Theta(x)\|_2}. \tag{3.80}$$

If the regularization functional has gradients of unit norm [36], then we can estimate $\lambda$ by

$$\hat{\lambda} = 2\mathrm{E}_{e \sim p_n} \|A^\mathrm{T}e\|_2. \tag{3.81}$$

In their codes, however, they did not use this method and the value of λ seems to be determined by trial and error. Here we demonstrate a series of experimental results on different λ. The reconstruction qualities are assessed on the PSNR (Peak Signal to Noise Ratio) and the SSIM (Structural Similarity Index), as shown below in Table 3.

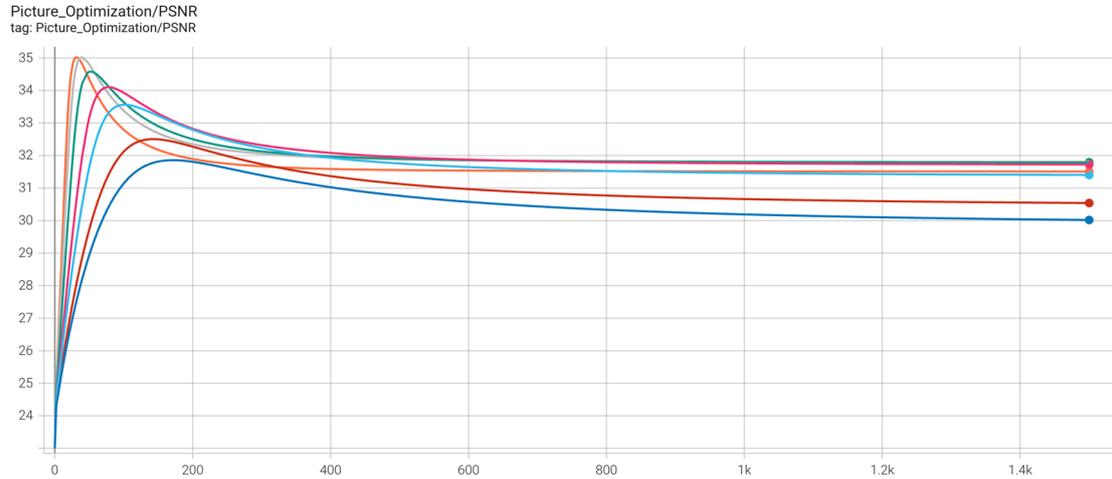

**Figure 21.** *Plots of PSNR against Gradient Descent steps for different regularization parameter λ.*

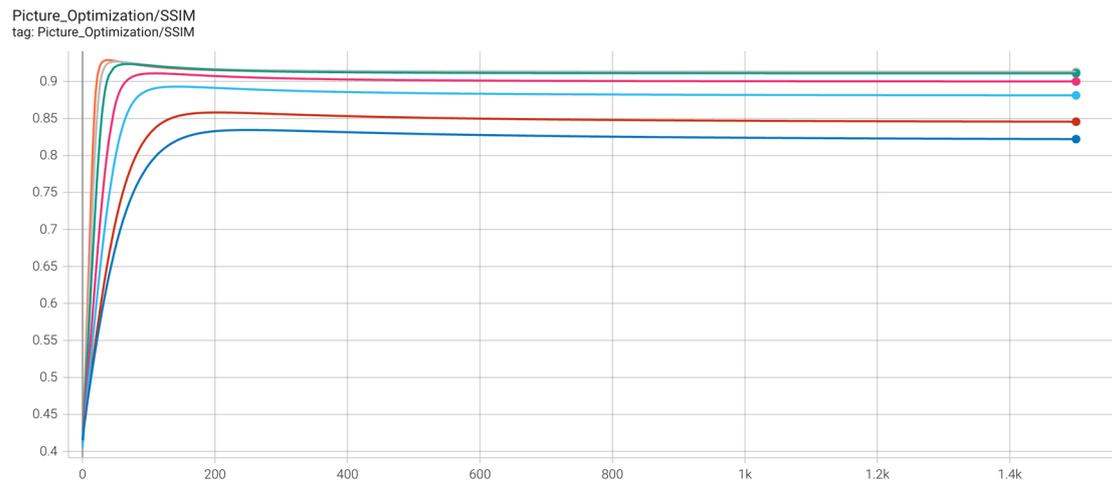

**Figure 22.** *Plots of SSIM against Gradient Descent steps for different regularization parameter λ.*

*Table 3. PSNR and SSIM values of Figure 21 and 22 at Gradient Descent step 1400.*

| Step Size $\tau$ | Regularization Parameter $\lambda$ | PSNR / dB | SSIM |
|---|---|---|---|
| 0.7 | 0.08 | 30.04 | 0.8224 |
| 0.7 | 0.10 | 30.56 | 0.8458 |
| 0.7 | 0.15 | 31.41 | 0.8813 |
| 0.7 | 0.20 | 31.73 | 0.9001 |
| 0.7 | 0.30 | 31.79 | 0.911 |
| 0.7 | 0.40 | **31.81** | **0.9131** |
| 0.7 | 0.50 | 31.51 | 0.9125 |

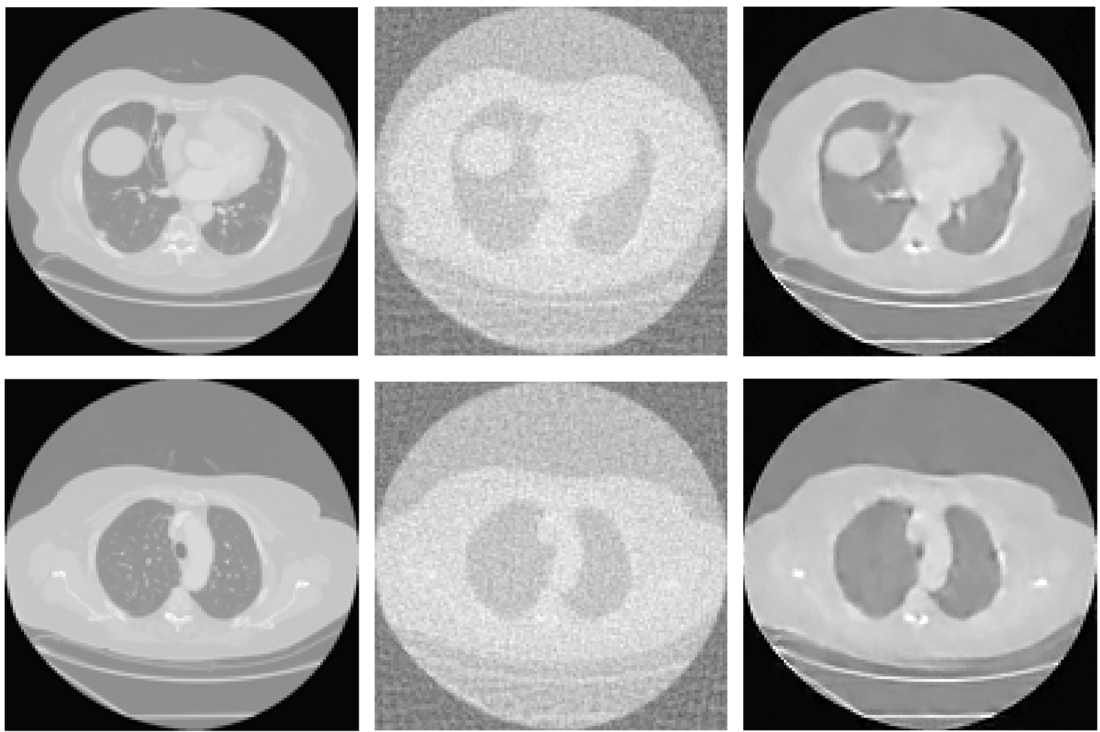

**Figure 23.** *Reconstruction results for λ =0.4 (top row) and λ =0.3 (bottom row). The left column are the ground truths, the middle column are the pseudo-inversed images, and the right columns are reconstructions. The data are drawn from the LIDC/IDRI database* [35].

**3.4 Comparisons of regularization-based methods and network-based methods**

In this section we give a brief account of the advantages and limitations of the two approaches – the regularization-based methods and the network-based methods – of we discussed till now.

The advantage of classical (without network) regularization methods lies in the clarity we have when we design everything in person. In spectral regularization methods, by spectral decomposition, we proved the effectiveness of filtering methods to deal with this inverse problem. We also demonstrated that reconstructions by the Tikhonov Filters follow exactly the form of the solution of the minimization objective in the Tikhonov Regularization problem. So, the problem of deblurring can be generalized as minimizing an objective containing a residual norm and a regularization term. By crafting the regularization term by hand, we experienced how to design a regularization functional corresponding to our prior knowledge and understood why this model can make reconstructions more in line with the prior information. All these works serve as fundamentals and help us to understand the image deblurring problem better.

The limitations of classical regularization methods are also obvious. First, non-linearity may present in the blurring mechanism, and for regularization to go we need to model explicitly the non-linearity. It can be a hard job. Second, the hand-crafted regularization term is always simplified to favor numerical computations. For example, in Section 3.2 the "sparse edge assumption" was modeled exponentially by a few parameters. Such a simplified model is not able to capture the great complexity in real life. And sometimes due to the complexity no analytical expression can be made. Third, in the optimization objective, there is a regularization parameter that must be determined but it relies on information we do not know in advance, *e.g.*, the noise in the data, the amount of "data misfit" represented in the residual norm, *etc*.

The advantages of network-based approaches compensate for the shortcomings of regularization-based methods. Neural network is designed to learn a complex mapping relationship between the input and the output. As we saw in Section 3.3, we let the network to learn a complex regularization functional, on which the distribution of the pseudo-inversed images gets rectified so the reconstruction can approximate the distribution of the ground truth image. And sometimes instead of outputting a scalar, the network can be designed to output an image which is directly what we want. So, the deblurring job is as simple as feeding a blurry image to the network and waiting for a sharp output. We do not have to design deliberately in advance the regularization parameter. Instead, we treat it as an alterable parameter subject to the real-time monitoring of the various indicators of the network in training.

The shortcomings of neural network-based methods mainly come from its "black box" attributes. Parameters are sometimes determined by trial and error without much room for explanation. Another concern is that the weights learned on one dataset may not be applied, with good quality, to other datasets. So sometimes we must sacrifice some accuracy on one dataset to avoid overfitting.

# Chapter 4

# Introduction to the DeblurGAN-v2 Method [14]

In this chapter, we introduce the DeblurGAN-v2 method. And in the next chapter, we analyze it on numerical tests and whereby find some improvements to it. As a method treating the blind image deblurring (BID) as a special case of image-to-image translation problem, DeblurGAN-v2 is based on the idea of Generative Adversarial Network (GAN). As an updated version of DeblurGAN, it incorporates the Feature Pyramid Network (FPN) framework for the generator and adopts the Relative Least Square (Ra-LS) adversarial loss for the Double-Scale discriminator.

**4.1 The Generative Adversarial Network (GAN)**

The idea of the Generative Adversarial Network was proposed by Ian Goodfellow *et al.* [10] in 2014. The motivation was to generate data close to the distribution of the truth but without any need to approximate the intractable density function hidden behind. In this sense, it can be categorized into the so-called *implicit generative models* that only learn a tractable data generation process.

GAN involves a generator and a discriminator. The discriminator, as a binary classifier, works to distinguish the generated data from the truth. The generator, working as an adversary to the discriminator, aims to produce data that are indistinguishable from the real. In terms of their respective goals, they are opponents. But during training they are also in a cooperative relationship in the sense that the generator is updated on information related to the discriminator. The generator gets rewards if it generates a datum that manages to fool the discriminator. The discriminator gets improved if it can correctly distinguish images of the true dataset from images of a constantly changing generated dataset. During the game, it is good if the two can tune their improvements and keep balanced. Otherwise, if the discriminator improves too fast, the generator may never catch up; if the discriminator improves too slow, the generator will be rewarded for constantly generating data of poor quality. If such a "consistent progress" is maintained, then in the end, the image generated by the generator will be very close to the target image. Because the generator has managed to survive the fierce competition with the discriminator.

**4.2 The Adversarial Loss Function**

For neural networks, the game between the generator and the discriminator can be represented by moving to or away from the minimum of an adversarial loss function. This cost can be designed by several ways. One version is the Min-Max objective:

$$\min_{G} \max_{D} \left\{ \mathrm{E}_{x \sim \mathrm{P}_r} \left[ \log(\mathrm{D}(x)) \right] + \mathrm{E}_{\hat{x} \sim \mathrm{P}_g} \left[ \log(1 - \mathrm{D}(\hat{x})) \right] \right\}, \text{ where } \begin{cases} \text{label "1" for real data } x \sim \mathrm{P}_r, \\ \text{label "0" for generated data } \hat{x} \sim \mathrm{P}_g. \end{cases} \quad (4.1)$$

On the discriminator's side, efforts are made to assign as more "1"s to the real data and more "0"s to the generated data as possible whereby to maximize the function in the curly brace. On the generator's side, it produces as more "real-like" images as possible whereby to fool the discriminator and get that function minimized. Although tractable for theoretical analysis, (4.1) would meet many problems in training, *e.g.*, model collapse, gradient vanishing/explosion, *etc.* [37] As an enhancement of (4.1), Mao [38] proposed an alternative version

$$\begin{cases} \min_{D} \mathrm{V}(\mathrm{D}) = \frac{1}{2} \mathrm{E}_{x \sim \mathrm{P}_r} \left[ (\mathrm{D}(x) - 1)^2 \right] + \frac{1}{2} \mathrm{E}_{\hat{x} \sim \mathrm{P}_g} \left[ (\mathrm{D}(\hat{x}))^2 \right] \\ \min_{G} \mathrm{V}(\mathrm{G}) = \frac{1}{2} \mathrm{E}_{\hat{x} \sim \mathrm{P}_g} \left[ (\mathrm{D}(\hat{x}) - 1)^2 \right] \end{cases} \quad (4.2)$$

which adopts the Least Square Loss instead. It was proved to produce smoother gradients without saturation, and training can be more stable compared to the case before. In the DeblurGAN-v2 method, the author followed this idea but made a few improvements to make a Relative Least Square Loss function as

$$\begin{cases} L_{\mathrm{D}}^{(\text{adversarial})} = \frac{1}{2} \left\{ \mathrm{E}_{x \sim \mathrm{P}_r} \left[ \left( \mathrm{D}(x) - \mathrm{E}_{\hat{x} \sim \mathrm{P}_g} \left[ \mathrm{D}(\hat{x}) \right] - 1 \right)^2 \right] + \mathrm{E}_{\hat{x} \sim \mathrm{P}_g} \left[ \left( \mathrm{D}(\hat{x}) - \mathrm{E}_{x \sim \mathrm{P}_r} \left[ \mathrm{D}(x) \right] + 1 \right)^2 \right] \right\} \\ L_{\mathrm{G}}^{(\text{adversarial})} = \frac{1}{2} \left\{ \mathrm{E}_{x \sim \mathrm{P}_r} \left[ \left( \mathrm{D}(x) - \mathrm{E}_{\hat{x} \sim \mathrm{P}_g} \left[ \mathrm{D}(\hat{x}) \right] + 1 \right)^2 \right] + \mathrm{E}_{\hat{x} \sim \mathrm{P}_g} \left[ \left( \mathrm{D}(\hat{x}) - \mathrm{E}_{x \sim \mathrm{P}_r} \left[ \mathrm{D}(x) \right] - 1 \right)^2 \right] \right\} \end{cases}$$
(4.3)

where the generator's cost is designed by "flipping" the discriminator's labels, *i.e.*, label "1" for generated data while label "0" for real data. Compared to (4.2), this Relative Square Loss function does not set specific goals (*i.e.*, 0s or 1s) for the discriminator to achieve as the indication of convergence. As long as there is enough distance between the outputs of the discriminator with respect to the two kinds of data, we say that it is able to distinguish them. In this gentle way it is much easier to train the network. The coefficient (1/2) may be set for the convenience of calculating the derivative during back-propagation. And by putting equal weights (*i.e.*, (1/2)) on the two mathematical expectations – with respect to the real data and the generated data, respectively – we have no bias towards either one of them. In other words, this metric ensures that there are equal numbers of samples from the real data and the generated data for us to evaluate the performance of the network fairly.

## 4.3 Structure of the Generator

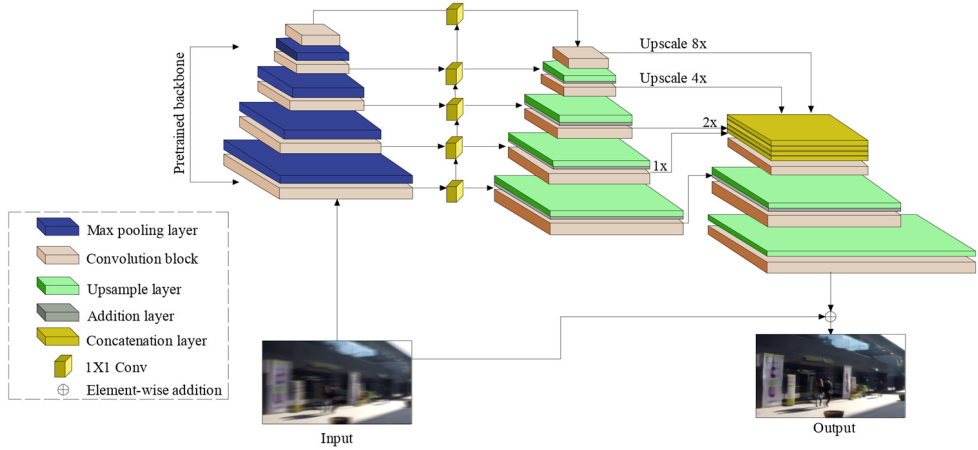

*Figure 24. The Feature Pyramid Network with Inception-ResNet-v2 backbone for the generator* [14].

The above structure may be formidable to read at the first glance, but we will find that it is a highly regular architecture if we manage to understand the role of each of its component parts.

As illustrated, the input is a blurry observation, and the output is the deblurred result. It is worth noting that the observation acts not only as the input to the generator, but also as the adding component in the end to yield the reconstruction. In this sense, the network only needs to learn the residual information – the mapping between the blurry and the sharp – rather than the whole things contained in the input. It is a necessary design. A thousand images may have a thousand types of content but the meaningful learning materials for the network may only be the essence of one type of blur, *e.g.*, the motion blur or the Gaussian blur.

To learn this essence of blur the author involves a Feature Pyramid Network (FPN) framework with the Inception-ResNet-v2 backbone. The left pyramid is a 5-layer CNN to extract features of the input. Similar to the structure we observed in Figure 14, the deeper we go, the thicker the feature map is while it shrinks in length and width as a result of Max-Pooling. These feature maps obtained are then used for Image-to-Image Translations, *i.e.*, to produce a high resolution deblurred image in this case. Some previous works [39] adopts the Encoder-Decoder structure as shown in Figure 25 (A).

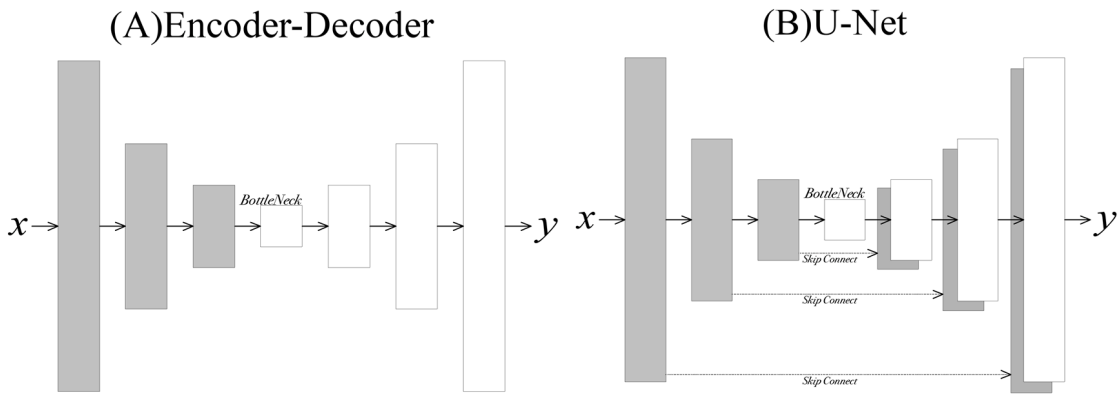

*Figure 25. Encoder-Decoder structure and the U-Net* [12] *structure. x is original image and y is the translated image.*

The problem is that the high-level information in the small bottleneck is not able to capture the important low-level information, which is shared between *x* and *y*. So, if we only use the information included in the bottleneck then the translation will lose a significant amount of information, *e.g.*, the common underlying structure of *x* and *y*. To address it, we can send low-level features extracted in maps before the bottleneck to corresponding layers in the decoder. This idea was proposed by [12] as the U-net shown in Figure 25 (B).

The DeblurGAN-v2 method adopts this idea of skip connection. As shown in the middle pyramid of Figure 24, a new feature map is obtained by convolution on the sum of the previous feature map up sampled and the feature map of the same level in the left pyramid. In this way, we manage to retain the underlying low-level features when building new feature maps. The top four feature maps are unified to the same scale and concatenated together, in the right pyramid, to form a new feature map, which is then up sampled and added with the largest feature map in the middle pyramid. The result then goes through a convolution layer and gets up sampled. The final convolution produces a feature map of the same number of channels as the input, *i.e.*, 3 in RGB images.

In summary, there are two highlights in this Feature Pyramid Network – the horizontal connection of feature maps of the same scale, as well as the vertical concatenation of feature maps of multiple scales. The former guarantees that low-level information is retained in the translation, and the latter ensures that all feature maps extracted, corresponding to multiple levels, are all used for reconstruction.

## 4.4 Structure of the Double-Scale Discriminator

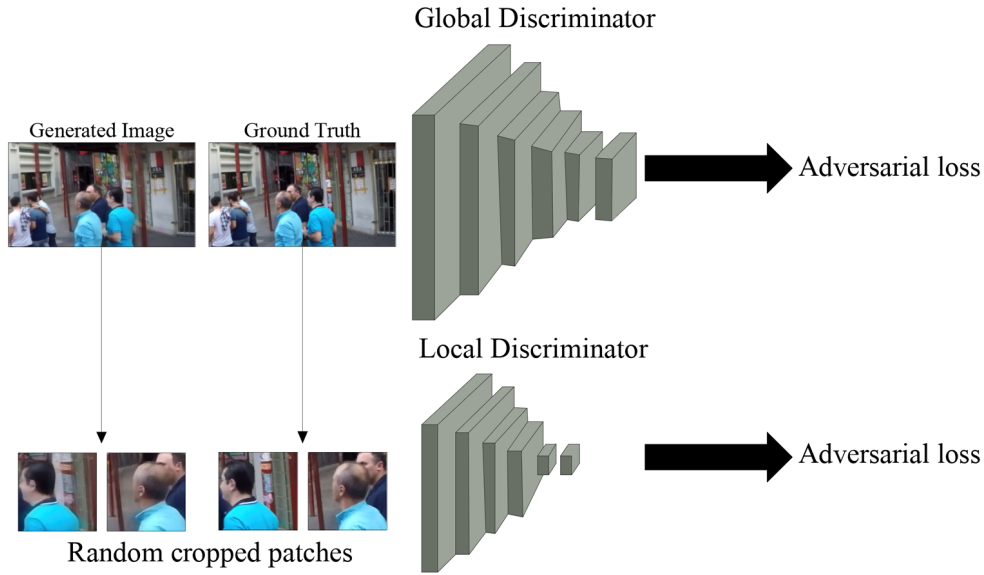

*Figure 26. The Double-Scale Discriminator structure* [14].

Compared with the pyramid-like architecture for the generator as shown in Figure 25, the structure for the discriminator is much simpler in that its task is much easier, *i.e.*, distinguishing the generated images from the ground truth images. For this job, plain CNN architecture is enough. A highlight here is that there are two discriminators working on the global scale and the local scale simultaneously. The former evaluates the two images from a global perspective while the latter focuses on localized details. It is worth noting that because the patches contain less information than the global images, for the purpose of saving computing resources, the local discriminator may contain fewer CNN layers than the global discriminator. In the code they used 5 layers for the global discriminator while 3 layers for the local discriminator.

Another comment on this discriminator is that due to its much simpler architecture compared with that of the generator, it contains fewer parameters than the generator has. It means that the discriminator is easier to train so it may keep dominance over the generator during the game. We will discuss this further in Section 5.3.1 and Section 5.3.2.1.

# Chapter 5

# Numerical Tests and potential improvements

Based on numerical tests, in this chapter we analyze the DeblurGAN-v2 and whereby suggest some improvements to it. We train, validate, and test algorithms on datasets arising from the GoPro [15], DVD [16], and NFS [17], which were designed to deal with motion blur initially. As suggested by Section 2.3, here we deal with the Gaussian blurs and the motion blurs.

**5.1 Performance on the Consistent Gaussian Blur**

For such a sophisticated algorithm designed to handle the complex motion blurs, we may wonder first whether it is able to deal with a much simpler case with high quality. Here we consider a consist systematic Gaussian blur, which is doubly symmetric with kernel (PSF) size 13 and standard deviation

$$\sigma = 0.3 \times \left(\left(kernel\_size - 1\right) \times 0.5 - 1\right) + 0.8 = 2.3, \quad \text{for } kernel\_size = 13 \text{ here.}$$

(5.1)

Example images before and after this Gaussian PSF are

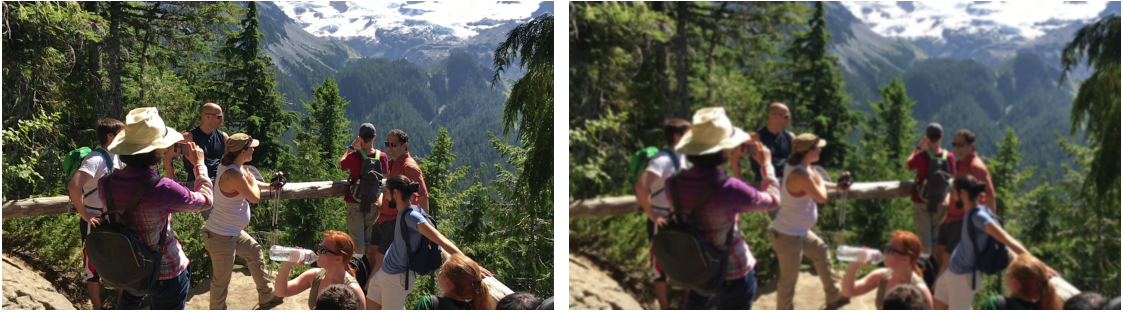

*Figure 27. Example images before and after this Gaussian blur. The left clear image is drawn from the DVD [16] training dataset.*

We use a merged dataset, by picking every second picture from the DVD and GoPro datasets and every tenth picture from the NFS dataset, to train the network. The merged dataset is divided into the Train folder and the Test folder. In the Train folder, there are over five thousand pairs of <blur images, sharp images>. While in the Test folder, the amount is 1990. We use the first 87% images in the Train folder to train the network, and the last 13% for validation. Images in the Test folder are only used to test the performance of the trained network, through metrics like PSNR and SSIM. For training and validation, we use several indicators to reflect the real-time performance of the network. Let us use D to denote the discriminator and G to denote the generator. These metrics are:
➢ The Adversarial Loss of D.

➢ The Total Loss of G, which is the weighted sum of the Perceptual Loss, the Content Loss, and the Adversarial Loss.
➢ The PSNR (Peak Signal-to-Noise ratio) between the generated image and the ground truth image.
➢ The SSIM (Structure Similarity Index) between the generated image and the ground truth image.

The following Figure 28 plots the variance of them.

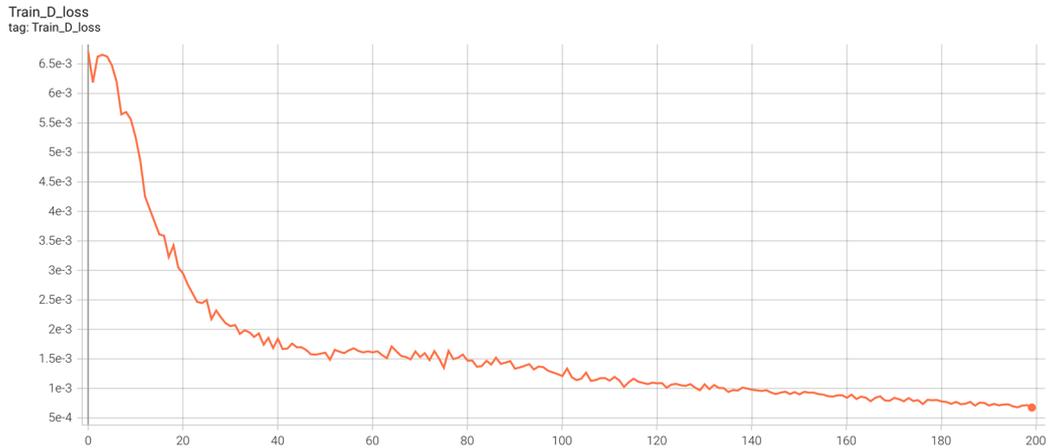

(*A*) *The adversarial loss of D in training.*

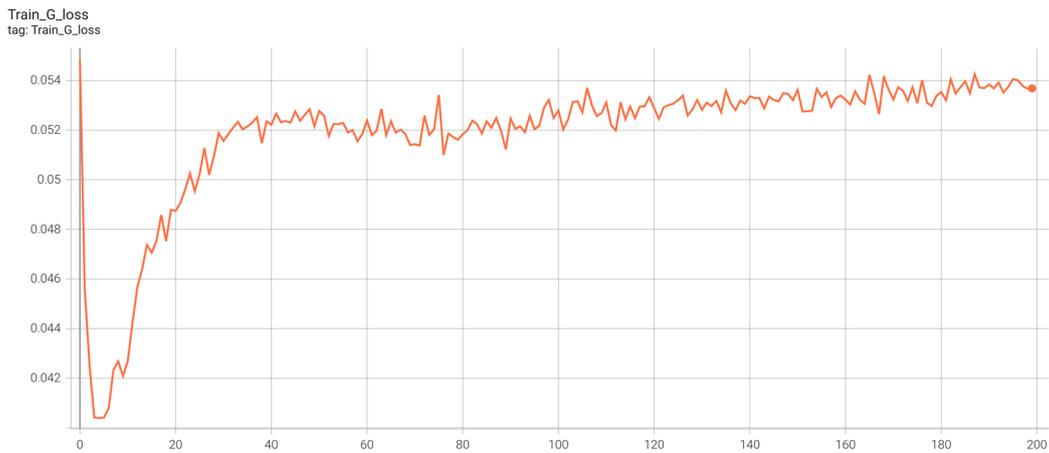

(*B*) *The total loss of G in training.*

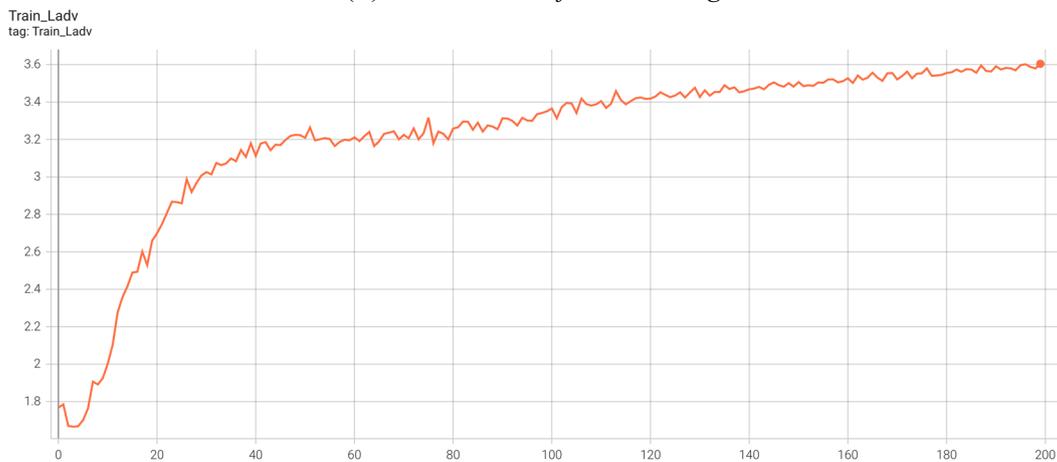

(*C*) *The adversarial loss of G in training.*

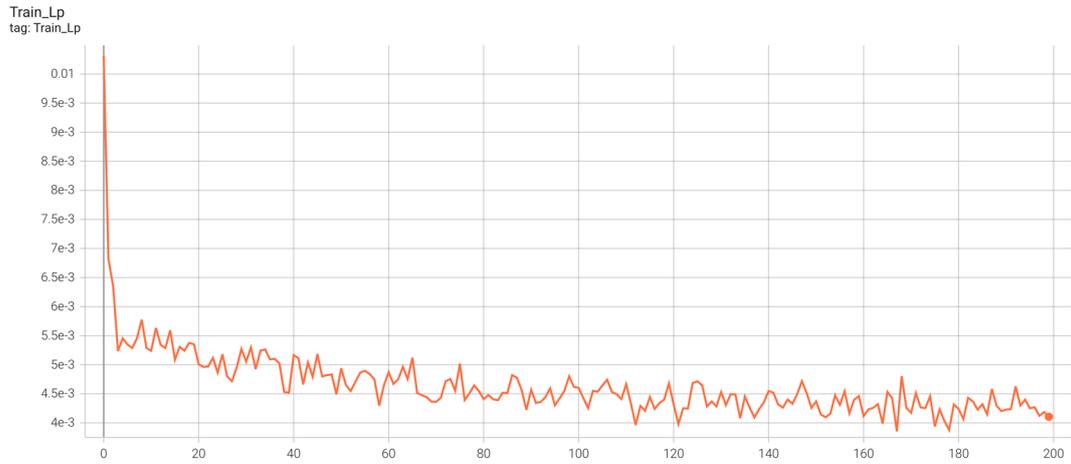

(*D*) *The perceptual loss $L_p$ of G in training with mean 0.0046 for the whole 200 epochs.*

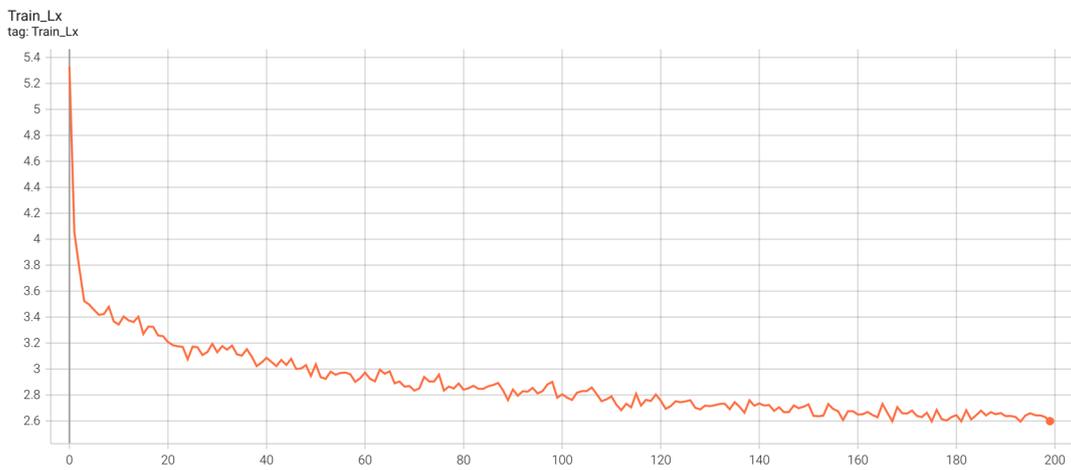

(*E*) *The content loss $L_x$ of G in training with mean 2.8886 for the whole 200 epochs.*

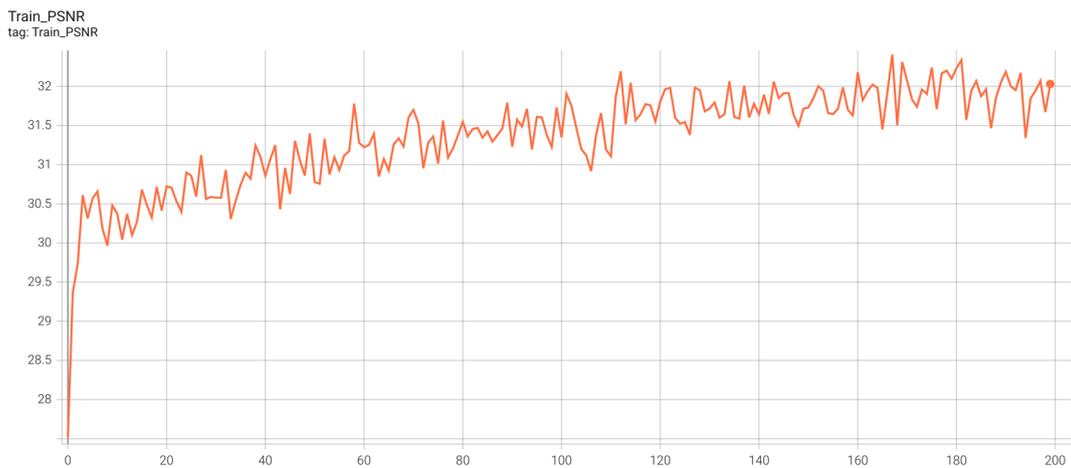

(*F*) *The PSNR between the generated image and the ground truth image in training.*

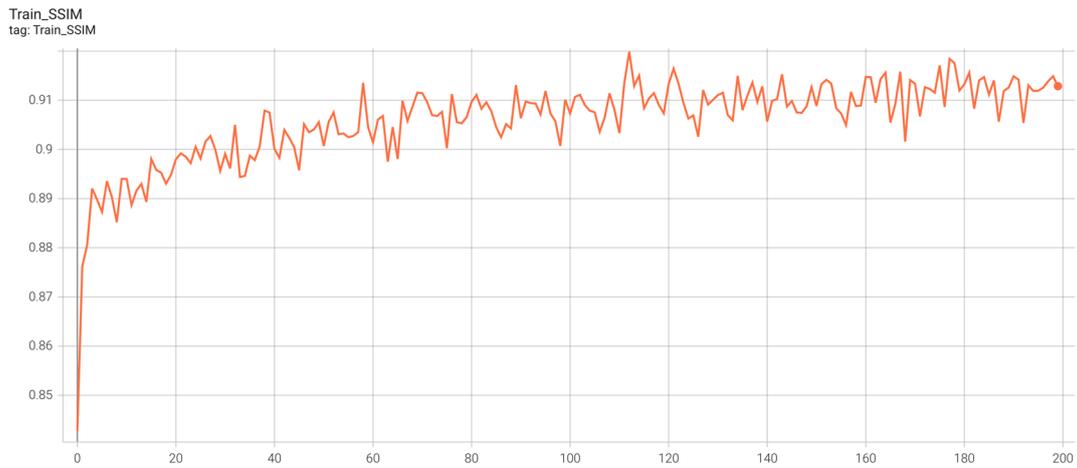

(*G*) *The SSIM between the generated image and the ground truth image in training.*

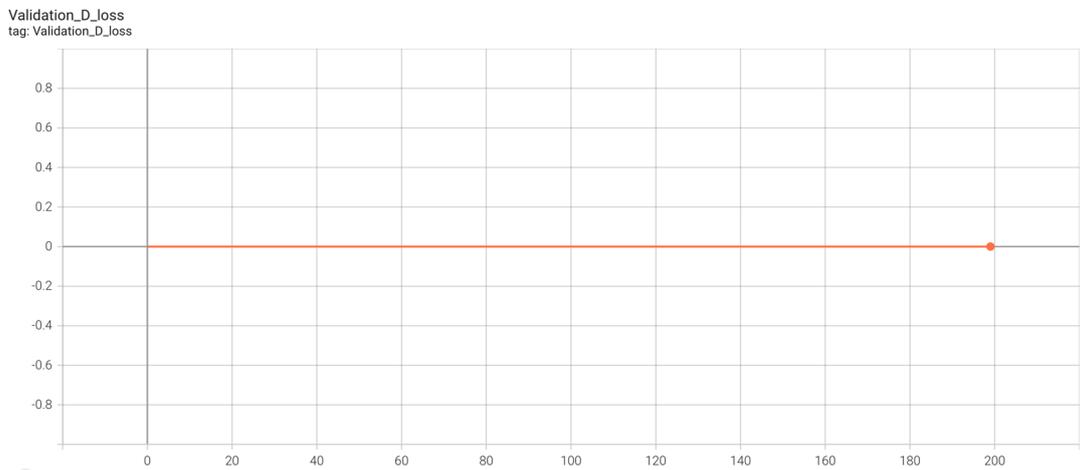

(*H*) *The adversarial loss of D in validation.*

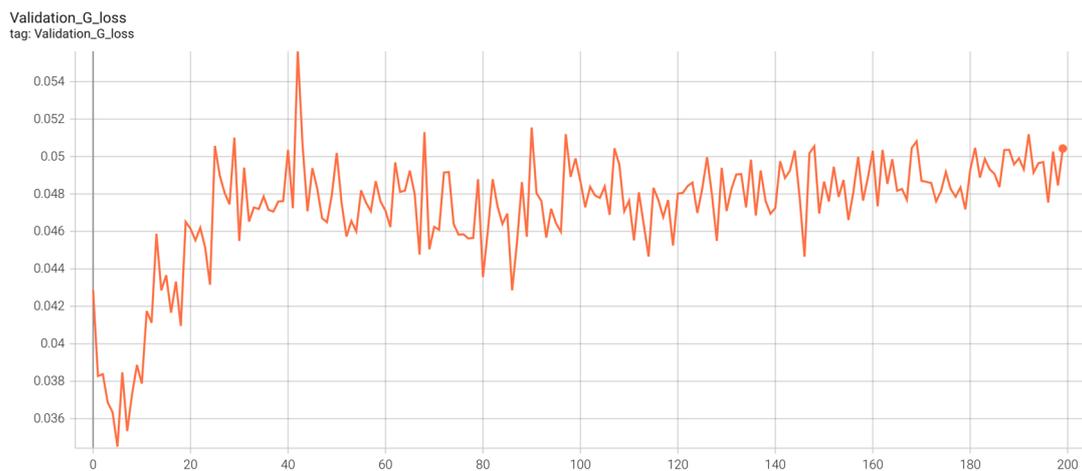

(*I*) *The total loss of G in validation.*

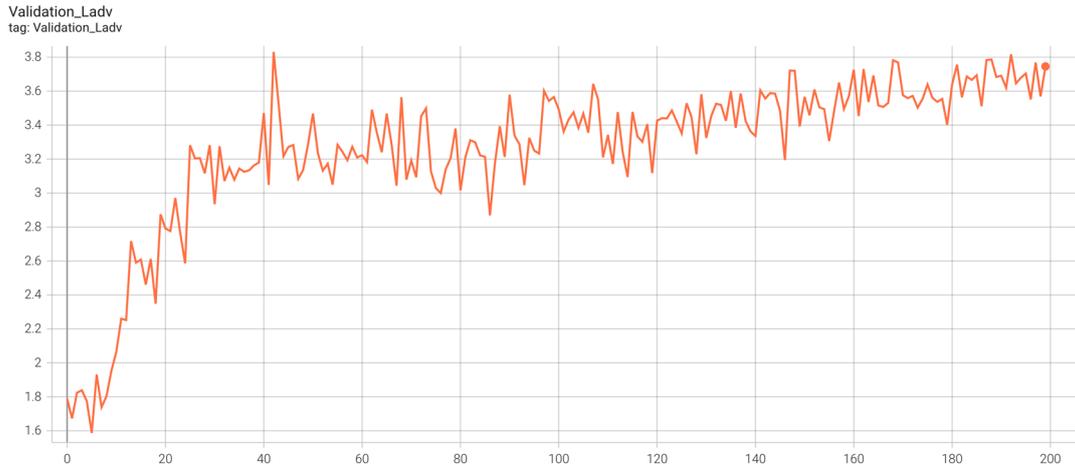

(*J*) *The adversarial loss of G in validation.*

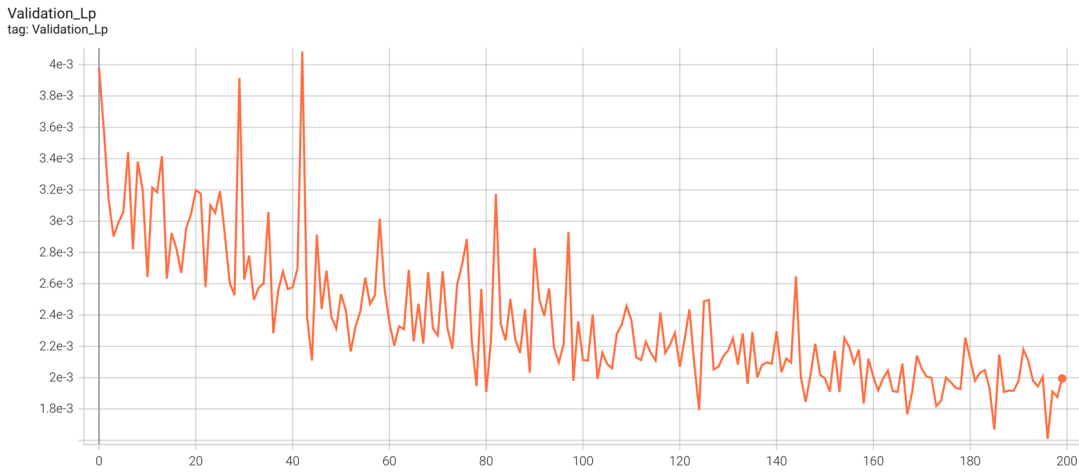

(*K*) *The perceptual loss $L_p$ of G in validation.*

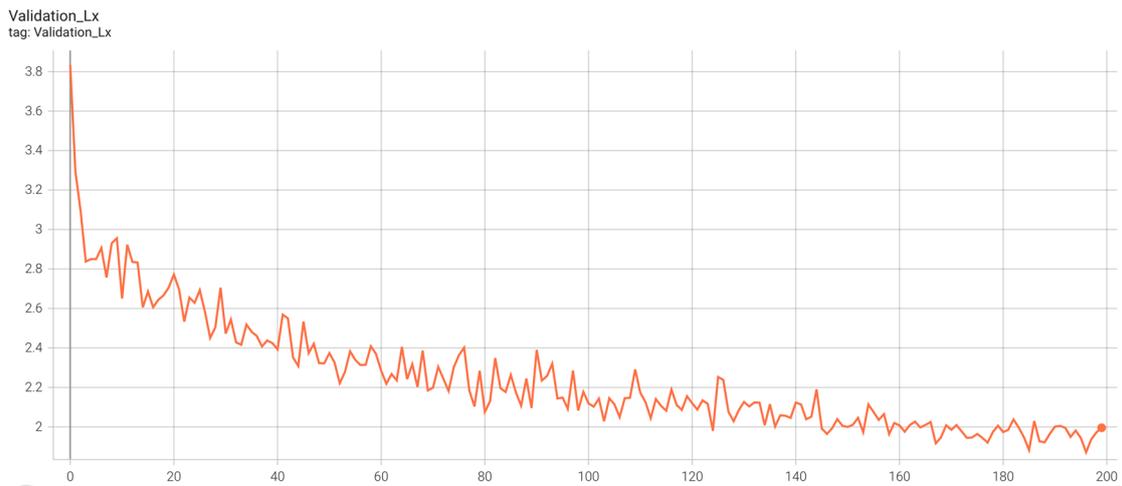

(*L*) *The content loss $L_x$ of G in validation.*

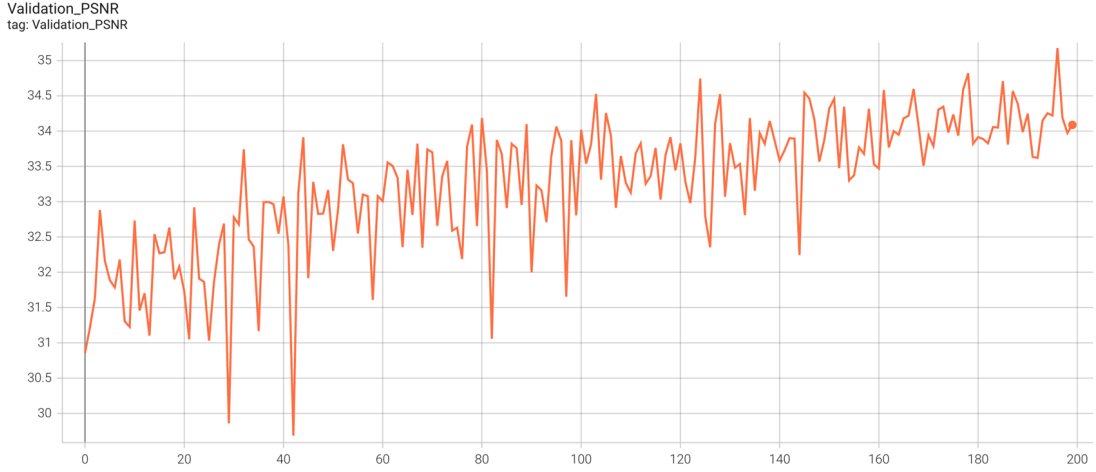

(*M*) *The PSNR between the generated image and the ground truth image in validation.*

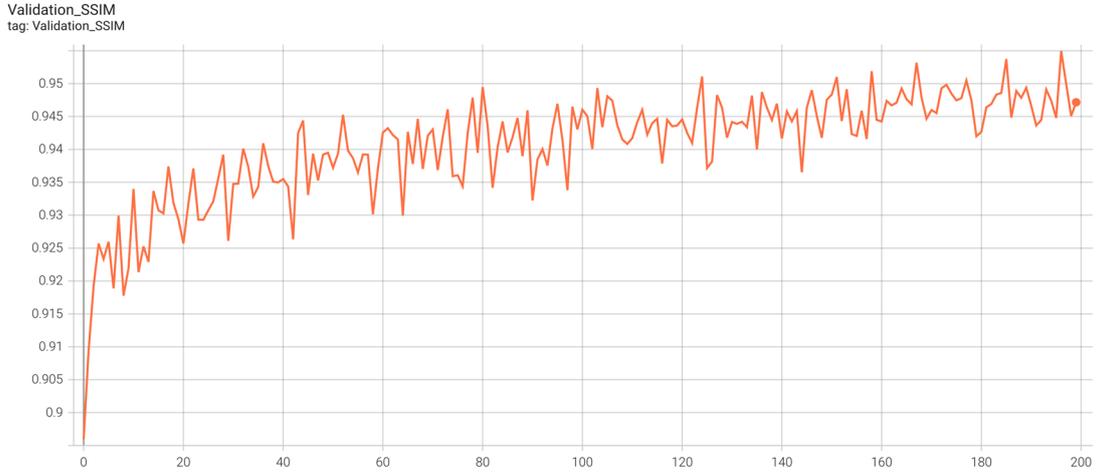

(*N*) *The SSIM between the generated image and the ground truth image in validation.*
**Figure 28.** *Metrics recorded in one experiment for the case of consistent Gaussian blur.*

The perceptual loss $L_p$ is the Minimum Square Error (MSE) between the generated image and the ground truth image, and the content loss $L_x$ is the MSE distance between the 14$^{th}$ feature maps on the VGG-19 architecture for the two kinds of images. As suggested by [40], if we use the global $L_p$ as the only metric then the reconstruction will be oversmoothed. To preserve details, $L_x$ is used as a complementary to $L_p$. The total loss of G is, by the author, the weighted sum of $L_p$, $L_x$, and its adversarial loss specified in (4.3), *i.e.*,

$$L_G = \lambda_1 \cdot L_p + \lambda_2 \cdot L_X + \lambda_3 \cdot L_G^{(\text{adversarial})}, \text{ where parameters were choosen to be } \begin{cases} \lambda_1 = 0.5 \\ \lambda_2 = 0.006 \\ \lambda_3 = 0.01 \end{cases}.$$

(5.2)

The weight $\lambda_1$ on $L_p$ was chosen to be 83 times as large as the weight $\lambda_2$ on $L_x$. We observe from Figure 28 (D) and (E) that the mean of $L_p$ on the whole 200 epochs is 0.0046, while the mean of $L_x$ is 2.8886. With such a huge difference it is reasonable to balance them in a loss function by setting the weights $\lambda_1$ and $\lambda_2$ in a sense as (5.2) suggests.

It is interesting to observe the behaviors of D and G during training through Figure 28. By comparing (A) and (C) we observe a "mirror symmetry" of the developing trend of their adversarial losses. It is understandable because, as (4.3) suggests, their adversarial losses are of exactly the same form except for the labels, *i.e.*, G's adversarial loss is designed by flipping D's labels. So, in the game, the degree of superiority of D reflects how much G falls behind, and vice versa. As observed from (A), D's adversarial loss declines on the whole without large fluctuations, which means that D has been in the leading position during the game. It seems a bad thing for G because speaking from its side it hardly succeeded in deceiving D during the process.

But does this mean that G was completely defeated in the game and did not make any achievements? Not necessarily. From Figure 28 (D) and (E) we observe that $L_p$ and $L_x$, the two metrics measuring the closeness between the generated image and the ground truth image, are all decreasing with a good trend. It means during the game the G has gradually learned how to approximate the real image via the competition with D. From (B) we also see that after 40 epochs there are fewer increments to $L_G$, which is a weighted sum of (C), (D), and (E). This is because in the later stages of training, the update rates of $L_p$, $L_x$ and $L_{adv}$ are all slowed down.

Figure 28 (F) and (G) record the improvements of PSNR and SSIM during training. (I) to (N) record the real-time performance of the network in training on the validation dataset. It is acceptable to see more fluctuations but a consistent trend there because the network is not trained upon this dataset. After the network runs through its 200 epochs, we test its performance on the test images from the merged dataset we made. It contains images sampled from the DVD, GoPro, and the NFS datasets. The results are summarized in Table 4 below.

*Table 4.* Performance of the trained network on different datasets, in the case of consistent Gaussian blur. The PSNR and SSIM are the averaged results from test images of corresponding dataset.

| Dataset | PSNR / dB | SSIM | *Weight File |
|---|---|---|---|
| Sampled DVD | 30.64661 | 0.92519 | 'best' |
| (746 images) | 30.63274 | 0.92320 | 'last' |
| Sampled GoPro | 33.58407 | 0.95036 | 'best' |
| (1096 images) | 33.88977 | 0.95203 | 'last' |
| Sampled NFS | 33.80677 | 0.94277 | 'best' |
| (148 images) | 34.20944 | 0.94519 | 'last' |
| Merged of above | 32.47762 | 0.94024 | 'best' |
| (1990 images) | 32.68526 | 0.94071 | 'last' |

*\*Weight File*: During training, the algorithm saves the "best weights" in real-time based on the validation results. So, together with the "last weights" which correspond to the latest update in the final epoch, there is a "best weights" file to record the network's best weights on the validation datasets.

The following Figure 29 uses three examples to demonstrate the effectiveness of the network to deal with this consistent Gaussian blur specified in (5.1), on the "best weights". The perceptual quality, from either the global or the local perspective, is good. A slightly unsatisfactory thing is that the reconstruction sometimes contains "artificial edges", as can be seen from the right side of (B) and (H), and the bottom of (E).

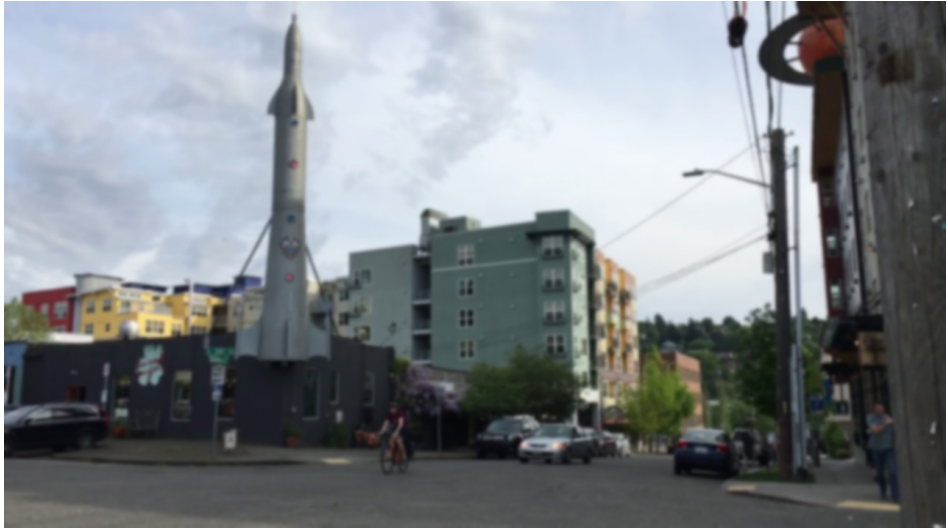

(*A*) *One Gaussian blurry observation.*

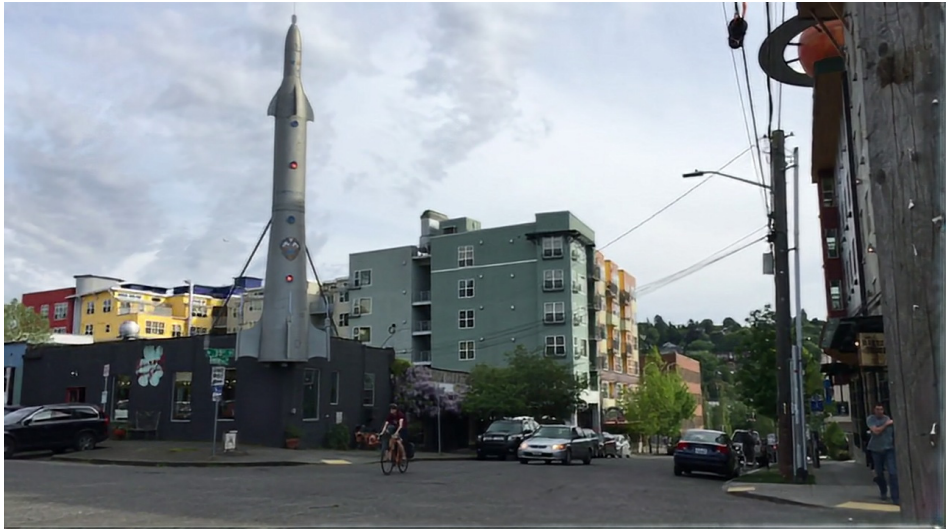

(*B*) *The deblurred result*

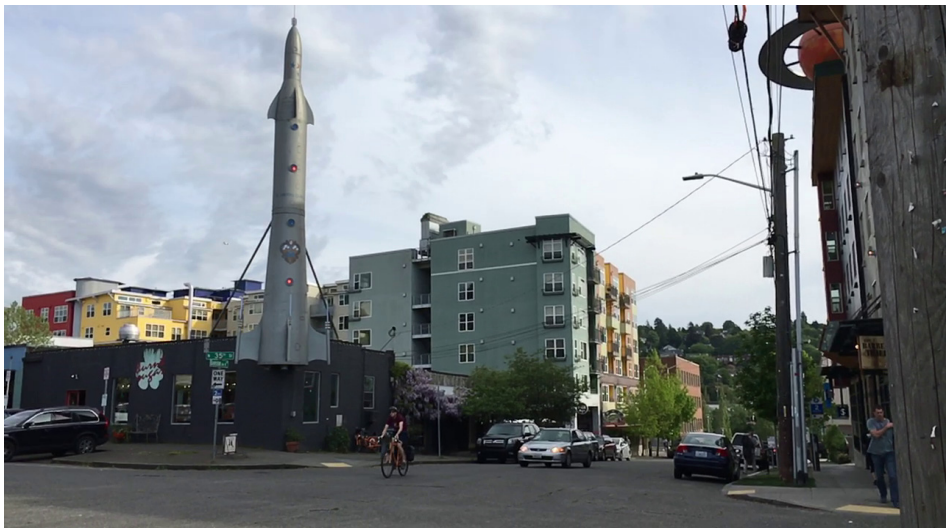

(*C*) *The ground truth from the DVD dataset* [16].

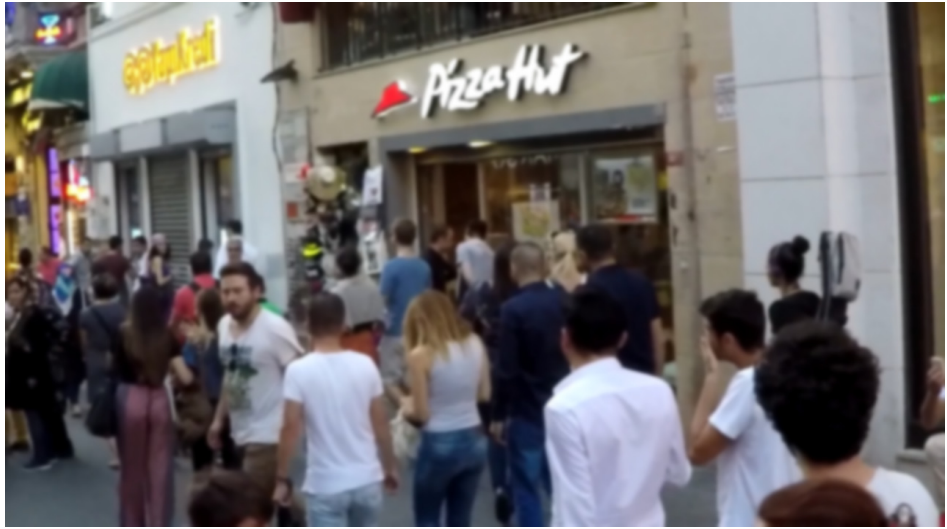

(*D*) *One Gaussian blurry observation.*

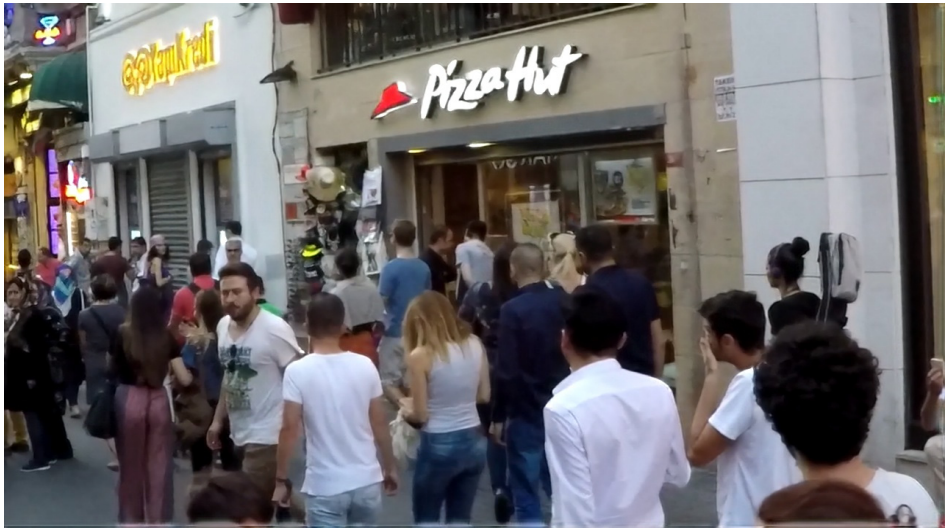

(*E*) *The deblurred result.*

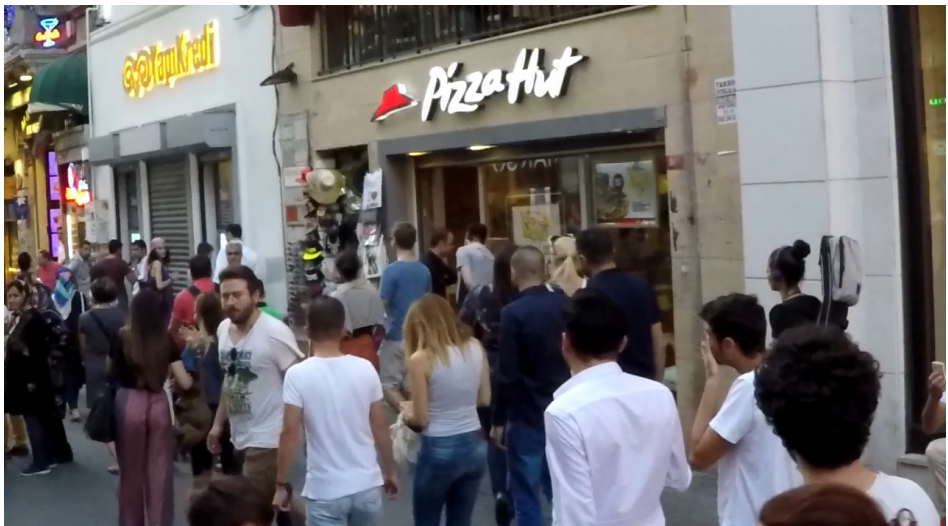

(*F*) *The ground truth from the GoPro dataset* [15].

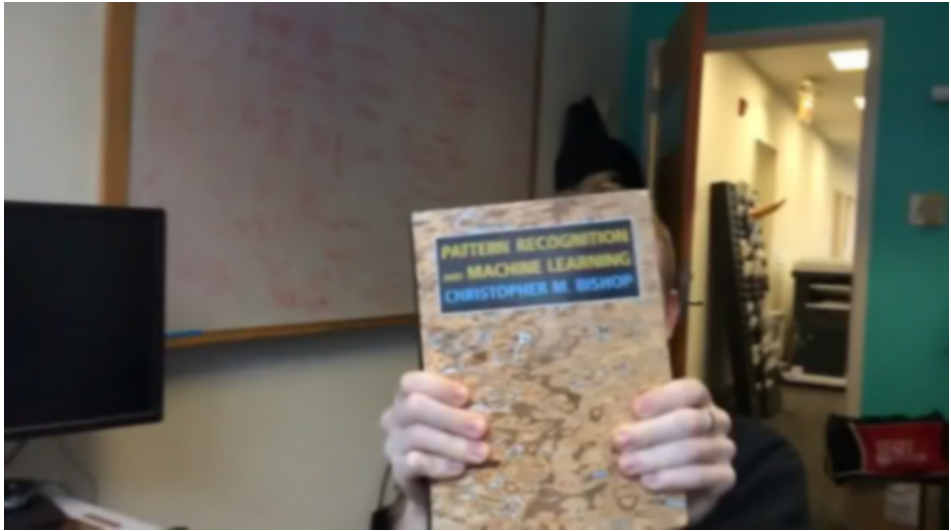

(*G*) *One Gaussian blurry observation.*

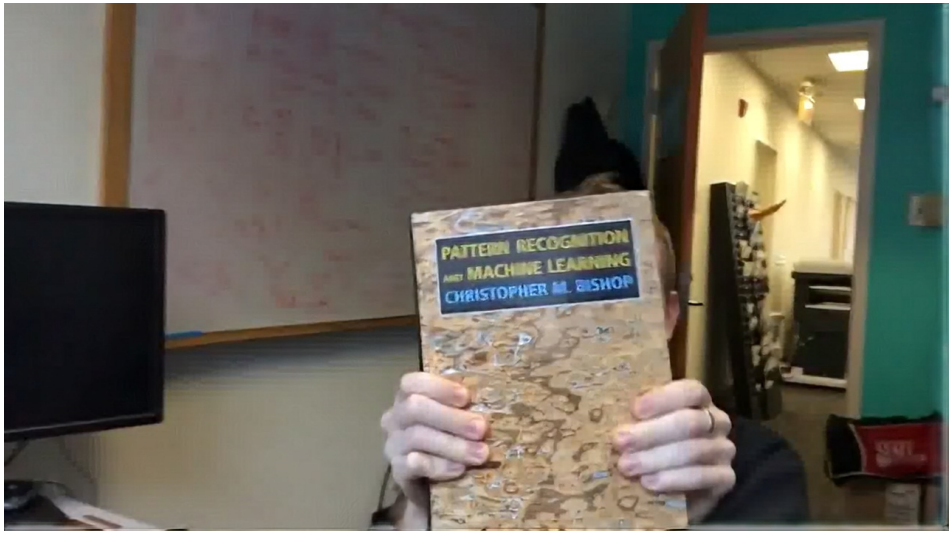

(*H*) *The deblurred result.*

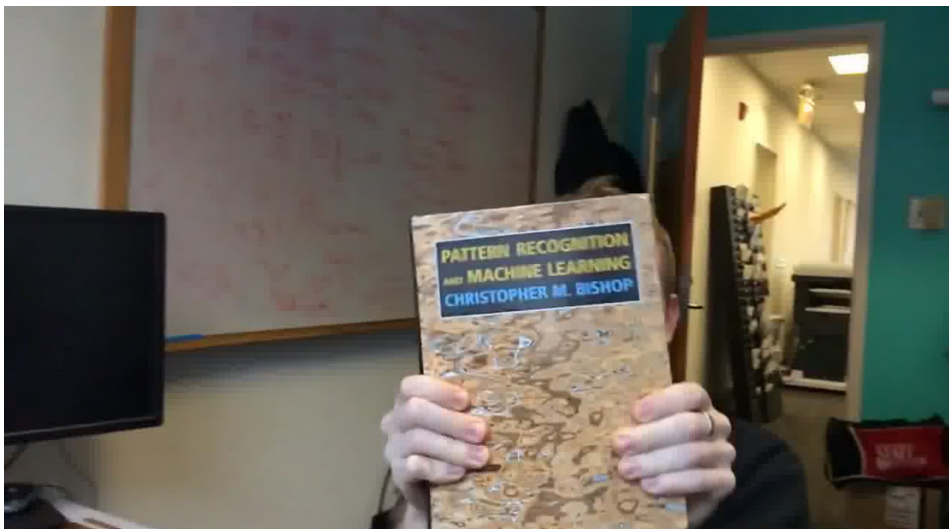

(*I*) *The ground truth from the NFS dataset* [17].

***Figure 29.*** *Deblurring results demonstration on three sets of examples.*

## 5.2 Performance on Motion Blurs

In the last section, we assigned a simple task to the network – learning a consistent and systematic blur from thousands of examples. In this section, we test the performance of the network on motion blurs, which are more complex and irregular as Figure 2 suggests. To make a comparison, we use the same default configuration as before. Measurements during training are displayed in the following Figure 30.

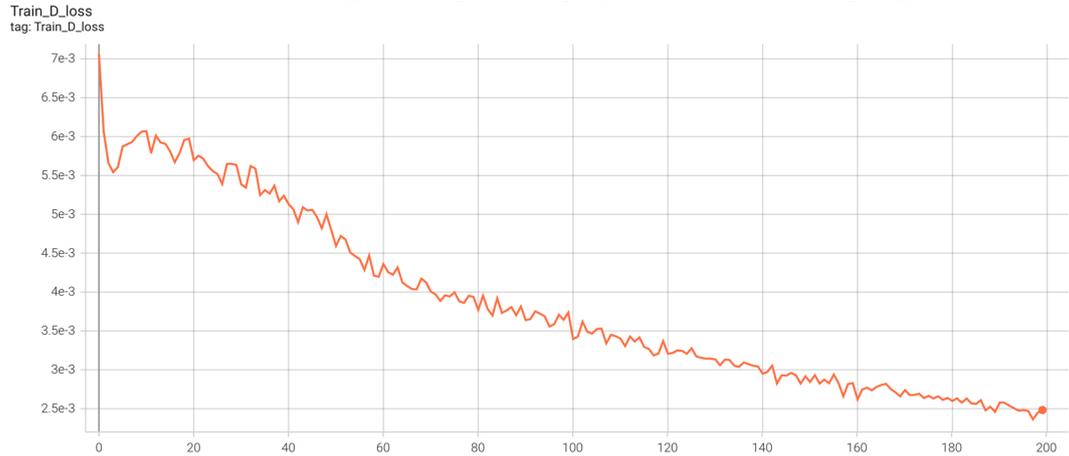

(*A*) *The adversarial loss of D in training.*

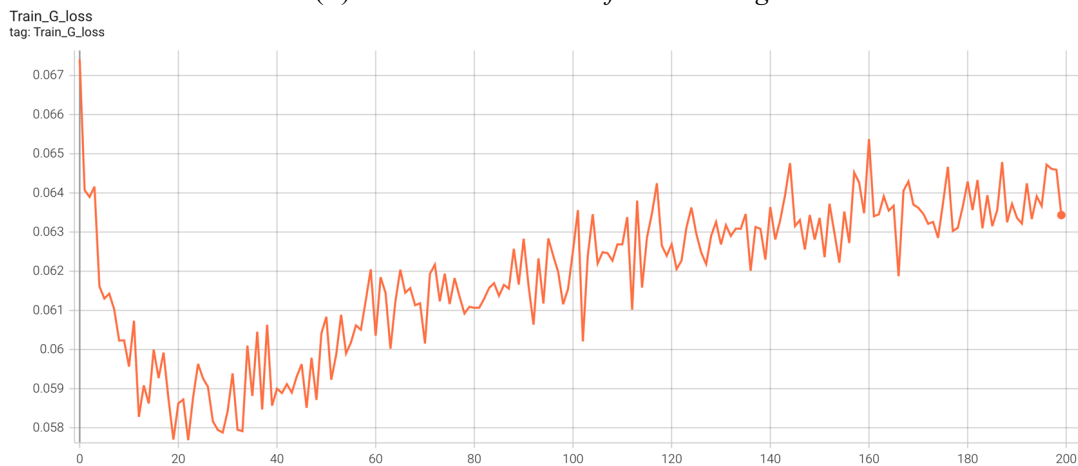

(*B*) *The total loss of G in training.*

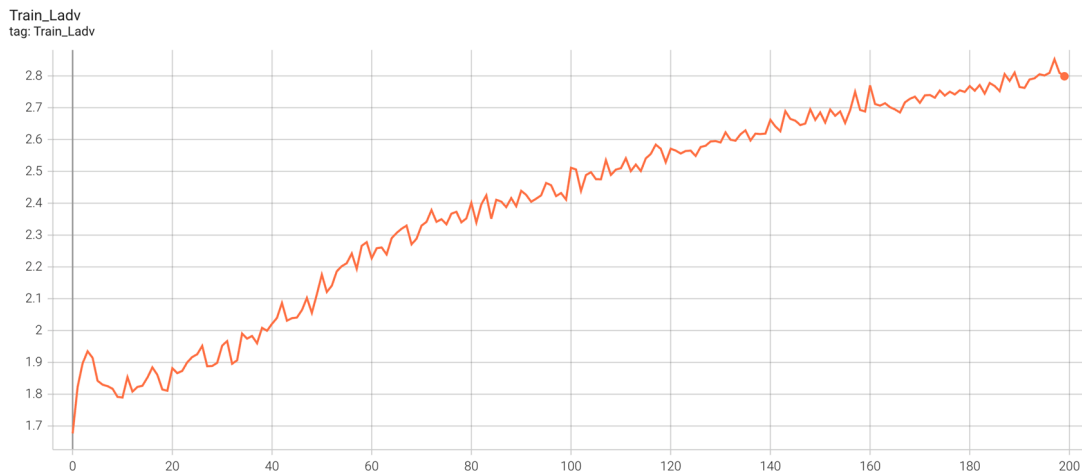

(*C*) *The adversarial loss of G in training.*

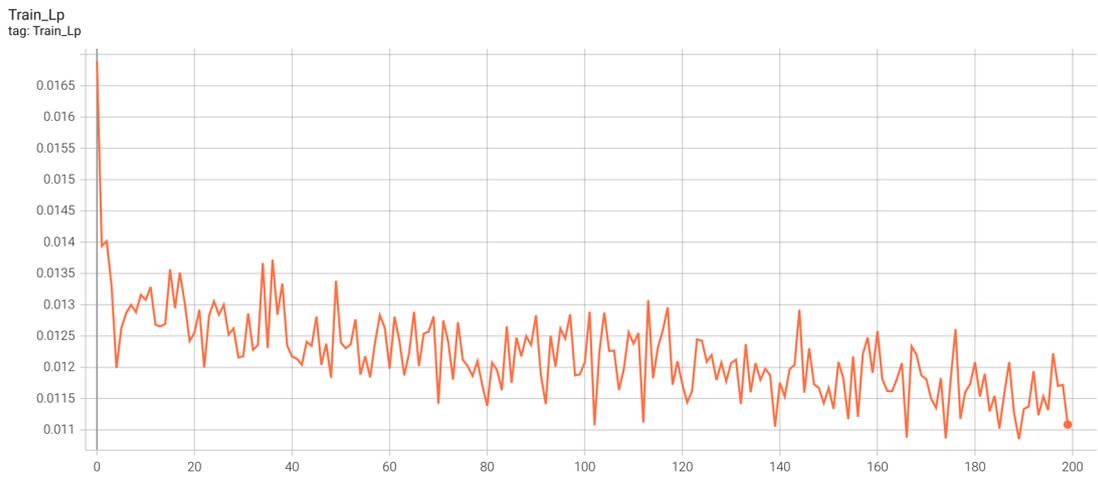

(*D*) *The perceptual loss $L_p$ of G in training with mean 0.0122 for the whole 200 epochs.*

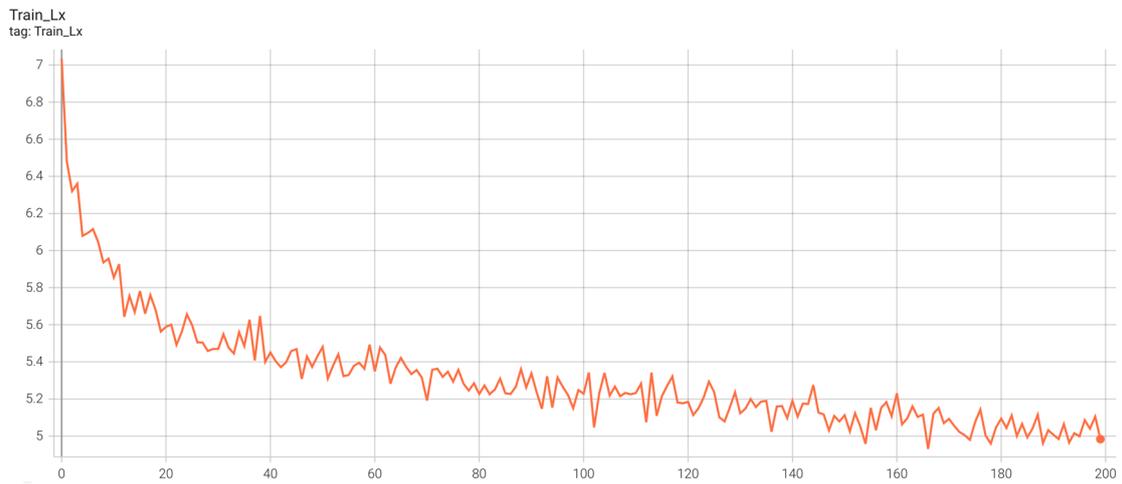

(*E*) *The content loss $L_x$ of G in training with mean 5.3109 for the whole 200 epochs.*

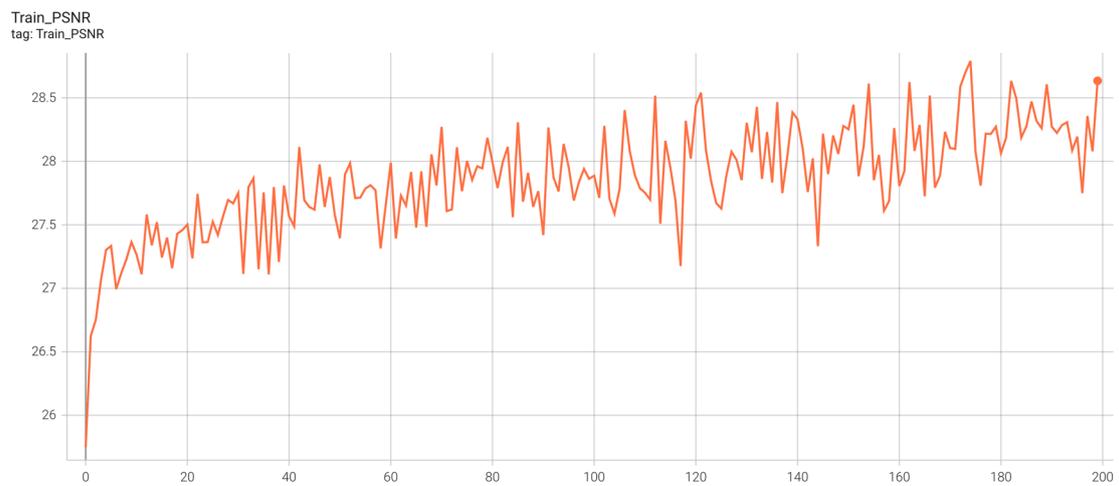

(*F*) *The PSNR between the generated image and the ground truth image in training.*

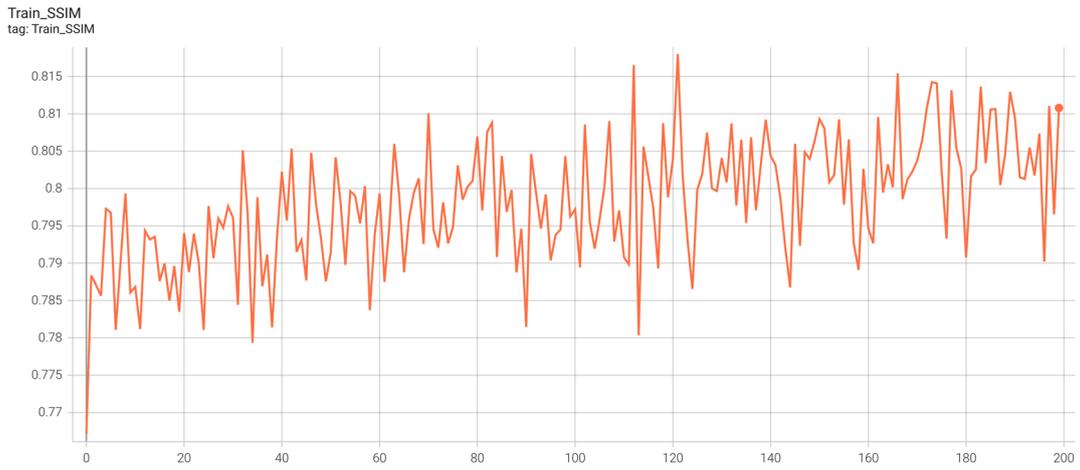

(*G*) *The SSIM between the generated image and the ground truth image in training.*

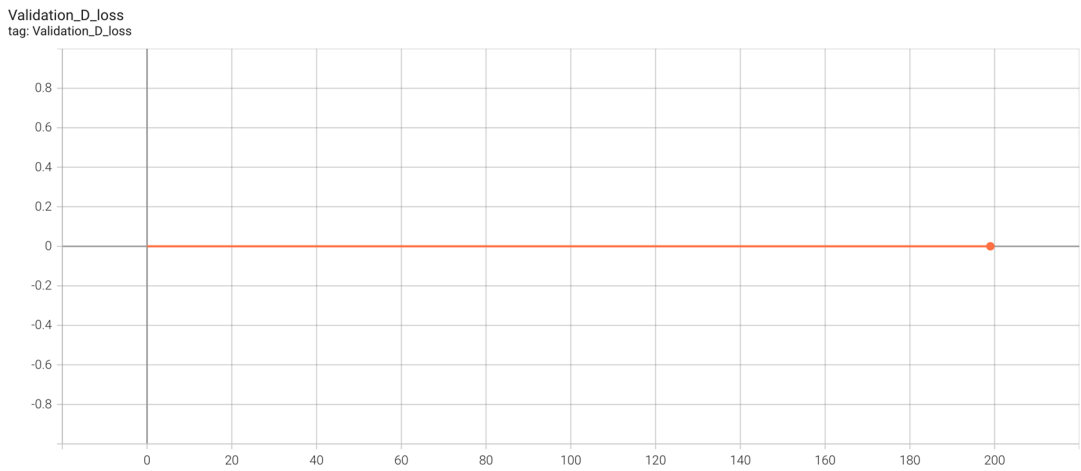

(*H*) *The adversarial loss of D in validation.*

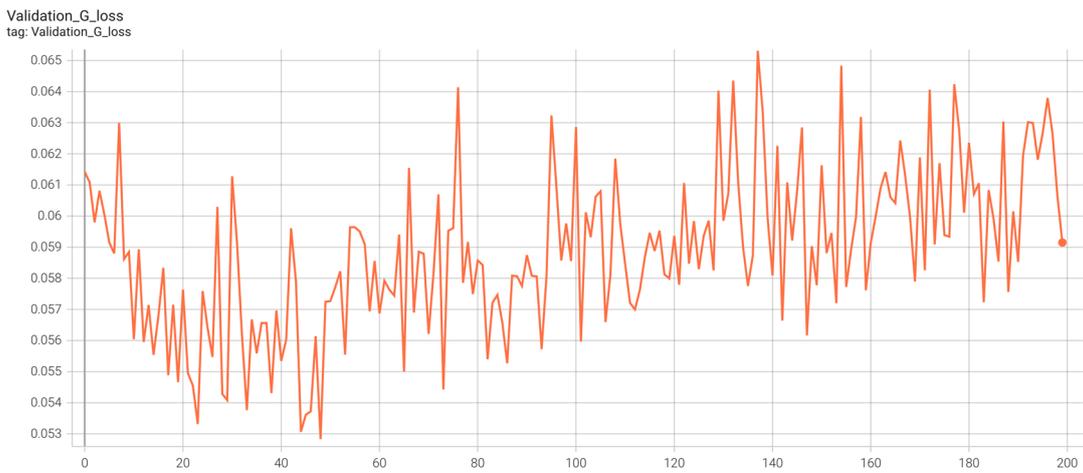

(*I*) *The total loss of G in validation.*

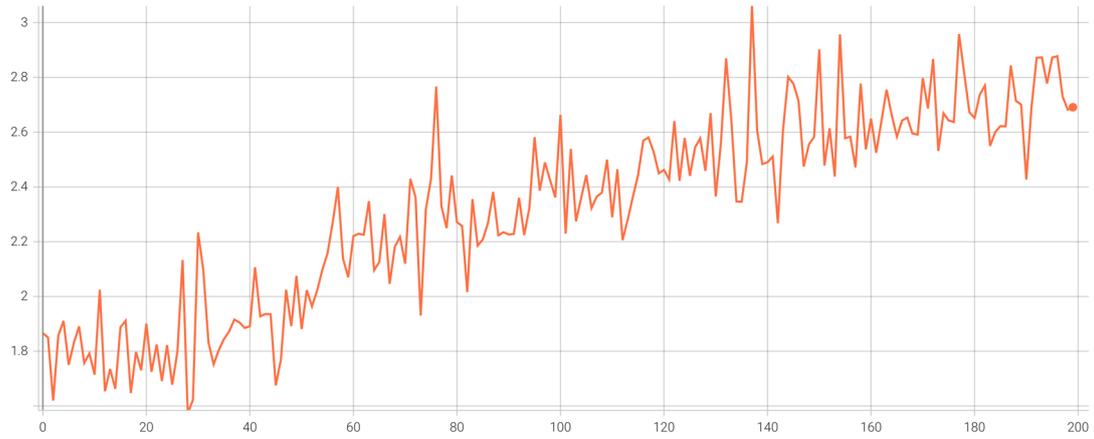

(*J*) *The adversarial loss of G in validation.*

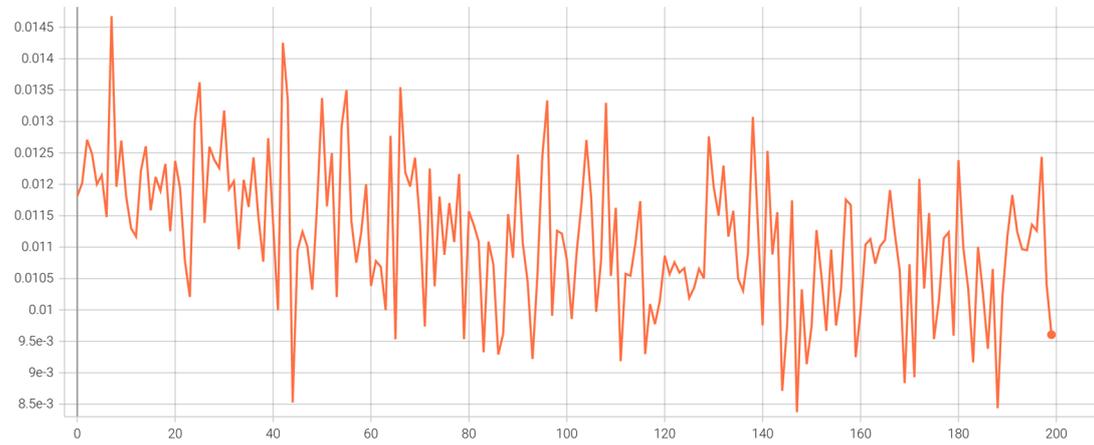

(*K*) *The perceptual loss $L_p$ of G in validation.*

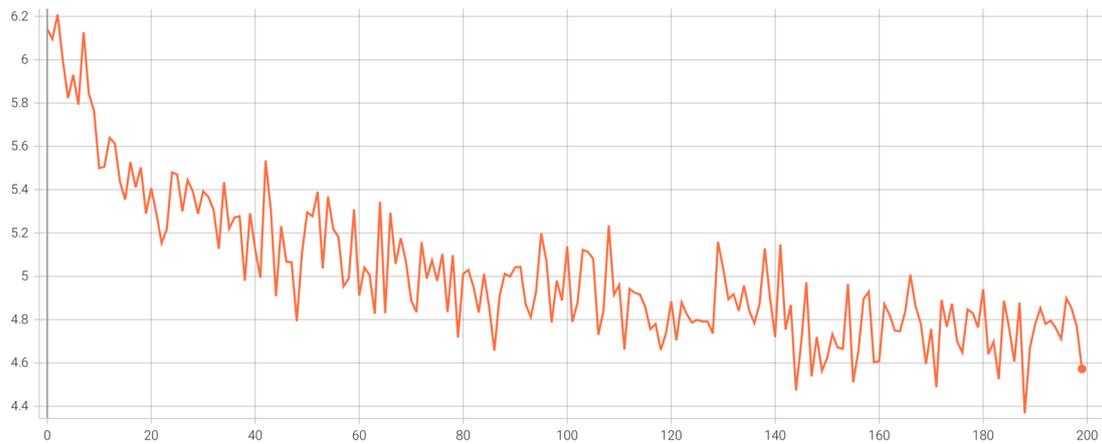

(*L*) *The content loss $L_x$ of G in validation.*

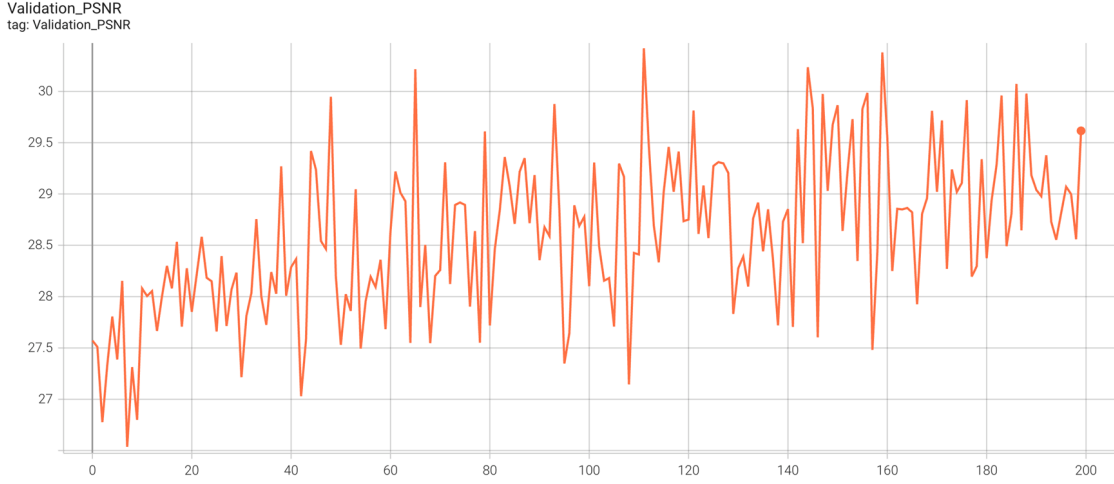

(*M*) *The PSNR between the generated image and the ground truth image in validation.*

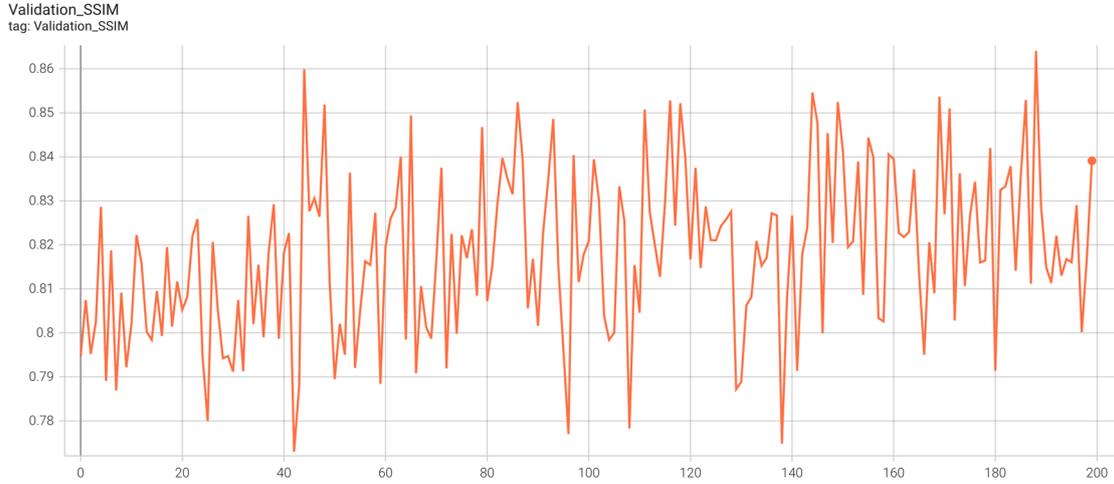

(*N*) *The SSIM between the generated image and the ground truth image in validation.*
***Figure 30.*** *Metrics recorded in one experiment for the case of motion blurs, with the same set of configurations as the previous experiment dealing with the consistent Gaussian blur.*

Compared to the case before, wilder fluctuations are observed especially for the PSNR and SSIM indices. This means that the network after training has not improved much on reconstruction quality. The performance on test datasets is summarized below:

***Table 5.*** *Performance of the trained network on different datasets, in the case of motion blurs. The PSNR and SSIM are the averaged results from testing images of corresponding dataset.*

| Dataset | PSNR / dB | SSIM | *Weight File |
|:---:|:---:|:---:|:---:|
| Sampled DVD | 26.30936 | 0.76595 | 'best' |
| (746 images) | 26.40772 | 0.77002 | 'last' |
| Sampled GoPro | 27.32346 | 0.80152 | 'best' |
| (1096 images) | 27.55554 | 0.80616 | 'last' |
| Sampled NFS | 30.11033 | 0.88418 | 'best' |
| (148 images) | 30.15745 | 0.88596 | 'last' |
| Merged of above | 27.14634 | 0.79461 | 'best' |
| (1990 images) | 27.32249 | 0.79858 | 'last' |

***Weight File****: as explained in **Table 4**.*

While the network with default configuration can deal with the consistent Gaussian blur as we saw in the previous section, it struggles to deal with motion blurs. Indicators, like the PSNR and the SSIM shown in table 5, reflect this limitation. We can also feel the gap between the reconstruction and the ground truth from the comparison in the following Figure 31.

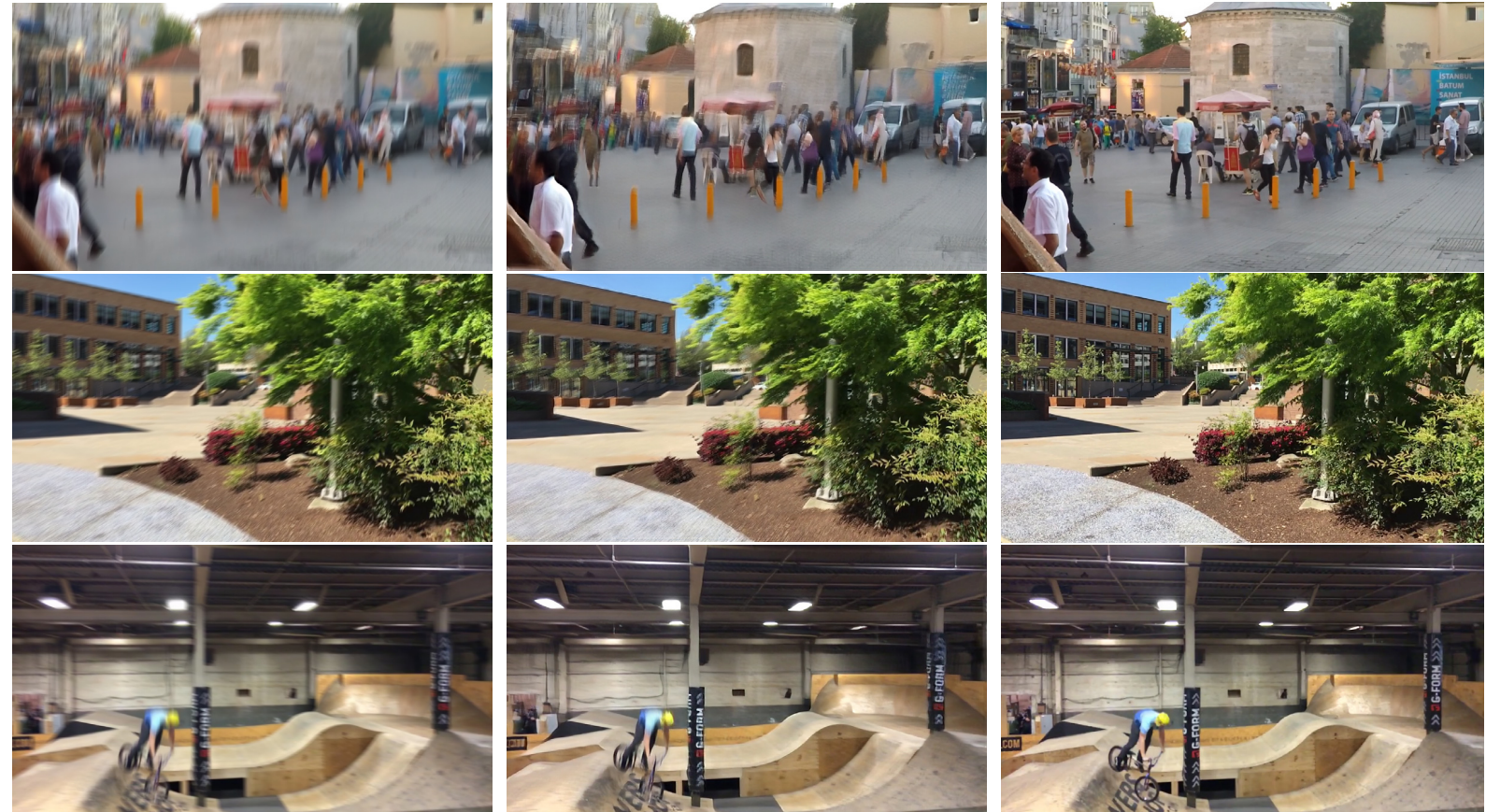

***Figure 31.*** *Comparison among the blurry observation (left column), the reconstruction (middle column) and the ground truth (right column). The three rows from top to bottom are testing images coming from the GoPro* [15], *DVD* [16], *and NFS* [17] *dataset, respectively. Corresponding to the measurements in Table 5, the baseline network performs the best on the NFS dataset* [17] *but poorly on the GoPro* [15] *and DVD* [16] *datasets.*

Hence, for degraded images caused by motion blurs we must find ways to make the network achieve more than this default baseline. In the next section we suggest several ways which may lead to a better reconstruction.

### 5.3 Suggestions for improvements

In this section, we propose possible directions for this DeblurGAN-v2 method to get enhanced. Our aim is to improve the network's performance in dealing with motion blurs. Network's performance is evaluated by metrics of PSNR and SSIM, as well as visual comparison of the reconstruction and the ground truth.

### 5.3.1 A similarity loss [41] and the Self-Attention module [42]

The SSIM index measures the structure similarities of two images, and the higher the SSIM index, the more similar the two images. Also, inspired by the idea of *Identity Loss* (*i.e.*, an $L_1$ Norm) provided by [41], we want the color distributions of the reconstruction and the ground truth to be similar. In this sense, we can design a new similarity loss, to replace the MSE loss $L_p$ in (5.2), as

$$L_{similar} = \mathrm{E}_{x \sim P_r, \hat{x} \sim P_g} \left[ \lambda \underbrace{\left| \hat{x} - x \right|}_{L_1 \text{ Norm}} + (1-\lambda) \cdot \underbrace{\left(1 - \mathrm{SSIM}(\hat{x}, x)\right)}_{\text{Structure Similarity index}} \right], \text{ where } \mathrm{SSIM}(x, y) \text{ denotes the SSIM index of two images.}$$

(5.3)

The parameter $\lambda$ controls the balance of the two terms – the $L_1$ Norm and the Structure Similarity index. Considering that the rationality of introducing the *Identity Loss* has yet to be proven, here we use a small and conservative weight on it by choosing $\lambda=0.2$ for the following experiments. We hope this similarity loss, which is more complex than the plain MSE loss, could make the reconstruction closer to the ground truth target in the sense of SSIM and color distributions.

Another modification is that we use the Self-Attention module, instead of the plain convolution, to process the top four small feature maps obtained in the middle pyramid of Figure 24, before up-sampling and concatenation in the right pyramid. Proposed by [42], the architecture is

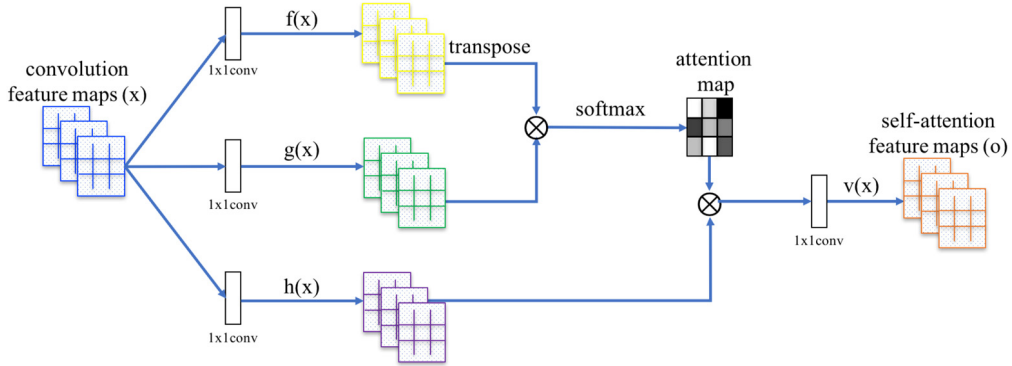

*Figure 32. The Self-Attention module proposed in [41]. Our codes are available at the Appendix.*

The $\otimes$ in the figure means the operation of matrix multiplication. Compared with the plain convolution where the operation is confined within the small area of the convolution window, matrix multiplication reflects cues from all locations. In this way, we hope to increase the receptive field of neurons, and therefore the generator could better generate details [41]. During programing, we found that this module can only be used upon feature maps of small size because the computation overhead for matrix multiplication is far more than that of the convolution.

We test the above two ideas on the GoPro dataset to see if they work. We set four experiments in a comparative sense, *i.e.*, one with the default configuration, two with one of these modifications respectively, and another one with both two modifications. Representative measurements recorded are shown in the following Figure 33.

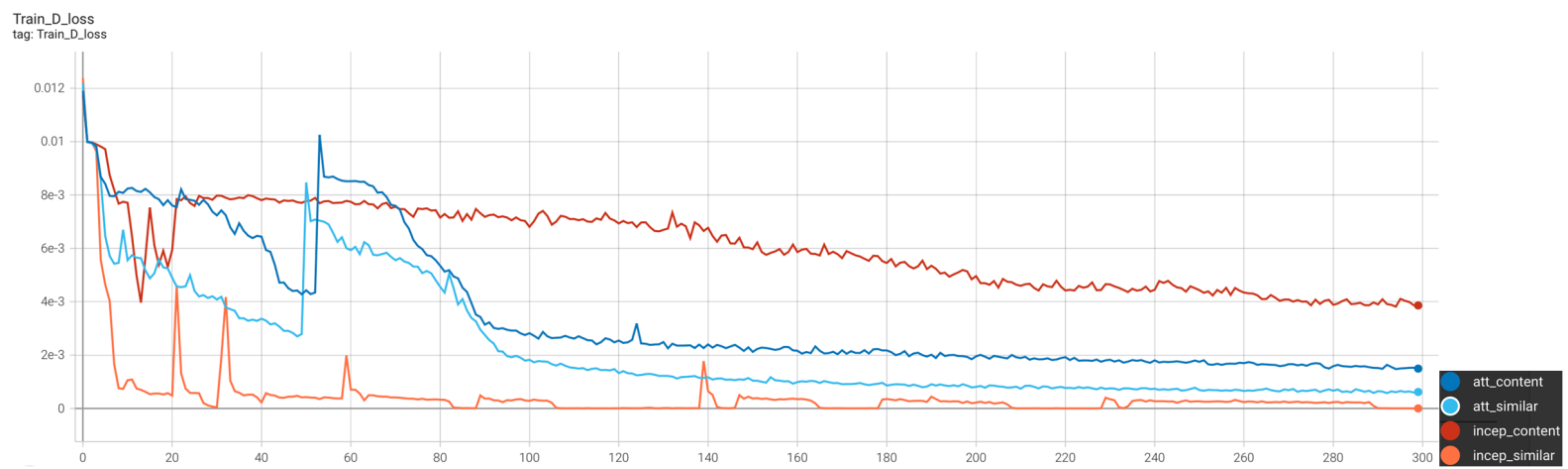

(*A*) *The adversarial loss of D in training.*

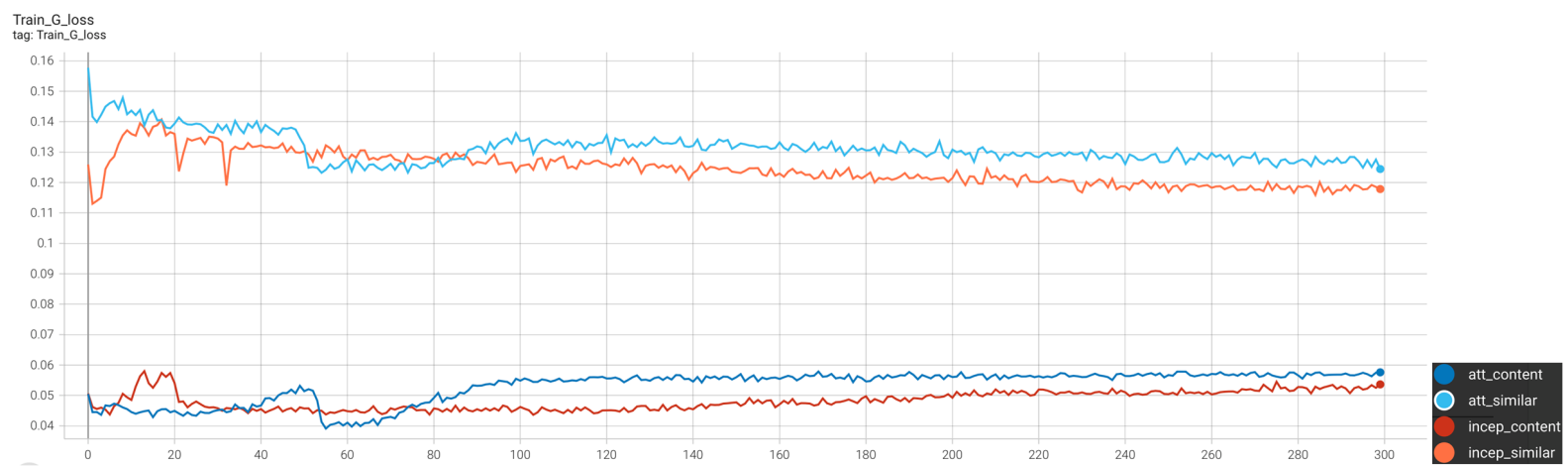

(*B*) *The total loss of G in training.*

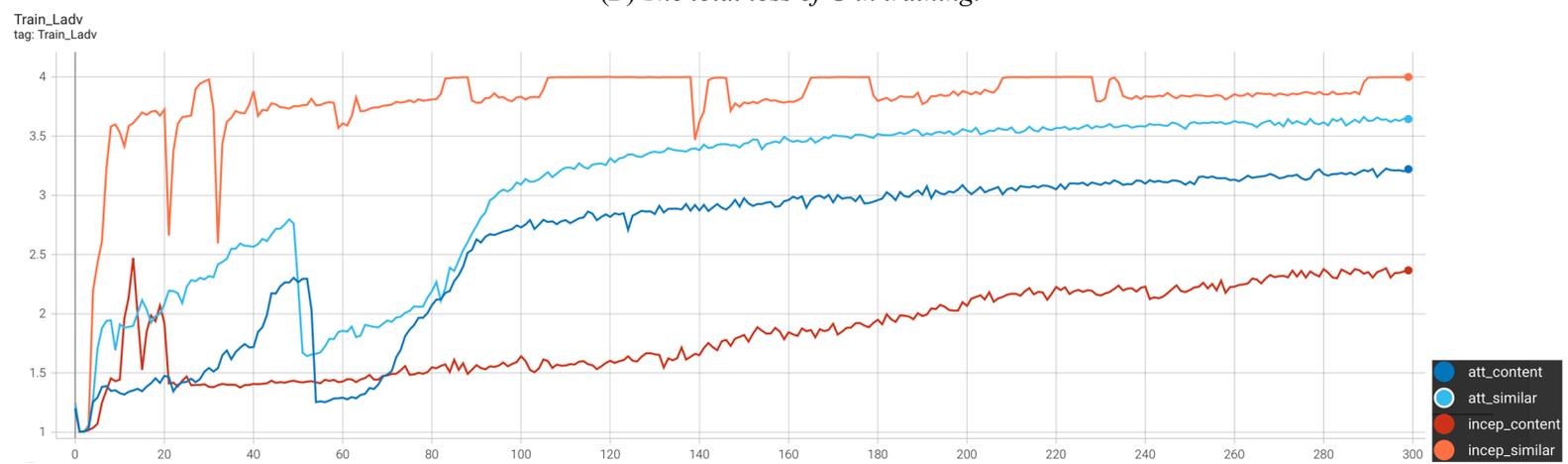

(*C*) *The adversarial loss of G in training.*

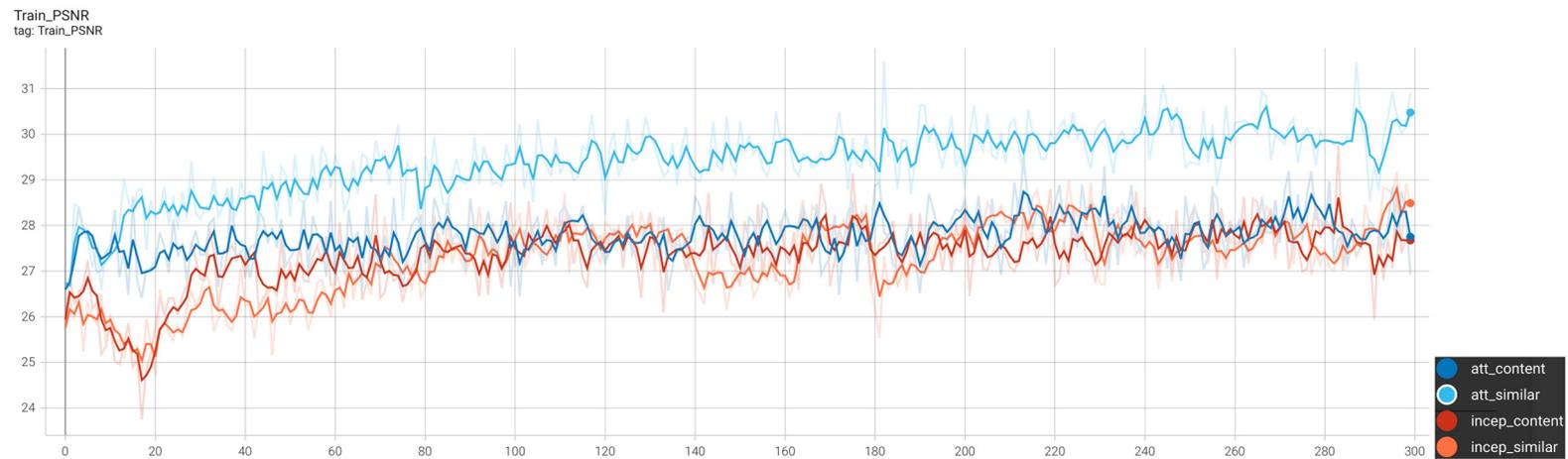

(*D*) *The PSNR between the generated image and the ground truth image in training (smoothed by 60% in TensorBoard).*

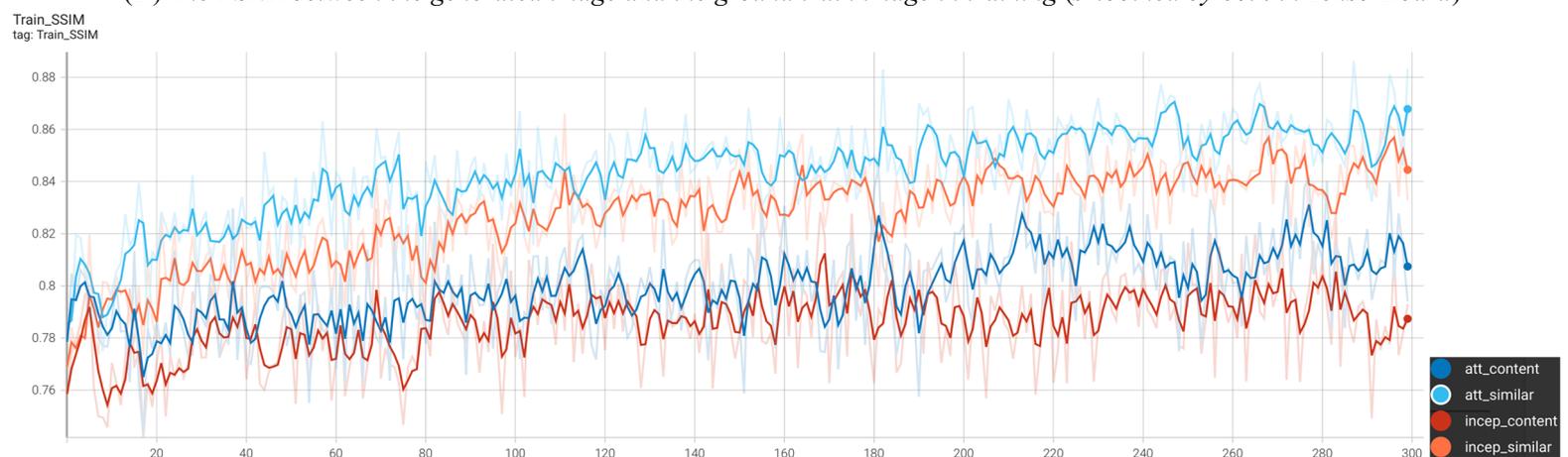

(*E*) *The SSIM between the generated image and the ground truth image in training (smoothed by 60% in TensorBoard).*

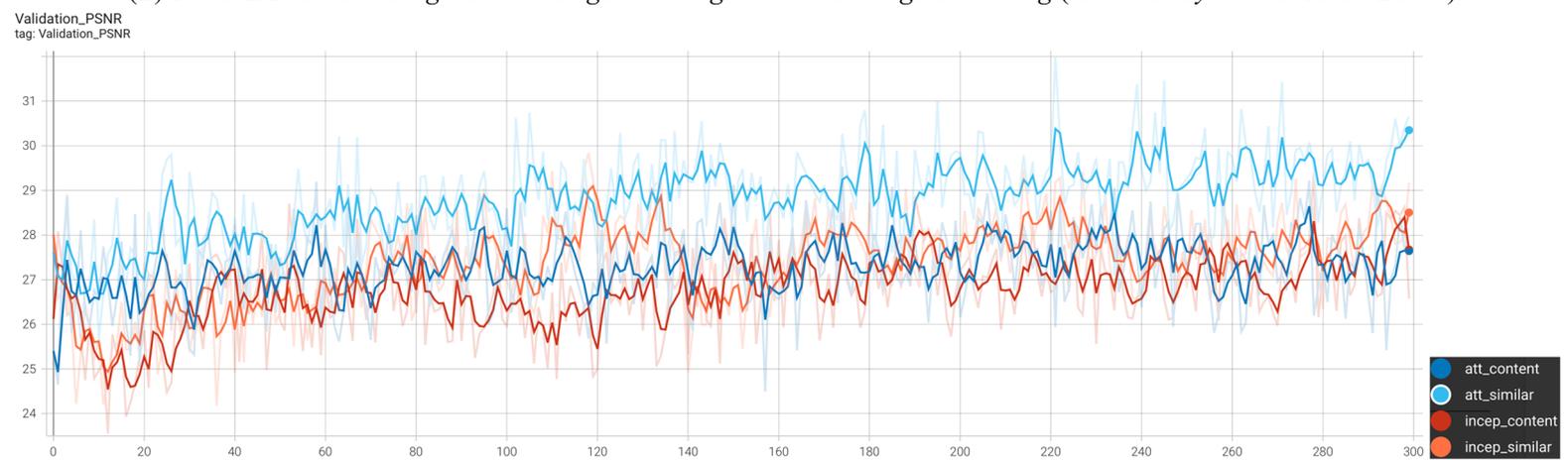

(*F*) *The PSNR between the generated image and the ground truth image in validation (smoothed by 60% in TensorBoard).*

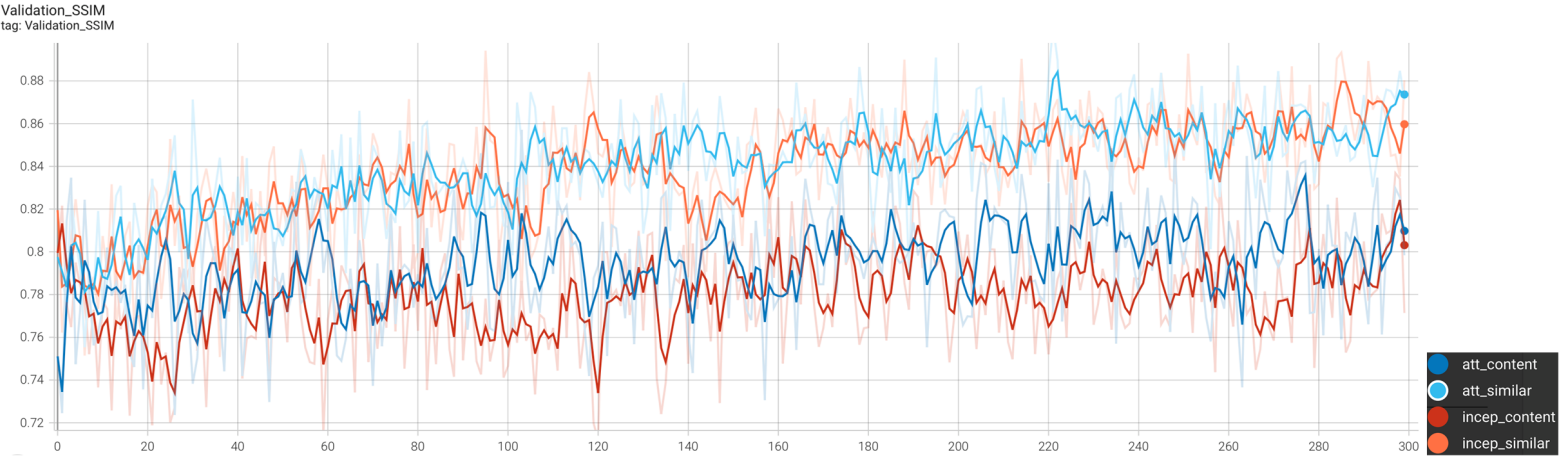

(*G*) *The SSIM between the generated image and the ground truth image in validation (smoothed by 60% in TensorBoard).*
***Figure 33.*** *Four comparative experiments showing network's performance under different configurations, where "incep_content" means using no self-attention module and no similarity loss, "incep_similar" means only using the similarity loss, "att_content" means only using the self-attention module, and "att_similar" means using both the self_attention module and the similarity loss.*

First, let us look at the measurements of PSNR and SSIM which reflect the reconstruction quality directly. During validation the network with the default configuration (*i.e.*, *incep_content*) performs the worst because of the almost lowest PSNR and SSIM it produces, as shown in (F) and (G). To the contrast, the network with both two modifications (*i.e.*, *att_similar*) performs the best during training and validation because of the highest PSNR and SSIM (on average) it produces, as shown from (D) to (G). The performances of networks with one of our modifications lie in the middle. It shows that our two ideas can bring some improvements upon the baseline. Performances of the four models on the testing GoPro dataset are summarized in the following Table 6. We see that our two modifications can improve PSNR by

$$(28.24175 - 26.39044)/26.39044 \approx 0.0701 = 7.01\%$$

and SSIM by

$$(0.83416 - 0.78238)/0.78238 \approx 0.0662 = 6.62\%$$

on the network in the default configuration.

***Table 6.*** *Performance comparison of four networks on the same testing dataset. PSNR and SSIM are averaged results from testing images.*

| **Dataset** | **Model** | **PSNR** / dB | **SSIM** | ***Weight File** |
|---|---|---|---|---|
| Sampled GoPro (1096 images) | *att_similar* (self-attention + similarity loss) | 27.88333 | 0.83133 | 'best' |
| | | **28.24175** | **0.83416** | 'last' |
| | *att_content* (self-attention only) | 27.17987 | 0.80422 | 'best' |
| | | 27.16117 | 0.80502 | 'last' |
| | *incep_similar* (similarity loss only) | 27.45496 | 0.80679 | 'best' |
| | | 27.56143 | 0.81255 | 'last' |
| | *incep_content* (default) | 26.39044 | 0.78238 | 'best' |
| | | 26.53476 | 0.78910 | 'last' |

***Weight File****: as explained in **Table 4**.*

We can also feel the progress by the following comparisons in perceptual quality.

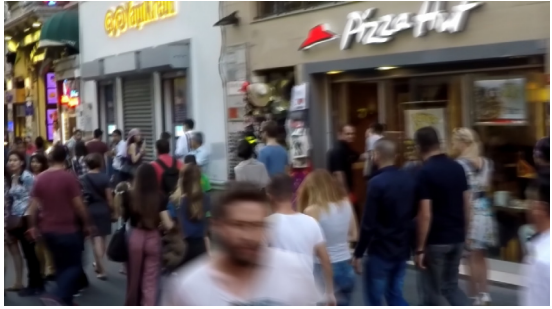
(*A*) *A blurry observation from GoPro* [15].

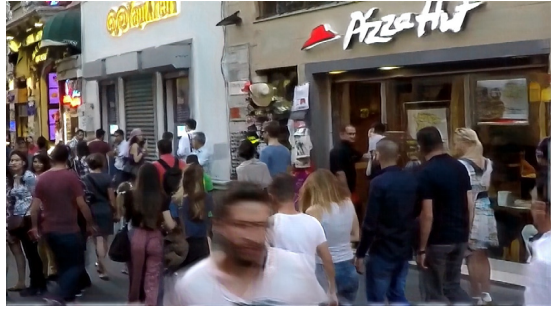
(*B*) *Restoration by default, "incep_content".*

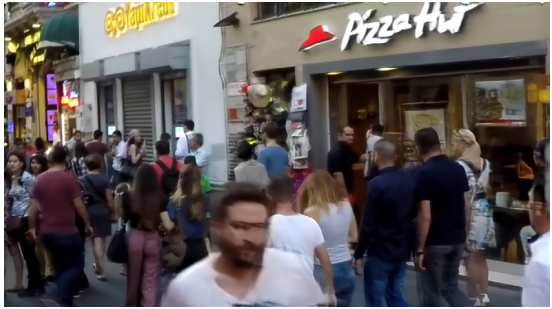
(*C*) *Restoration by "incep_similar".*

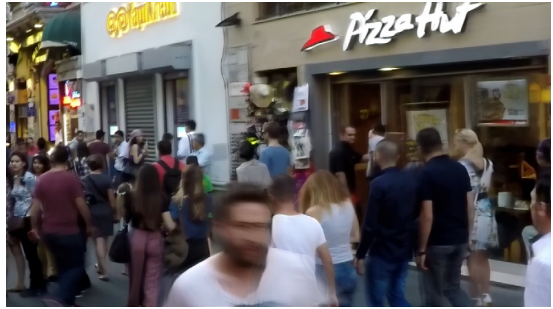
(*D*) *Restoration by "att_content".*

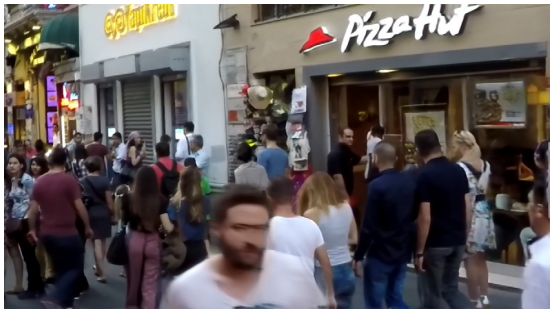
(*E*) *Restoration by "att_similar".*

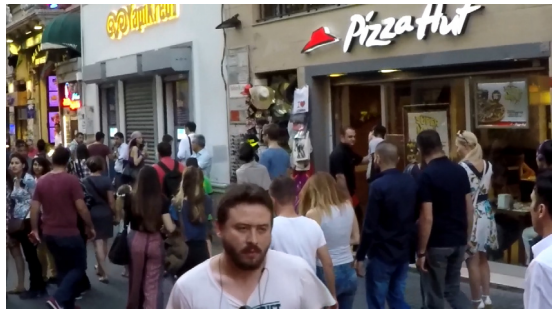
(*F*) *The ground truth.*

**Figure 34.** *Perceptual quality comparison of the results from the four models. We use the "last weight" file for demonstration, and the blurry observation is randomly selected from the testing dataset.*

Although there is still a big gap between the reconstruction of our best model and the ground truth, from this random example we have observed the progress. The face of the man in white in (B) is distorted and full of artificial effects, while it appears more natural in (E) with the blurry impact mitigated.

It is also worth to investigate the variations of the adversarial losses of G and D during training. From Figure 33 (A) it is observed that D no longer has an all-time dominance over G, because its adversarial loss has opportunities to rise up at some times. It may indicate that G has once succeeded in deceiving D. It is not an easy thing for G in this case, because G has a much more complex architecture involving much more parameters than D. In this sense G is much harder to train than D. If G has once won a victory in this unequal situation, then it can be said that G has made considerable progress for a time. We see this as a positive sign for an effective game between G and D. We also observed from Figure 33 (C) that compared to the case of the network with default configuration (*i.e.*, *incep_content*), our modifications can make the adversarial loss of G converge better to a constant when 300 epochs are done. It means that G does not collapse in the competition with D but survives in the end.

### 5.3.2 Other possible directions for improvement

In this section we point out some another possible improvement directions, but due to the limitation of the time and computation power, we did not validate them. These suggestions are based on our best knowledge and experience accumulated along the way, so we believe they are useful for future works.

### 5.3.2.1 Train G more often than D

The GAN works in the sense of the cooperation between G and D. However, as we saw from Figure 24 and Figure 26, the architecture of G is much more complex than that of D. Also, as we discussed before in Section 5.3.1, it means that G has much more learnable parameters than D has. Consequently, G is much harder to train than D. Therefore, in order to increase the chance of G in the game, we can train G multiple times after training D once in a round, and therefore try to make G keep up with the pace of D.

### 5.3.2.2 Multi-scale Perceptual Loss for G

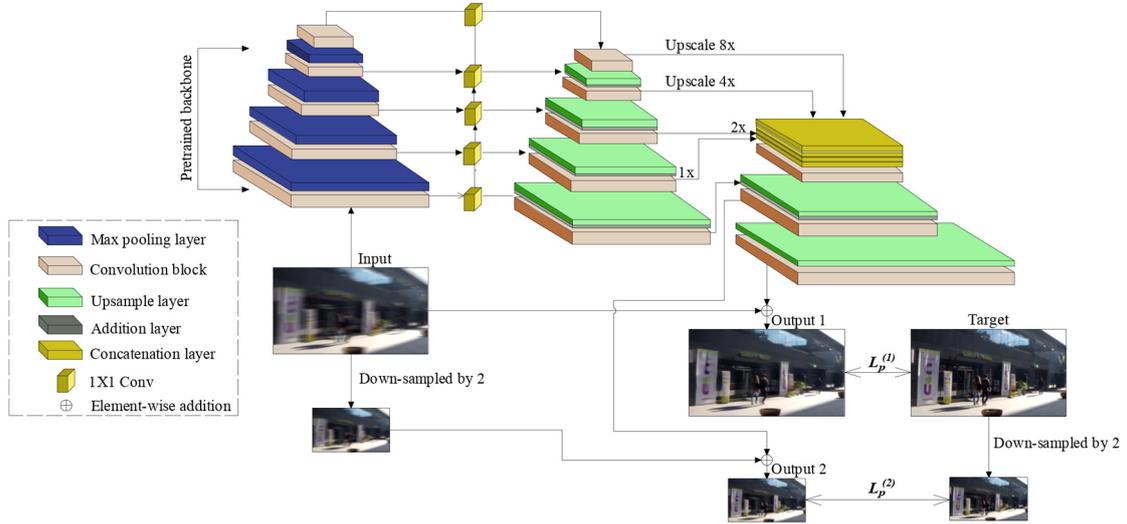

***Figure 35.*** *Illustration of the G's multi-scale perceptual loss.*

Another potential improvement direction is to use a multi-scale perceptual loss as

$$L_p = \gamma_1 L_p^{(1)} + \gamma_2 L_p^{(2)}, \text{ where } \gamma_1 > \gamma_2 \qquad (5.4)$$

instead of $L_p^{(1)}$ only for (5.2). $L_p^{(1)}$ is the MSE loss by the last feature map, and $L_p^{(2)}$ is the MSE loss by the penultimate feature map. Since our generator contains multi-scale features, we naturally hope that each loss on each feature map can be small. And because the output of the last feature map is what we care about most, we may give it a higher weight, *i.e.*, $\gamma_1 > \gamma_2$. Specific values for them may be determined depending on experiment's results.

# Chapter 6

# Conclusions and future works

Blind Image Deblurring (BID) is an ill-posed problem where infinite number of solutions are possible. Thus, we need to regularize our reconstruction to make it approximate the ground truth as close as possible. We can either design a regularization functional in the optimization objective, as in the regularization-based method, or use data-driven approaches. For the former, it can be a challenging job to design a regularization function that is both computationally convenient and able to capture the great complexity in practice. Also, the choice of regularization parameter can be a tough problem as it often relies on a proper estimate of noise and no prior knowledge is available. The performance of data-driven methods relies on datasets and the ability of the network to learn. And although some uncertainty still exists, for GAN new research results are constantly emerging as well as related theories.

The contributions of this thesis are mainly summarized in the following four points. First, we gave a very detailed account on image deblurring principles from the perspective of spectral decomposition. Second, based on maximum a posterior estimation we demonstrate how to apply our prior knowledge of images to crafting a regularization functional by hand. Third, we validate the idea of training a network to learn the regularization functional, by analyzing the result of a method that is based on the Wasserstein Generative Models. Finally, we analyze a more advanced deblurring method using GAN based on numerical tests. We also suggest some potential ways to improve it and test some of them based on experiment results. Although we have not understood in much depth the working mechanism of GAN, the idea of exploring based on experiments is beneficial. Recording our exploration journey, this work may provide some useful experience to researchers in the image processing or the computer vision field.

There are some issues remaining to be settled as our future works. First, we did not explore much in depth for choosing a suitable regularization parameter. More often, this variable is set as a hyperparameter which is not learnable from data. It means that we must determine it based on experiments, *i.e.*, trail-and-error. Second, we did not validate all our suggestions in Section 5.3. It was due to a limited time and the limited computation power we have. Sometimes the experiment results are unstable, and we need to pick one that best suits our expectation. So, for an idea we need to run the algorithm several times during which some parameters are altered. And for one round it takes one or two days. With these two issues settled, we have very reasons to believe that our work can be extend to more general cases to try to deal with blurs of more kinds.

neural networks. Science, 313(5786):504–507, 2006.

[40] Christian Ledig, Lucas Theis, Ferenc Husz´ar, Jose Caballero, Andrew Cunningham, Alejandro Acosta, Andrew Aitken, Alykhan Tejani, Johannes Totz, and Zehan Wang. Photorealistic single image super-resolution using a generative adversarial network[C]. In Proceedings of the IEEE conference on computer vision and pattern recognition, pages 4681–4690, 2017.

[41] Qu, Yangyang, Chao liu, and Yongsheng Ou. LEUGAN: Low-Light Image Enhancement by Unsupervised Generative Attentional Networks. (2020): n. pag. Print.

[42] H. Zhang, I. Goodfellow, D. Metaxas, and A. Odena. Self-attention generative adversarial networks[C]. International conference on machine learning. PMLR, 2019: 7354-7363.

[43] https://archive.siam.org/books/fa03/.

# Appendix

We benefit a lot from the Open-Source project by Orest Kupyn *et al*. [14] available at https://github.com/VITA-Group/DeblurGANv2. Based on that we add a new Self-Attention module, and our codes are available here.

# The Self-Attention module

```python
import torch
import torch.nn as nn
from pretrainedmodels import inceptionresnetv2
from torch.autograd import Variable
from .spectral import SpectralNorm
import torch.nn.functional as F

class Self_Attn(nn.Module):
    def __init__(self, inChannels, k=8):
        super(Self_Attn, self).__init__()
        embedding_channels = inChannels // k                              # C_bar
        self.key     = nn.Conv2d(inChannels, embedding_channels, 1)
        self.query   = nn.Conv2d(inChannels, embedding_channels, 1)
        self.value   = nn.Conv2d(inChannels, embedding_channels, 1)
        self.self_att = nn.Conv2d(embedding_channels, inChannels, 1)
        self.gamma   = nn.Parameter(torch.tensor(0.0))
        self.softmax = nn.Softmax(dim=1)

    def forward(self, x):
        """
        inputs:
            x: input feature map [Batch, Channel, Height, Width]
        returns:
            out: self attention value + input feature
            attention: [Batch, Channel, Height, Width]
        """
        batchsize, C, H, W = x.size()
        N = H * W                                                    # Number of features
        f_x = self.key(x).view(batchsize, -1, N)                     # Keys      [B, C_bar, N]
        g_x = self.query(x).view(batchsize, -1, N)                   # Queries   [B, C_bar, N]
        h_x = self.value(x).view(batchsize, -1, N)                   # Values    [B, C_bar, N]

        s = torch.bmm(f_x.permute(0,2,1), g_x)                       # Scores    [B, N, N]
        beta = self.softmax(s)                                       # Attention Map [B, N, N]

        v = torch.bmm(h_x, beta)                                     # Value x Softmax    [B, C_bar, N]
        v = v.view(batchsize, -1, H, W)                              # Recover input shape [B, C_bar, H, W]
```

```python
        o = self.self_att(v)                        # Self-Attention output      [B, C, H, W]

        y = self.gamma * o + x                      # Learnable gamma + residual
        return y

class FPNHead(nn.Module):
    def __init__(self, num_in, num_mid, num_out):
        super().__init__()
        self.attn=Self_Attn(num_in)
        self.block = nn.Conv2d(num_in, num_out, kernel_size=3, padding=1, bias=False)
        # self.block1 = nn.Conv2d(num_mid, num_out, kernel_size=3, padding=1, bias=False)

    def forward(self, x):
        return self.block(self.attn(x))

class ConvBlock(nn.Module):
    def __init__(self, num_in, num_out, norm_layer):
        super().__init__()

        self.block = nn.Sequential(
            nn.Conv2d(num_in, num_out, kernel_size=3, padding=1),
            norm_layer(num_out),
            nn.ReLU(inplace=True),
        )

    def forward(self, x):
        x = self.block(x)
        return x

class FPNInceptionAtt(nn.Module):
    def __init__(self, norm_layer, output_ch=3, num_filters=128, num_filters_fpn=256):
        super().__init__()

        # Feature Pyramid Network (FPN) with four feature maps of resolutions
        # 1/4, 1/8, 1/16, 1/32 and `num_filters` filters for all feature maps.
        self.fpn = FPN(num_filters=num_filters_fpn, norm_layer=norm_layer)

        # The segmentation heads on top of the FPN

        self.head1 = FPNHead(num_filters_fpn, num_filters, num_filters)
        self.head2 = FPNHead(num_filters_fpn, num_filters, num_filters)
```

```python
        self.head3 = FPNHead(num_filters_fpn, num_filters, num_filters)
        self.head4 = FPNHead(num_filters_fpn, num_filters, num_filters)
        # self.attn1 = Self_Attn(4 * num_filters, "relu")
        # self.attn1 = Self_Attn(num_filters, "relu")
        # self.attn = Self_Attn(32)
        self.smooth = nn.Sequential(
            SpectralNorm(
                nn.Conv2d(4 * num_filters, num_filters, kernel_size=3, padding=1)
            ),
            nn.ReLU(),
        )

        self.smooth2 = nn.Sequential(
            SpectralNorm(
                nn.Conv2d(num_filters, num_filters//2, kernel_size=3, padding=1)
            ),
            nn.ReLU(),
        )

        self.final = nn.Conv2d(num_filters//2, output_ch, kernel_size=3, padding=1)

    def unfreeze(self):
        self.fpn.unfreeze()

    def forward(self, x):
        map0, map1, map2, map3, map4 = self.fpn(x)

        map4 = nn.functional.upsample(self.head4(map4), scale_factor=8, mode="bilinear")
        map3 = nn.functional.upsample(self.head3(map3), scale_factor=4, mode="bilinear")
        map2 = nn.functional.upsample(self.head2(map2), scale_factor=2, mode="bilinear")
        map1 = nn.functional.upsample(self.head1(map1), scale_factor=1, mode="bilinear")

        smoothed = self.smooth(torch.cat([map4, map3, map2, map1], dim=1))
        smoothed = nn.functional.upsample(smoothed, scale_factor=2, mode="bilinear")
        smoothed = self.smooth2(smoothed + map0)
        smoothed = nn.functional.upsample(smoothed, scale_factor=2, mode="bilinear")
        # smoothed = self.attn(smoothed)
        final = self.final(smoothed)
        # return torch.tanh(final)
        res = torch.tanh(final) + x
        return torch.clamp(res, min=-1, max=1)

class FPN(nn.Module):
```

```python
def __init__(self, norm_layer, num_filters=256):
    """Creates an `FPN` instance for feature extraction.
    Args:
    num_filters: the number of filters in each output pyramid level
    pretrained: use ImageNet pre-trained backbone feature extractor
    """

    super().__init__()
    self.inception = inceptionresnetv2(num_classes=1000, pretrained="imagenet")

    self.enc0 = self.inception.conv2d_1a
    self.enc1 = nn.Sequential(
        self.inception.conv2d_2a,
        self.inception.conv2d_2b,
        self.inception.maxpool_3a,
    )  # 64
    self.enc2 = nn.Sequential(
        self.inception.conv2d_3b,
        self.inception.conv2d_4a,
        self.inception.maxpool_5a,
    )  # 192
    self.enc3 = nn.Sequential(
        self.inception.mixed_5b, self.inception.repeat, self.inception.mixed_6a,
    )  # 1088
    self.enc4 = nn.Sequential(
        self.inception.repeat_1, self.inception.mixed_7a,
    )  # 2080
    self.td1 = nn.Sequential(
        nn.Conv2d(num_filters, num_filters, kernel_size=3, padding=1),
        norm_layer(num_filters),
        nn.ReLU(inplace=True),
    )
    self.td2 = nn.Sequential(
        nn.Conv2d(num_filters, num_filters, kernel_size=3, padding=1),
        norm_layer(num_filters),
        nn.ReLU(inplace=True),
    )
    self.td3 = nn.Sequential(
        nn.Conv2d(num_filters, num_filters, kernel_size=3, padding=1),
        norm_layer(num_filters),
        nn.ReLU(inplace=True),
    )
    self.pad = nn.ReflectionPad2d(1)
    self.lateral4 = nn.Conv2d(2080, num_filters, kernel_size=1, bias=False)
```

```python
        self.lateral3 = nn.Conv2d(1088, num_filters, kernel_size=1, bias=False)
        self.lateral2 = nn.Conv2d(192, num_filters, kernel_size=1, bias=False)
        self.lateral1 = nn.Conv2d(64, num_filters, kernel_size=1, bias=False)
        self.lateral0 = nn.Conv2d(32, num_filters // 2, kernel_size=1, bias=False)

        for param in self.inception.parameters():
            param.requires_grad = False

    def unfreeze(self):
        for param in self.inception.parameters():
            param.requires_grad = True

    def forward(self, x):

        # Bottom-up pathway, from ResNet
        enc0 = self.enc0(x)
        enc1 = self.enc1(enc0)    # 256
        enc2 = self.enc2(enc1)    # 512
        enc3 = self.enc3(enc2)    # 1024
        enc4 = self.enc4(enc3)    # 2048

        # Lateral connections

        lateral4 = self.pad(self.lateral4(enc4))
        lateral3 = self.pad(self.lateral3(enc3))
        lateral2 = self.lateral2(enc2)
        lateral1 = self.pad(self.lateral1(enc1))
        lateral0 = self.lateral0(enc0)

        # Top-down pathway
        pad = (1, 2, 1, 2)    # pad last dim by 1 on each side
        pad1 = (0, 1, 0, 1)
        map4 = lateral4
        map3 = self.td1(
            lateral3 + nn.functional.upsample(map4, scale_factor=2, mode="nearest")
        )
        map2 = self.td2(
            F.pad(lateral2, pad, "reflect")
            + nn.functional.upsample(map3, scale_factor=2, mode="nearest")
        )
        map1 = self.td3(
            lateral1 + nn.functional.upsample(map2, scale_factor=2, mode="nearest")
        )
        return F.pad(lateral0, pad1, "reflect"), map1, map2, map3, map4
```